\documentclass{article} 
\usepackage{fullpage}
\usepackage{authblk}

\usepackage[american]{babel}

\usepackage{natbib} 
\usepackage{mathtools} 
\usepackage{booktabs} 
\usepackage{tikz} 
\usepackage{varwidth,array,ragged2e}

\newcolumntype{M}{>{$}l<{$}}
\newcolumntype{L}{>{\varwidth[c]{\linewidth}}l<{\endvarwidth}}
\title{\textsc{Modeling Extremes with $d$-max-decreasing Neural Networks}}
\usepackage[utf8]{inputenc} 
\usepackage[T1]{fontenc}    

\usepackage{url}            
\usepackage{booktabs}       
\usepackage{amsfonts}       
\usepackage{nicefrac}       
\usepackage{microtype}      
\usepackage{bm}
\usepackage{bbm}
\usepackage{amsmath}
\usepackage{amssymb}
\usepackage{amsthm}
\usepackage{xcolor}
\usepackage{graphicx}
\usepackage{comment}
\usepackage{xr}
\usepackage{subcaption}
\usepackage{bbold}
\usepackage{algorithm}
\usepackage{algorithmic}
\newtheorem{theorem}{{\bf Theorem}}

\newtheorem{proposition}{{\bf Proposition}}
\newtheorem{remark}{{\bf Remark}}

\newtheorem{definition}{{\bf Definition}}

\newcommand{\clamp}{\operatorname{clamp}}

\usepackage{array}
\usepackage{wrapfig}
\usepackage{multirow}
\usepackage{tabularx}
\makeatletter
\newcommand*{\addFileDependency}[1]{
  \typeout{(#1)}
  \@addtofilelist{#1}
  \IfFileExists{#1}{}{\typeout{No file #1.}}
}
\makeatother

\usepackage{subfiles}
\author[1]{Ali Hasan}
\author[2]{Khalil Elkhalil}
\author[2]{Yuting Ng}
\author[3]{Jo\~ao Pereira}
\author[1]{Sina Farsiu}
\author[4]{Jose Blanchet}
\author[2]{Vahid Tarokh}
\affil[1]{Department of Biomedical Engineering, Duke University}
\affil[2]{Department of Electrical Engineering, Duke University}
\affil[3]{Department of Mathematics, University of Texas at Austin}
\affil[4]{Department of Management Science and Engineering, Stanford University}

\begin{document}
\maketitle
\begin{abstract}

We propose a novel neural network architecture that enables non-parametric calibration and generation of multivariate extreme value distributions (MEVs). MEVs arise from Extreme Value Theory (EVT) as the necessary class of models when extrapolating a distributional fit over large spatial and temporal scales based on data observed in intermediate scales. In turn, EVT dictates that $d$-max-decreasing, a stronger form of convexity, is an essential shape constraint in the characterization of MEVs. As far as we know, our proposed architecture provides the first class of non-parametric estimators for MEVs that preserve these essential shape constraints. We show that our architecture approximates the dependence structure encoded by MEVs at parametric rate. Moreover, we present a new method for sampling high-dimensional MEVs using a generative model. We demonstrate our methodology on a wide range of experimental settings, ranging from environmental sciences to financial mathematics and verify that the structural properties of MEVs are retained compared to existing methods.

\end{abstract}
\section{Introduction}

\begin{figure}
    \centering
    \includegraphics[width=0.4\textwidth]{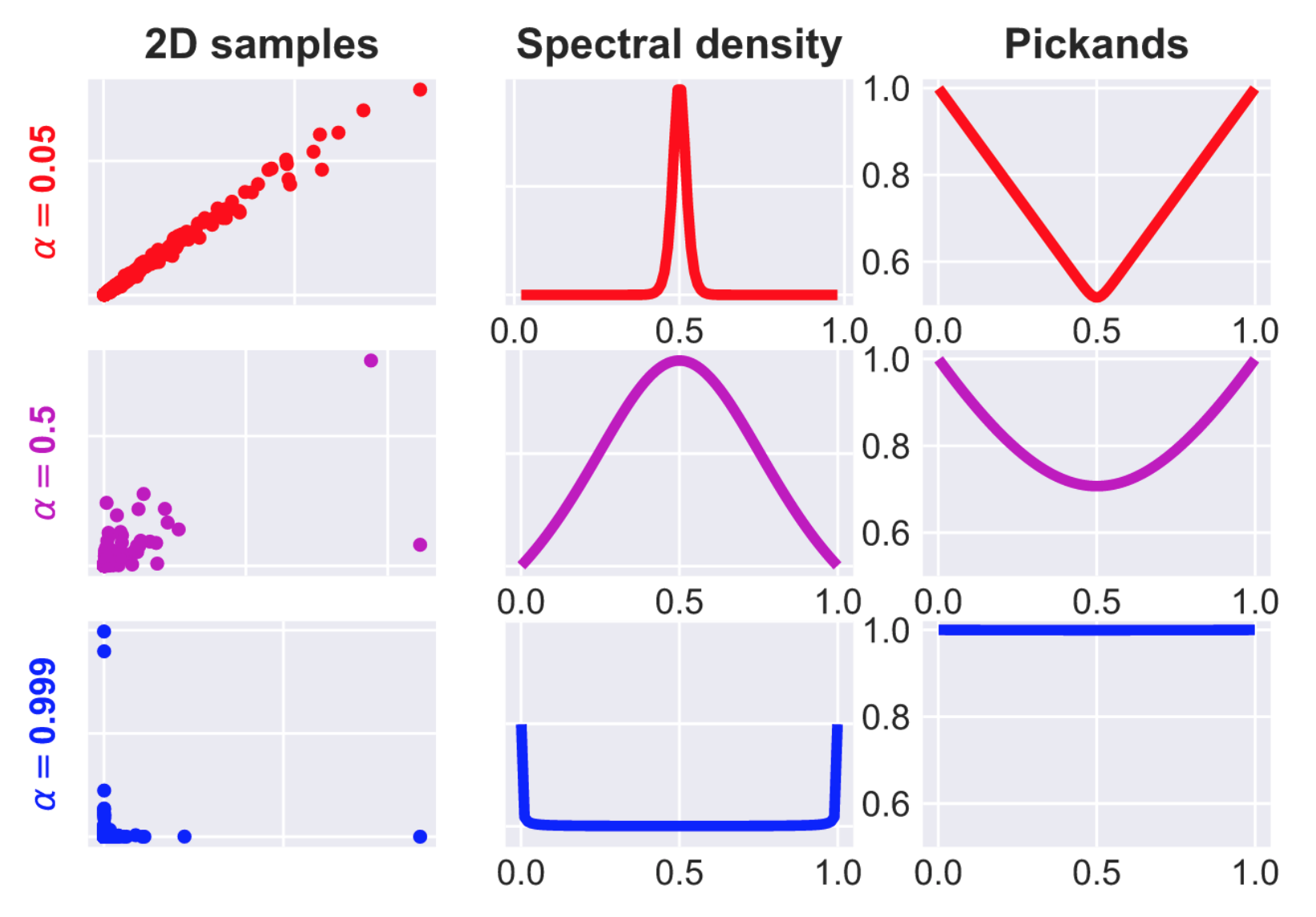}
    \caption{ 
    Equivalent representations of MEVs in dimension two, from dependent on the top row to independent at the bottom row. Left column, samples from MEV; Middle column, spectral representation; Right column, Pickands dependence function. We propose methods for estimating the Pickands function (section~\ref{sec:dmnn}), recovering the spectral density (section~\ref{sec:generative}) and sampling MEVs (section~\ref{sec:sampling}).}
    \vspace{-15pt}
    \label{fig:illustration}
\end{figure}
Modeling the occurrence of extreme events is an important task in many disciplines such as medicine, environmental science, engineering, and finance.
For example, understanding the probability of a patient having an adverse reaction to medication or the distribution of economic shocks is critical to mitigating the associated effects of these events~\citep{dey2016extreme}. 
However, these events are rare in occurrence and therefore are often difficult to characterize with traditional statistical tools. 
This has been the primary focus of extreme value theory (EVT), which describes how to extrapolate the occurrence of rare events outside the range of available data.
In the one-dimensional case, EVT provides remarkably simple models for the asymptotic distribution of the maximum of an infinite number of independent and identically distributed (i.i.d.) random variables, which is due to the celebrated Fisher-Tippet-Gnedenko theorem~\citep{embrechts_book}. 
These are known as the generalized extreme value (GEV) distributions \citep{dehaan_book}.

Perhaps more relevant to practical use-cases is to consider simultaneous extremes in the multi-dimensional scenario.
For example, how are extreme weather patterns related in geographical areas or how do extremes of different financial instruments relate?   
Unlike the one-dimensional case, multivariate extreme value (MEV) distributions generally do not endow simple analytical forms of the underlying density.
This leads to difficulties in performing inference tasks using conventional methods.
Instead, MEV distributions are characterized by tail dependence functions embedded in extreme value copulas~\citep{pickands1981multivariate,segers_copulas}.
\subsection*{Background: Extreme Value Copulas}
We start with a brief overview of multivariate EVT and provide additional background material in Appendix~\ref{sec:bg}.
 Let $\Delta_{d-1}$ denote the unit $d-$dimensional simplex.
Let $X_i=(X_{1}^{(i)},\ldots,X_{d}^{(i)}) \in \mathbb{R}^d$ for $i \in \{1, \ldots, n \}$ be a sample of i.i.d. random vectors with common continuous probability distribution $F$, marginals $F_1, \ldots, F_d$ and copula $C_F$. Recall that $C_F:[0, 1]^d \to [0, 1]$ satisfies:
\begin{equation*}
     C_F(\mathbf{u}) = \mathbb{P} \left[F_1(X_{1}) \leq u_1, \ldots, F_d(X_{ d}) \leq u_d \right].
\end{equation*}
Let the vector of component-wise maxima be given by:
$
    M^{(n)} = \left(M_{1}^{(n)}, \ldots, M_{d}^{(n)} \right), 
$
where $M_{k}^{(n)} = \max_{i=1,\ldots,n} X_{k}^{(i)}$ for $k \in \{1, \ldots, d\}$.  
Let $C_n$ be the copula of $\bar{M}^{(n)}$ given by:
$
    \bar{M}^{(n)} = \left(\frac{M_{1}^{(n)} - b_{1}^{(n)}}{a_{1}^{(n)}}, \ldots, \frac{M_{d}^{(n)} - b_{d}^{(n)}}{a_{d}^{(n)}} \right),
$
where each component-wise maxima $M_{k}^{(n)}$ is normalized with sequences of real numbers $a_{k}^{(n)} > 0$ and $b_{k}^{(n)}$ such that the corresponding limiting marginal is non-degenerate.
Then the following property known as max-stability holds:
\begin{equation*}
    C_n(u_1, \ldots, u_d) = C_F(u_1^{1/n}, \ldots, u_d^{1/n})^n,\; \forall \; \mathbf{u}\in[0,1]^d.
\end{equation*}

We are interested in finding the limiting copula $C$ of $C_n$ as $n \to \infty$. 
The limiting copula is then called an extreme value copula and we say that $C_F$ is in the \emph{maximum domain of attraction} of $C$, denoted as $C_F \in \text{MDA}(C)$. The limiting extreme value copula $C$ has the form \citep{segers2012max}:
\begin{equation}
\label{eq:pickands_copula}
    \begin{split}
    C(\mathbf{u})  =  \exp\Bigg[&\left( \sum_{k=1}^d \log u_k\right) \\
    &A\left(\frac{\log u_1}{\sum_{k=1}^d \log u_k}, \ldots, \frac{\log u_d}{\sum_{k=1}^d \log u_k} \right) \Bigg] ,
    \end{split}
\end{equation}
where $A$ is known as a Pickands dependence function that defines the joint dependence of a MEV.

\begin{definition}[Pickands dependence function]
\label{def:pickands}
A function $A : \Delta_{d-1} \to [1/d, 1]$ is called a Pickands dependence function if it satisfies the following properties:
\begin{enumerate}
    \item $A$ is homogeneous of order 1 and fully $d$-max-decreasing where $d$ is the dimension;
    \item $A$ satisfies $\label{pickands_bounds} \max_{k = 1,\ldots,d} w_k \leq A(\mathbf{w}) \leq 1$ for all $\mathbf{w} \in \Delta_{d-1}$.
    \item $A(\mathbf{e}_k) = 1$ where $\mathbf{e}_k$ is the $k^\text{th}$ canonical basis vector.
\end{enumerate}
\end{definition}

We give the functional definition of fully $d$-max-decreasing in Appendix~\ref{sec:fdmd}\footnote{Intuitively, fully $d$-max-decreasing describes a stronger form of convexity needed to ensure that subsets of margins remain valid MEVs.
See \citet[Theorem 5.2.2]{hofmann2009characterization} and \citet[Theorem 6]{ressel2013homogeneous} for further details.} and instead give the spectral correspondence of $A$ here. 

\begin{definition}[Spectral form of Pickands dependence function]
\label{definition_integral}
For any Pickands dependence function $A$, there exists a Borel measure (spectral measure) $\Lambda$ on $\Delta_{d-1}$ satisfying $\int_{\Delta_{d-1}} s_k \, \mathrm{d} \Lambda(\mathbf{s}) = 1$ for $k \in \{1,\ldots,d\}$ such that 
\begin{equation}
A(\mathbf{w}) = \int_{\Delta_{d-1}} \max_{k=1,\ldots,d}w_k s_k \, \mathrm{d}\Lambda(\mathbf{s}), \:\: \mathbf{w} \in \Delta_{d-1}.
\label{eqn:pickands_integral}
\end{equation}
\end{definition}

The equality $\int_{\Delta_{d-1}} s_k \, \mathrm{d} \Lambda(\mathbf{s}) = 1$ is only used as a convention to standardize the margins, and is not essential in maintaining the $d$-max decreasing property~\citep{fougeres2013dense}.
To provide some intuition on the aims of this paper, Figure~\ref{fig:illustration} illustrates the relationship between different equivalent representations for a canonical parametric MEV -- the symmetric logistic distribution with dependence parameter $\alpha=0.05$ leaning towards complete dependence and $\alpha=0.999$ leaning towards complete independence. The proposed methods estimates the Pickands function (right most column) and recovers the spectral measure (middle column) which enables sampling MEVs (left most column).

\textbf{Related Work.}
A number of techniques have been developed to estimate extreme value copulas from data.
The most relevant to the present work is that by \cite{pickands1981multivariate} where a non-parametric estimator of the Pickands function was first proposed. Following works such as \citet{caperaa1997nonparametric} and  \citet{bucher2011new} describe alternative takes on estimating the dependence function. 
The above methods, however, do not guarantee that the estimate completely satisfy the conditions of a valid Pickands dependence function.
In \citet{marcon2017multivariate}, the authors consider a projection of a nonparametric estimator to a convex function represented as a Bernstein polynomial.
However, the number of parameters required significantly increases with both the amount of data and the dimensionality, making it difficult for higher dimensional problems or problems with many data points. 
Finally, a number of Pickands estimators were compared and described in \citet{vettori2018comparison}, and notably none of the estimators reviewed satisfied all requirements of the Pickands function in cases where $d > 2$.
For additional details, please refer to the review on extreme value copulas in \citet{segers_gudendorf_copulas}. 
A theoretical review of $d$-max-decreasing functions and their applications to copulas is given in \citet{ressel2019copulas}.

\textbf{Our Contributions.} 
\begin{enumerate}
    \item We present fully $d$-max-decreasing neural networks, an architecture constrained to represent Pickands dependence functions of MEVs.
    \item We prove that, in the limit, the proposed architecture can approximate arbitrary Pickands functions.
    \item We propose a generative neural network representation of the spectral density of Pickands functions.
    \item We propose an extension of the Pickands Estimator to train neural networks.
\end{enumerate}

\section{Neural Representations of Extreme Value Distributions}
\label{sec:Pickands_ICNN}
Our main results propose two architectures for representing MEVs: a deterministic method for representing the Pickands dependence function, and a stochastic method for representing the spectral measure. 
While both represent equivalent quantities, each is more suited for a particular task.
The deterministic representation is more suitable for estimating exceedance probabilities whereas the spectral representation is more suitable for sample generation.

\subsection{Fully $d$-max-decreasing Neural Networks}
\label{sec:dmnn}
We are interested in finding a flexible parameterization of $A$ that enforces all the properties given in Definition~\ref{def:pickands}. The most difficult property to enforce is being fully $d$-max-decreasing.
To that end, we propose a new architecture inspired by Maxout Networks~\citep{goodfellow2013maxout} and Input Convex Neural Networks (ICNNs)~\citep{amos2017input}.
The proposed architecture, dubbed \emph{$d$-max Neural Networks (dMNNs)}, has additional restrictions to fulfill the conditions of the Pickands dependence function.

\begin{theorem}[Fully $d$-max-decreasing Architecture]
\label{thm:arch}
Let $A_{\bm \theta}^{(m)}(\mathbf{w})$ be a function defined as:
\begin{equation}
 \begin{split}
& A_{\bm \theta}^{(m)}(\mathbf{w}) \\
& := \max \bigg( \max_{k = 1,\ldots,d} w_k,  L^{(m)}(\mathbf{w}) +(1 - L^{(m)}(\mathbf{e})^T \mathbf{w}) \bigg),
\end{split}
    \label{eq:arch}
\end{equation}
where
\begin{align*}
L^{(m)}(\mathbf{w}) &= \frac1{n_m}\sum_{j=1}^{n_m}\left ( \ell^{(m)} \circ \ell^{(m-1)}\circ \cdots \circ \ell^{(1)}(\mathbf{w}) \right)_j,  \\
\ell^{(i)}(\mathbf{h}^{(i-1)})_j &= \max_{k=1,\ldots,n_{i-1}} \left (\Theta_{j,\cdot}^{(i)} \odot {h}^{(i-1)} \right )_{k},\\
\mathbf{h}^{(i-1)} &= \ell^{(i-1)}\circ \cdots \circ \ell^{(1)}(\mathbf{w}),\\
L(\mathbf{e}) &=(L(\mathbf{e}_1),\ldots,L(\mathbf{e}_d))^T,
\end{align*}
$m$ is the number of layers, $n_i$ is the width of the $i^\text{th}$ layer, $\;\Theta^{(i)} \in  \mathbb{R}^{n_{i} \times n_{i-1}}_+$ are the weights of the $i^\text{th}$ layer, constrained to be all positive, and $\mathbf{e}_i$ is the $i^\text{th}$ canonical basis vector.
$\odot$ denotes component-wise multiplication.

Then, $A_{\bm\theta}^{(m)}(\mathbf{w})$ is a fully $d$-max-decreasing function. Moreover, $A_{\bm\theta}^{(m)}(\mathbf{w})$ represents a valid Pickands dependence function.
\end{theorem}

\begin{proof}[Intuition of proof]
The proof uses the idea that $\mathbb{E}_\mathbf{s}[\max_{k=1,\ldots,d}(w_ks_k)], \:\: \mathbf{s} \in \Delta_{d-1}$ is fully $d$-max-decreasing and certain compositions of this function retain this property. The full proof is given in Appendix~\ref{sec:proof_arch}.
\end{proof}



For notational convenience, we drop the $(m)$ unless needed. 
To get an intuition behind the structure of the architecture, note that in the single layer case in the limit as $n_1\to\infty$, the weights $\bm \theta$ correspond to samples of the spectral measure in Definition~\ref{definition_integral} and the expectation is computed empirically. 
While the proposed architecture is guaranteed to enforce the properties of the Pickands function, and is thus fully $d$-max-decreasing, we are also interested in seeing how well it can approximate an arbitrary Pickands dependence function.
We present results in the following theorem:

\begin{theorem}[Uniform Convergence]
\label{thm:approx}
Suppose that $\bm \theta$ are samples from the true spectral measure and $A$ is the true Pickands function.
The empirical process $$\mathbb{G}_n = \sqrt{n} \left ( A_{\bm \theta}^{(1)}(\mathbf{w}) - A(\mathbf{w}) \right )$$ converges to a zero mean Gaussian process as $n \to \infty$ where $A_{\bm \theta}^{(1)}$ is a single layer $d$MNN of width $n$.
\end{theorem}
\begin{proof}[Intuition of proof]
We first establish pointwise convergence. Then we show $A$ is Lipschitz over a bounded set whose covering number grows in accordance with functions that are $P-$Donsker. 
The full proof is given in Appendix~\ref{sec:proof_pointwise}. 
\end{proof}
The result in Theorem~\ref{thm:approx} has many implications on the properties of the proposed network since it, for example, allows us to quantify the uncertainty associated with our function estimates.
Using the proposed architecture, we mitigate issues faced by previous estimators, such as \citep{bucher2011new, caperaa1997nonparametric, marcon2017multivariate}, in enforcing the $d$-max-decreasing property, inequalities, and endpoints of the function. 

\subsection{A Generative Model for the Spectral Measure}
\label{sec:generative}
While the spectral measure can be computed from the weights of the proposed $d$MNN, we propose an alternative representation of the spectral measure using a generative neural network. We model $\mathbf{y} \sim \Lambda$ in \eqref{eqn:pickands_integral} as the output of a generative neural network $G( \, \cdot \,; \bm \phi) \in \mathbb{R}^d_+$ with parameters $\bm \phi$, i.e. $\mathbf{y} = G(\mathbf{z}; \bm \phi)$ which maps input samples $\mathbf{z} \sim p_z$ to $\mathbf{y}$, where $p_z$ is a distribution that is easy to sample from (such as a multivariate Gaussian distribution). 
This leads us to a representation of $A$ in terms of the generator:
\begin{equation}
A_G(\mathbf{w}) := \mathbb{E}_{\mathbf{y} \sim G} \left [ \max_{k=1,\ldots,d} w_k y_k \right],
\label{eq:generator}
\end{equation}
where $\mathbb{E}[y_k] = 1$. 
The expectation is taken empirically with a large number of samples from $G$.
\begin{remark}
The function given by~\eqref{eq:generator} satisfies all the necessary conditions for a valid Pickands function.
\label{rmk:generator}
\end{remark}
Following Remark~\ref{rmk:generator}, we informally note that it follows from the universal approximation theorem of neural networks that if $G$ is sufficiently expressive then \eqref{eq:generator} can represent an arbitrary Pickands dependence function.

\paragraph{Use Cases of Each Representation.}
The difference between the representation given by the $d$MNN~\eqref{eq:arch} and the generative neural network~\eqref{eq:generator} is: in the $d$MNN case the spectral measure is modeled by a discrete number of elements as dictated by the $d$MNN architecture, while in the generator case the implicit distribution of the spectral measure is modeled. 
The $d$MNN is useful in representing probabilistic quantities since it provides a deterministic representation of the CDF and therefore
it does not exhibit the variance of the generative representation. On the other hand, the generative model is capable of simulating many realizations of the MEV, particularly useful for sampling applications.
\section{Parameter Estimation}
Fitting data to high dimensional copulas is often a difficult task since the probability density function (PDF) is not directly modeled. 
In general, specific parametric families are used to make the process easier, such as in Archimedean copulas.
While it is theoretically possible to first obtain the underlying PDF via differentiating the CDF and then fit the $d$MNN with Maximum Likelihood Estimation (MLE), the procedure is computationally complex, especially in high dimensions.
The main drawback of such a method lies in the need to differentiate the $d-$variate CDF, since nested differentiation with existing automatic differentiation methods may result in numerical errors \citep{margossian2019review}.
Instead, we use specific properties of MEVs to transform the parameter fitting procedure into MLE over univariate random variables.
We additionally present the analogs for survival distributions in Appendix~\ref{sec:survival}. 

\subsection{Fitting the Dependence Function}
Let $F_k$ denote the univariate marginal CDF (which can be fitted using MLE as in \cite{embrechts_book} or the $L$-moments method of \cite{L_moments}) of the $k^\text{th}$ normalized component wise maxima $\bar{M}_k^{(n)} = \frac{M_k^{(n)} - b_k^{(n)}}{a_k^{(n)}}$, $k \in \{1, \ldots, d\}$. In addition, let $\mathbf{w}=\left(w_1, \ldots, w_d\right) \in \Delta_{d-1}$. 
We introduce the transformation on $\bar{M}_k^{(n)}$:
\begin{align}
    \label{transform1}
    \widetilde{M}_k^{(n)} & = - \log (F_k(\bar{M}_k^{(n)})), \: \forall k \in \{1, \ldots, d\}, \\ 
    \label{transform2}
    Z_w & = \min_{k=1,\ldots,d} \widetilde{M}_k^{(n)} / w_k.
\end{align}
Then, we have: $\mathbb{P} \left[ Z_w > z\right] = e^{- z A(\mathbf{w})}$ 
(for the full derivation, see Section 3 of \cite{segers_gudendorf_copulas}).
This transformation casts the original multi-dimensional distribution into the new variables $Z_w$ that are exponentially distributed with rate parameter given by the Pickands dependence function $A(\mathbf{w})$.
From this transformation, we can fit the model $A_{\bm \theta}(\mathbf{w})$ to samples $Z_w$ using MLE. 
This can be done by training the model $A_{\bm \theta}(\mathbf{w})$ with stochastic gradient descent (SGD) to match the data points $Z_w$ as follows:
\begin{align}
    \label{MLE_loss}
    A^{\star}_{\bm \theta}(\mathbf{w}) = \arg \min_{ \bm \theta} \mathbb{E}_{Z_w} \mathcal{L}(Z_w; {\bm \theta}),
\end{align}
where $\mathcal{L}(Z_w; {\bm\theta}) = A_{\bm\theta}(\mathbf{w}) Z_w - \log A_{\bm \theta}(\mathbf{w})$.
Alternative losses could be considered by reformulating the loss with respect to the estimators defined in \citet{bucher2011new} and \citet{caperaa1997nonparametric}.
We empirically found that the MLE approach described in~\eqref{MLE_loss} provides the best performance, and it follows naturally from the original formulation of~\citet{pickands1981multivariate}. 
The training procedure is summarized in Algorithm~\ref{alg_train}. 

\begin{algorithm}[h!]
	\caption{Fitting the Pickands-$d$MNN to Data} 
	\label{alg_train}
	\begin{algorithmic}[1]
	\STATE \textbf{Input:} $\left \{ \left(X_1^{(i)}, \ldots, X_d^{(i)} \right) \right \}_{i=1}^N$, $N=B \times n$ samples of i.i.d. random vectors where $B$ is the number of blocks of data and $n$ is the size of each block.
	\STATE Take component-wise maxima over each block: $\left \{ \left(M_{1}^{(n,b)}, \ldots, M_{d}^{(n,b)} \right)\right \}_{b=1}^B$ where $M_{k}^{(n,b)}=\max _{i=(b-1)n+1,...,bn}X_k^{(i)}$, $(k, b) \in \{1, \ldots, d\} \times \{1, \ldots, B\}$. 
	\STATE Fit a GEV to each component-wise maxima $\{ M_{k}^{(n,b)} \}_{b=1}^B$, obtain $\{\bar{M}_{k}^{(n,b)} \}_{b=1}^B$, then estimate marginals $F_k$ for each $k \in \{1, \ldots, d \}$.
	\STATE \textbf{Initialize} the parameters ${\bm \theta} \geq 0$ of the $d$MNN \\ 
    \textbf{Repeat}: 
    \STATE Randomly sample a minibatch of training data $\{\bar{M}_{k}^{(n,b)} \}_{b \in \text{batch}}$
    and uniformly sample $\mathbf{w} \in \Delta_{d-1}$.
    \STATE  Transform samples according to Equations \eqref{transform1} and \eqref{transform2} to obtain transformed samples $\{Z_{w, b} \}_{b \in \text{batch}}$. 
    \STATE Compute gradient
     $\nabla_{{\bm\theta}}  \sum_{b \in \text{batch}} \mathcal{L}\left(Z_{w, b}; {\bm \theta} \right)$.\\
    \STATE Update $\bm \theta$ with Adam \citep{adam} \\ 
    \textbf{Until} convergence \\ 
    \textbf{Output:} $A^{\star}_{\bm \theta}(\mathbf{w})$. 
	\end{algorithmic} 
\end{algorithm}

\subsection{Fitting the Generator}
Recall that we have an equivalent representation of $A$ given by $A_G$ in \eqref{eq:generator} where $G(\cdot; \bm \phi)$ is a function, with parameters $\bm \phi$, of random variables.
We fit the parameters $\bm \phi$ of the generator by solving the following optimization problem:
\begin{equation}
  \min_{\bm \phi} \mathbb{E}_{Z_w} \mathcal{L}(Z_w;\bm \phi)
 + \eta \left\| \mathbb{E}_{\mathbf{y}  } [\mathbf{y}] - \mathbf{1}_d \right\|_2^2,
\label{eqn:segers_opt}
\end{equation}
with $\mathcal{L}$ now defined using the representation of $A_G$ in \eqref{eq:generator}:
 $$
 \mathcal{L}(Z_w; \bm \phi) = \mathbb{E}_{\mathbf{y}}[\max_{k=1\ldots d}y_k w_k]Z_w - \log \mathbb{E}_{\mathbf{y}}[\max_{k=1\ldots d}y_k w_k]
 $$
 where $\mathbf{y} = (y_1, \ldots, y_d) =  G(\mathbf{z}; \bm \phi)$, $\mathbf{y} \in \mathbb{R}^d_+$ and $\mathbf{z} \in \mathbb{R}^k \sim p_z$ with $\eta >0$ as a regularization factor. 
 Note that the second expectation in~\eqref{eqn:segers_opt} is only needed to enforce the margins. It need not be strictly enforced, enforcing approximately only results in minor changes in the tail index. 
The expectations with respect to $\mathbf{y}$ in \eqref{eqn:segers_opt} are approximated using the sample mean with samples from the generator. 

To summarize the parameter estimation section, we bypass the need to differentiate the CDF and use properties of MEVs to estimate the parameters of the distribution from data. 
Both representations of the Pickands function presented can be used with this technique. 
\section{Sampling}
\label{sec:sampling}
While learning MEV distributions from data is important for computing probabilities, it is also useful to simulate possible scenarios by sampling from an estimated MEV distribution.
We introduce a sampling technique using the proposed architectures to efficiently sample from arbitrary MEVs.
To the best of our knowledge, there are no general sampling methods for arbitrary extreme value copula that scale to high dimensions.
This is because MEV sampling algorithms assume knowledge of the spectral measure, and do not consider sampling when given only the Pickands function.
It then becomes necessary to recover the spectral measure from a given Pickands function or from data, which we previously described two methods for doing so.
We additionally note that the traditional method of conditional sampling for copulas is ineffective since it requires both computing high order derivatives and using numerical root-finding techniques.
We base our sampling procedure on algorithms for the infinite dimensional analogue of MEV distributions known as \emph{max-stable processes} \citep{dombry2016exact}. 
Max-stable processes have the property that finite dimensional marginals are MEVs and have a spectral representation in terms of the spectral measure $\Lambda$ for stationary processes.
This ultimately allows us to recast MEV sampling in terms of prior work on sampling from max-stable processes, where established methods exist.

\subsection{Margins of Max-Stable Processes as MEV Distributions}
A stationary max-stable process has the form: 
\begin{equation}
\label{eq:max_stable}
\max_{i \geq 1} \xi_i y_i(x), \:\: x \in \mathbb{X} \subset \mathbb{R}^k
\end{equation}
where $\xi_i$ is the $i^\text{th}$ realization of a Poisson point process with intensity $\xi^{-1} \mathrm{d} \xi$. 
$y_i$ is the $i^\text{th}$ sample from the spectral measure. 
Additionally, $\mathbb{E}[y(x)] = 1, x \in \mathbb{X}$ is generally assumed to enforce unit Frechet margins. 
For a finite number $d$ of $\{x_j\}_{j=1}^d$, this corresponds to a $d$-dimensional spectral measure with the same properties as in Definition~\ref{definition_integral}. 
The key idea is to use the representation in \eqref{eq:max_stable} to sample from the full MEV distribution with only knowledge of the spectral measure. 
We use the algorithm mentioned in \citet[Algorithm 1]{hofert2018hierarchical} for sampling from the full distribution given samples of the spectral measure. We give the details of the algorithm in Appendix~\ref{sec:algs} Algorithm~\ref{alg:sampling}.

\subsection{Sampling from the $d$MNN}
Suppose we fit a single layer $d$MNN using Algorithm~\ref{alg_train} with weights given by ${\bm \theta } \in \mathbb{R}_+^{w \times d}$ where $w$ is the width of the network and $d$ is the data dimension.
Consider the transformation $\hat{\theta}_{i,j} = \theta_{i,j} / \sum_{j=1}^d \theta_{i, j}$ where we transform the weights of the network to the unit simplex $\Delta_{d-1}$, and $i,j$ refer to the row and column indices. 

We then choose a number $N$ and compute
$$
\max_{i = 1, \ldots, N} \xi_i \hat{\theta}_{i + j}, \quad j \sim \text{rand}(\{1,\ldots, w-N\})
$$
where $\xi_i$ is defined as per~\eqref{eq:max_stable}.
While this method is effective in sampling, a possible issue is the finite number of $\hat{\theta}$ dictated by the width $w$ of the network. 
The generative model on the other hand allows for unlimited generation of samples of the spectral measure. 


\subsection{Sampling from the Generative Model}
Suppose we fit a generative model $G(z; \bm \phi)$ to data following the optimization procedure in \eqref{eqn:segers_opt}. 
Then sampling proceeds similarly to the case with the $d$MNN except in this case we do not use the weights of the network explicitly, but sample from the model:
$$
\max_{i=1,\ldots,N}\xi_i y_i \:\: \text{where} \:\: y_i = G(z_i; \bm \phi), z_i \sim p(z)
$$
where the notation is maintained as above with $p(z)$ defining an easy to sample prior distribution.

As a final note regarding the sampling methods, one particularly useful way of combining the methods is to first estimate $A_{\bm\theta}$ from data using an estimator such as the $d$MNN.
Then, fit the generator to $A_{\bm\theta}$ by taking the mean squared error (MSE) between the two representations, i.e.
\begin{align*}
{\bm \phi}^\star = \arg \min_{\bm \phi} &\left \| A_{\bm\theta} - \mathbb{E}_{\mathbf{y} \sim G_{\bm \phi}} \left[ \max_{k=1, \ldots, d} w_k y_k \right] \right \| \\
& + \eta \| \mathbb{E}[\mathbf{y}] - \mathbf{1}_d \|.
\end{align*}
This provides a simple way to recover the spectral density of any given EVC and thus an effective way to sample from arbitrary MEVs.
We detail this algorithm in Appendix~\ref{sec:algs} Algorithm~\ref{alg:train_gen}.
\section{Results}
\label{sec:results}
\begin{figure}
    \centering
\begin{subfigure}{.22\textwidth}
  \centering
  \includegraphics[width=\linewidth]{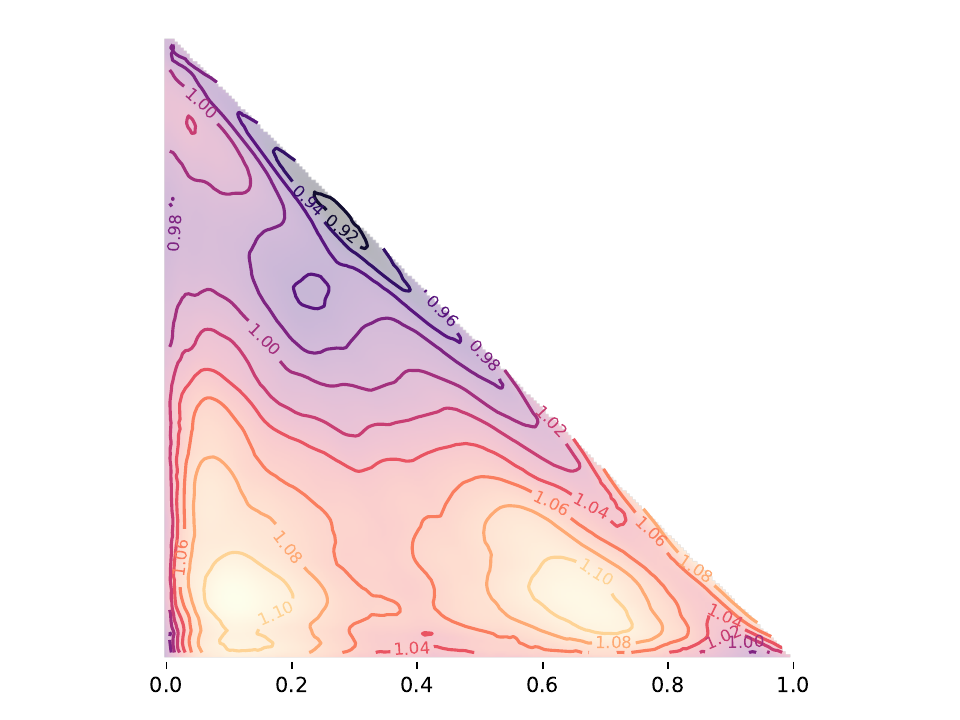}  
  \caption{Pickands 3d Margins}
  \label{fig:pick_marg3}
\end{subfigure}
\begin{subfigure}{.22\textwidth}
  \centering
  \includegraphics[width=\linewidth]{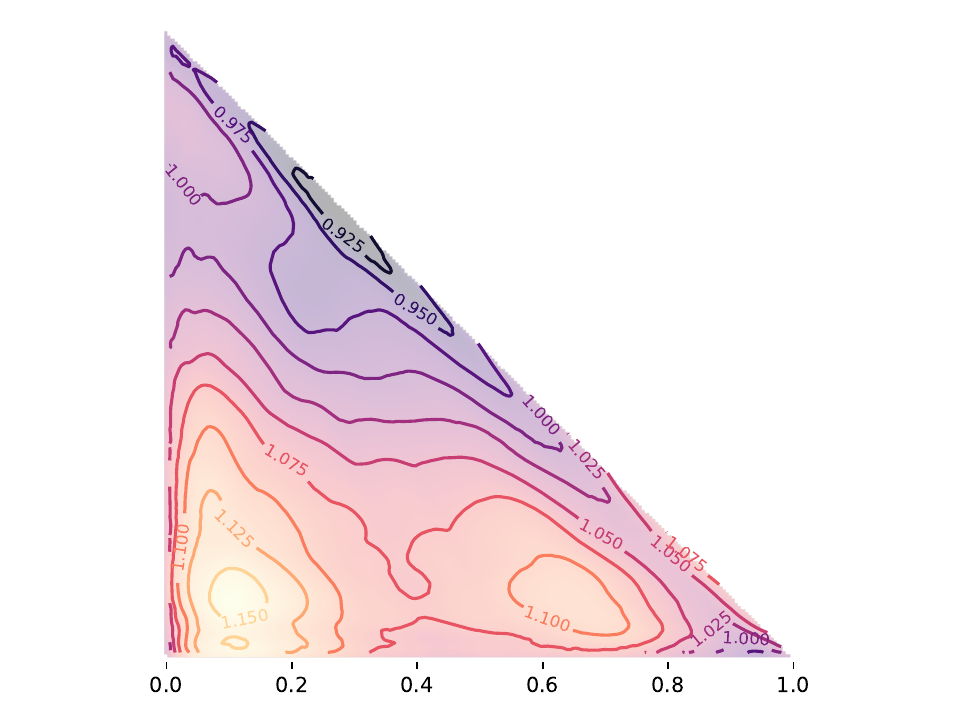}  
  \caption{CFG 3d Margins}
  \label{fig:cfg_marg3}
\end{subfigure}
\begin{subfigure}{.22\textwidth}
  \centering
  \includegraphics[width=\linewidth, trim=2pt 0pt 2pt 0pt]{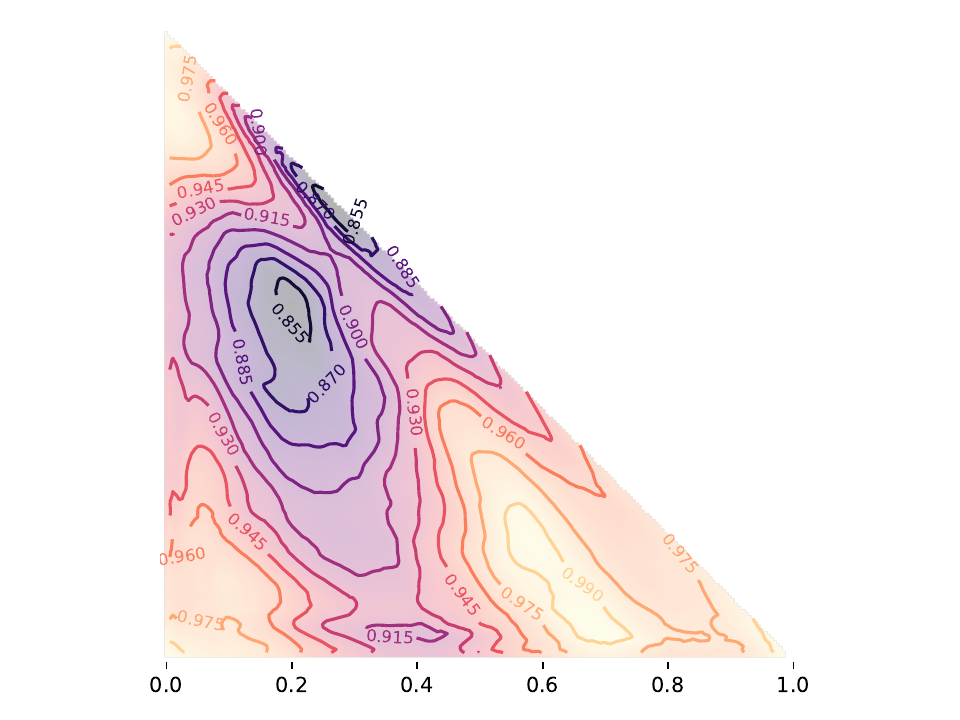}  
  \caption{BDV 3d Margins}
  \label{fig:bdv_marg3}
\end{subfigure}
\begin{subfigure}{.22\textwidth}
  \centering
  \includegraphics[width=\linewidth]{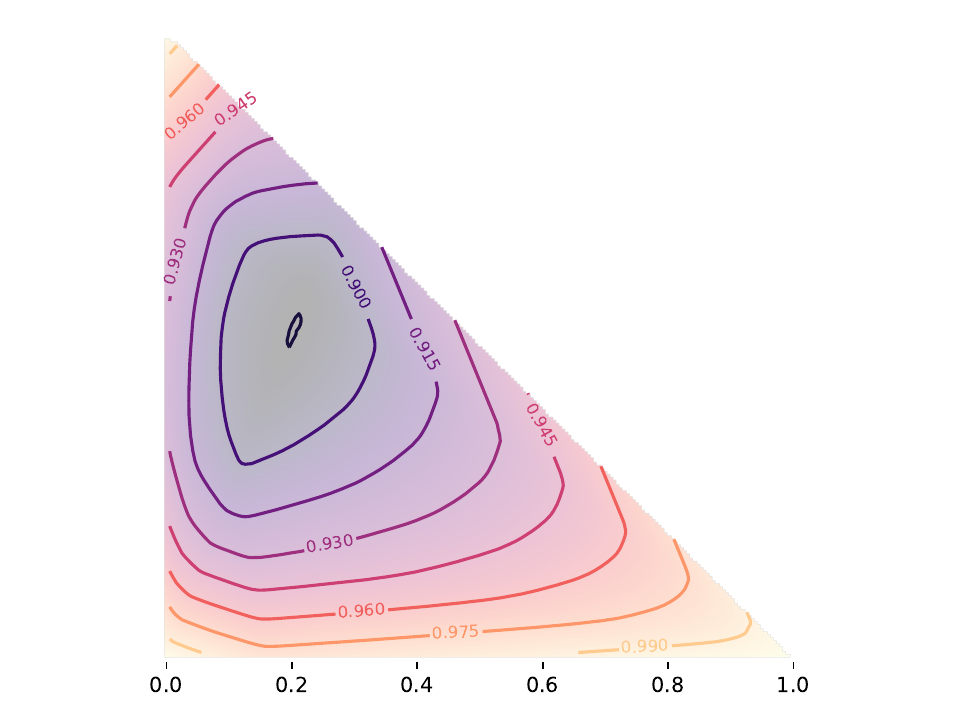}  
  \caption{$d$MNN 3d Margins}
  \label{fig:net_marg3}
\end{subfigure}
\label{fig:marg3}
\caption{Qualitative comparison of 3d margins from learned 10d MEV for the commodities dataset. The $d$MNN retains margins that are valid Pickands dependence function. The other estimators are non-convex and outside the required bounds. Contours plotted with solid lines. See additional figures in Appendix~\ref{sec:large_figs} and~\ref{sec:more_experiments}, Figures~\ref{fig:net_marg_ozone} to \ref{fig:net_marg_crypto}.}
\end{figure}
In this section, we provide numerical results that compare the estimation capabilities of the proposed $d$MNN-based model with well-known estimators from the literature: Pickands \citep{pickands1981multivariate}, CFG \citep{caperaa1997nonparametric}, and the estimator described in \citep{bucher2011new} which we refer to as BDV. These estimators are described in greater detail in Appendix~\ref{sec:estimators}. 
We start by evaluating the performance for estimating survival probabilities on known parametric models, followed by real data.
We conclude with experiments on sampling from a MEV, where we use the proposed generative model for high dimensional data with different dependence structures.
To align with the results in Theorem~\ref{thm:approx}, for the experiments presented in this section, we use a single layer $d$MNN with a width of $512$. Additional experiments with two different architectures are presented in Appendix~\ref{sec:more_experiments}.

\begin{figure}
    \centering
\begin{subfigure}{.23\textwidth}
  \centering
  \includegraphics[width=\linewidth]{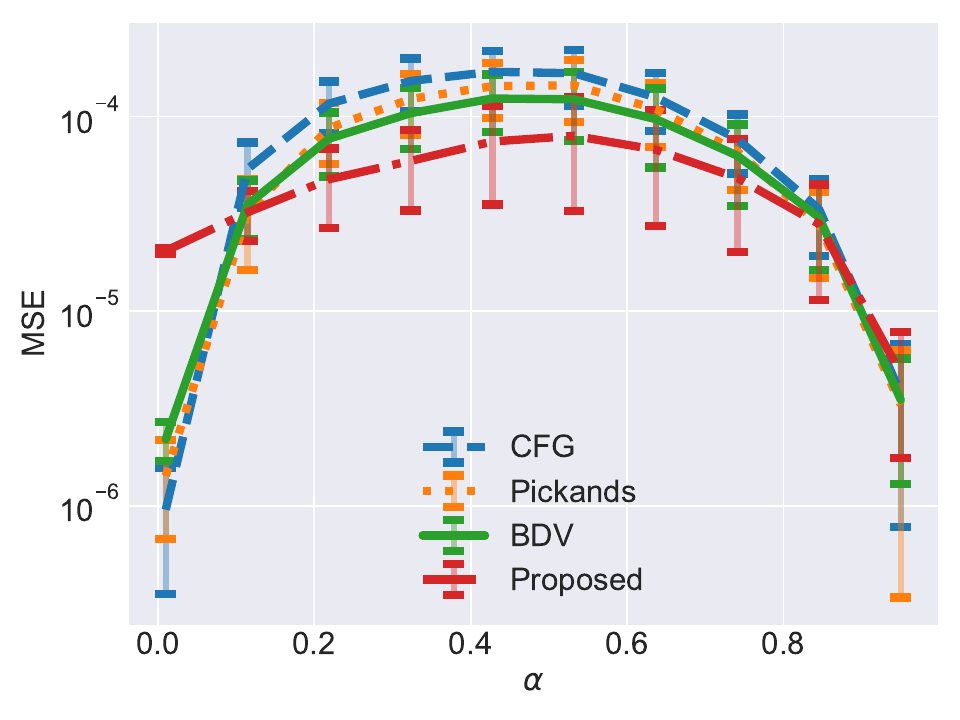} 
  \caption{$A_\text{SL}$ MSE ($d=2$)}
  \label{fig:sl_survival}
\end{subfigure}
\begin{subfigure}{.23\textwidth}
  \centering
  \includegraphics[width=\linewidth]{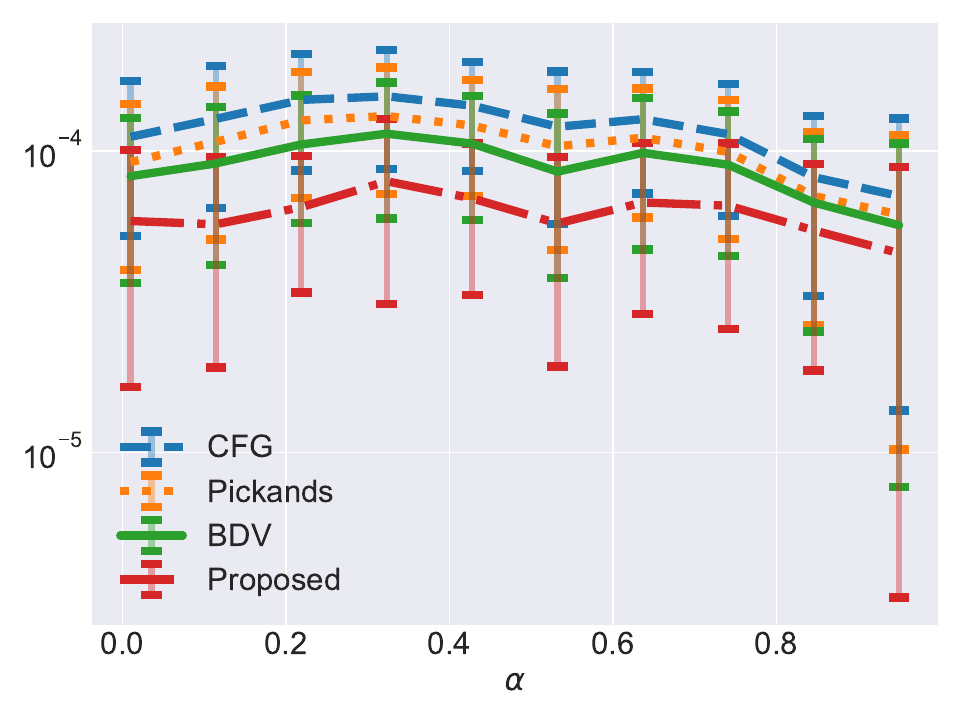}
  \caption{$A_\text{ASL}$ MSE ($d=2$)}
  \label{fig:asl_survival}
\end{subfigure} 
\caption{MSE of survival probabilities for $d=2$ with $100$ samples for $A_\text{SL}$ (\ref{fig:sl_survival}) and $A_\text{ASL}$ (\ref{fig:asl_survival}). Thresholds are above the $75$th percentile.}
\end{figure}
\begin{figure*}
    \centering
\begin{subfigure}{.247\textwidth}
  \centering
  \includegraphics[width=\linewidth]{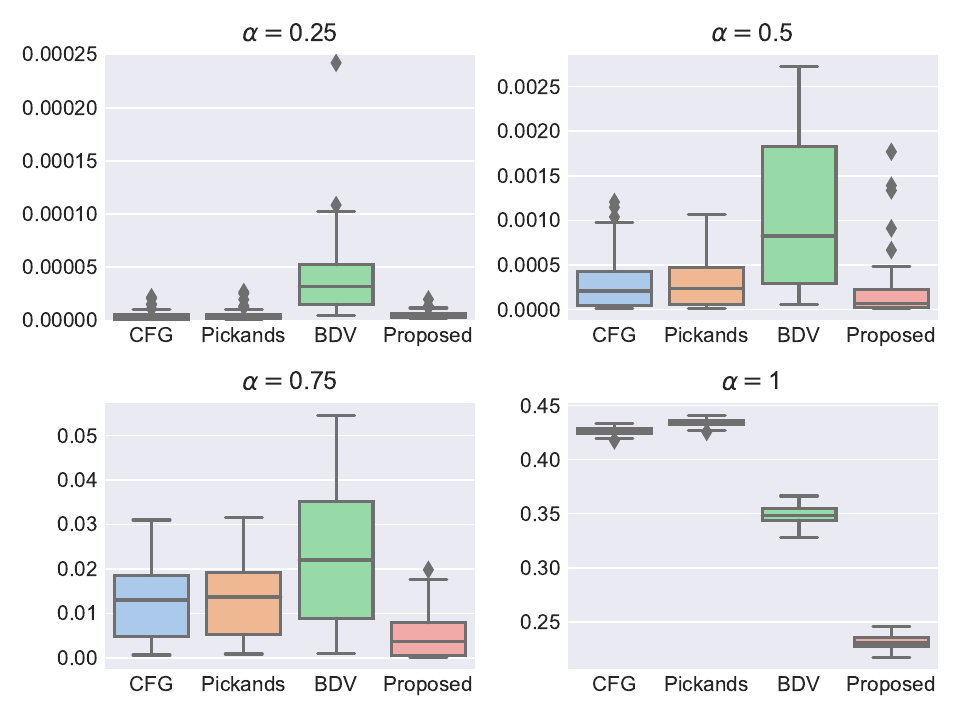} 
  \caption{$A_\text{SL}$ MSE ($d=256$)}
  \label{fig:sl_mse_est_all_a}
\end{subfigure}
\begin{subfigure}{.247\textwidth}
  \centering
  \includegraphics[width=\linewidth]{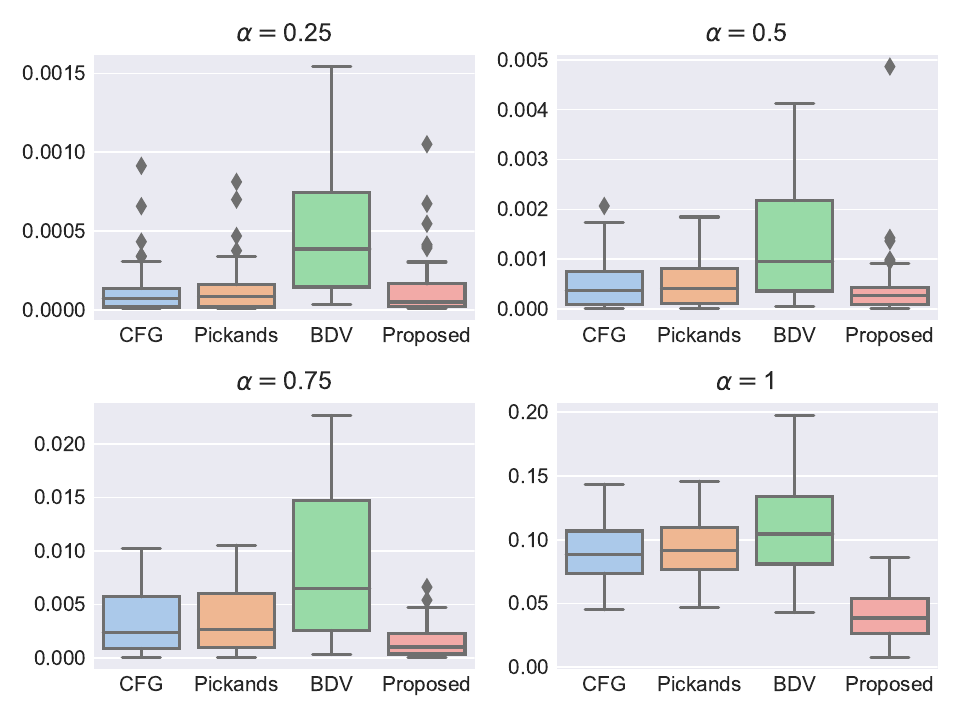}
  \caption{$A_\text{ASL}$ MSE ($d=256$)}
  \label{fig:asl_mse_est_all_a}
\end{subfigure} 
\begin{subfigure}{.247\textwidth}
  \centering
\includegraphics[width=\linewidth]{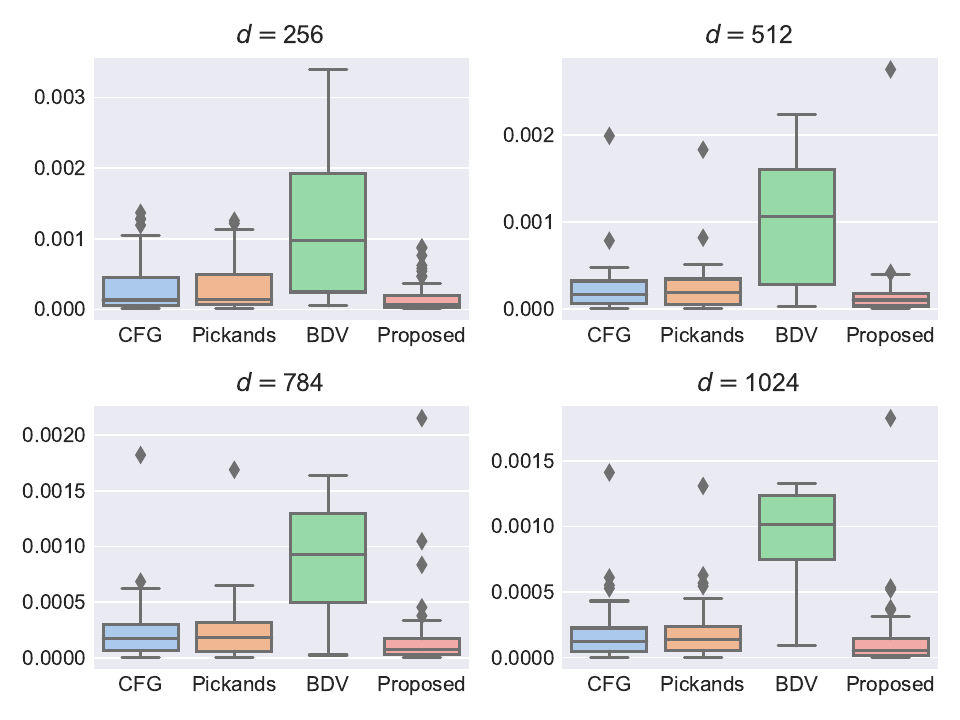}  
  \caption{$A_\text{SL}$ MSE ($\alpha=0.5$)}
  \label{fig:sl_mse_est_all_d}
\end{subfigure}
\begin{subfigure}{.24\textwidth}
  \centering
    \includegraphics[width=\linewidth]{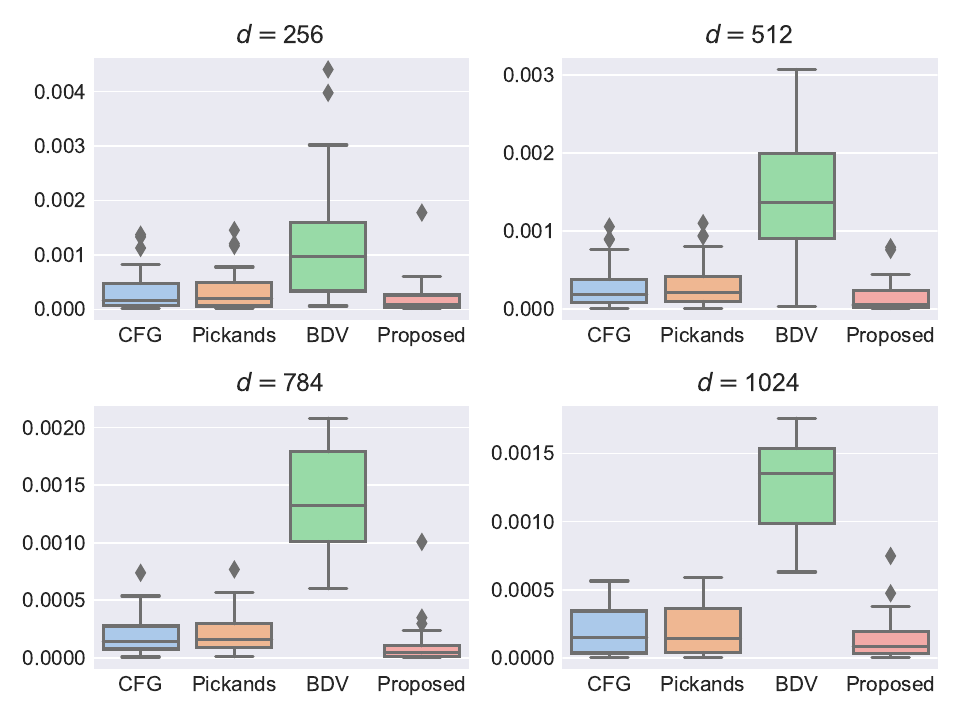}  
  \caption{$A_\text{ASL}$ MSE ($\alpha=0.5$)}
  \label{fig:asl_mse_est_all_d}
\end{subfigure}  

\caption{Comparison of $||\hat{A}(\mathbf{w}) - A(\mathbf{w})||_2^2$ for different estimators $\hat{A}$ for different dependence $\alpha = \{0.25, 0.50, 0.75, 1.0\}$ with fixed $d=256$ (\ref{fig:sl_mse_est_all_a}, \ref{fig:asl_mse_est_all_a}) and for fixed  $\alpha=0.5$ with different $d = \{256, 512, 728, 1024\}$ (\ref{fig:sl_mse_est_all_d}, \ref{fig:asl_mse_est_all_d}). The reference $A(\mathbf{w})$ are $A_\text{SL}$ (\ref{fig:sl_mse_est_all_a}, \ref{fig:sl_mse_est_all_d}) and $A_\text{ASL}$ (\ref{fig:asl_mse_est_all_a}, \ref{fig:asl_mse_est_all_d}). Results are over 50 runs with 100 training samples for each run.} 
    \label{fig:sl_asl_mse}
\end{figure*}
\begin{table}[h!]
\footnotesize
    \centering
    \begin{tabular}{@{}lL@{}}
    Pickands function & Parameters \tabularnewline
    \toprule
    $A_{\text{SL}}(\mathbf{w}) = \left( \sum_{k=1}^d w_k^{1/\alpha}\right)^{\alpha}$ & $\alpha \in (0,1]$ \tabularnewline
    \multirow{3}{*}{$A_\text{ASL}(\mathbf{w}) = \sum_{b \in \mathcal{P}_d}\bigg ( \sum_{i \in b} (\lambda_{i,b}w_i)^{1 / \alpha_b} \bigg)^{\alpha_b}$} &
    $\alpha_b \in (0, 1] $ \tabularnewline
    &  $\lambda_{i, b} \in [0, 1]$ \tabularnewline
    & $ \sum_{i \in b} \lambda_{i,b} =1$
    \end{tabular}
    \caption{Parametric Pickands functions for the symmetric $A_\text{SL}$ and asymmetric $A_\text{ASL}$ logistic copulas and their valid parameter ranges. $\mathcal{P}_d$ refers to the power set of $\{1,\ldots,d\}$. All functions are defined for domain $\mathbf{w} \in \Delta_{d-1}$.}
    \label{tab:sl}
\end{table}
\textbf{Synthetic data.} We consider two canonical families of extreme value distributions known as the symmetric logistic ($A_\text{SL}$) and the asymmetric logistic ($A_\text{ASL}$) families where the underlying Pickands function is given by \cite{segers_copulas} listed in Table~\ref{tab:sl}.
$\alpha \in (0, 1]$ is the parameter modeling the degree of dependence between variables ranging from complete dependence ($\alpha=0$) to complete independence ($\alpha=1$).
Exact sampling from distributions of this type are described in \cite{stephenson2003simulating}. 
Note that for both the symmetric and asymmetric copulas, the marginals are distributed according to the standard Fr\'echet distribution. 
We start by comparing the MSE of survival probabilities for $d=2$ where the true Pickands dependence function is given by the symmetric or asymmetric model described above for different degrees of dependence $\alpha$. 
We compute the exact values of the survival probability and consider survival probabilities associated with margins above the 75th percentile. 
As shown in Figures~\ref{fig:sl_survival} and~\ref{fig:asl_survival}, the proposed Pickands-$d$MNN estimator achieves the lowest MSE performance for most degrees of dependence $\alpha$ for the symmetric logistic model and all the degrees of the asymmetric logistic model. 
The proposed method performs worse comparatively in the full dependence case of the symmetric logistic (when all components of the vector are the same) which we suspect is due to difficulties in the optimization procedure of the $d$MNN.
We additionally showcase the ability of the proposed method to model high dimensional extreme value distributions. 
To do this, we train the Pickands-$d$MNN with data for $d=256$ with $\alpha \in \{0.25, 0.50, 0.75, 1.0\}$ and for $d = \{ 256, 512, 728, 1024\}$ with $\alpha = 0.5$. 
Then, we compute the MSE between the Pickands-$d$MNN and the true Pickands function via Monte Carlo with 10,000 uniformly sampled points in $\Delta_{d-1}$. 
The results are illustrated for varying $\alpha$ in Figures~\ref{fig:sl_mse_est_all_a} and~\ref{fig:asl_mse_est_all_a} and for $\alpha = 0.5$ in Figures~\ref{fig:sl_mse_est_all_d} and~\ref{fig:asl_mse_est_all_d}. 
While all hyperparameters were fixed at the beginning and not fine-tuned, we note that performance may improve if additional fine-tuning is performed using a validation set. 

\begin{table*}[tbh!]
\newcommand{\timesten}{\text{\tiny $\times10$}}
\centering
\begin{tabular}{@{}lclllll@{}} 
\toprule
 & $d$ & Train/Test Length & \textsc{Pickands} & \textsc{CFG} & \textsc{BDV} & \textsc{Proposed}  \\ 
\midrule
 Wind & 10 & day/week &
 ${4.48(18.6)}\timesten^{-4}$ & $\textit{4.15(15.1)}\timesten^{-4}$ &   $\bf 4.10(16.3)\timesten^{-4}$ &   $ 4.37(17.5)\timesten^{-4}$ \\
 Ozone & 4 & day/week &
 $3.06(4.66)\timesten^{-2}$ &  $2.99(4.56)\timesten^{-2}$ & $\textit{2.86(4.46)}\timesten^{-2}$ & $\bf 2.73(4.25)\timesten^{-2}$ \\
 Commodities & 10 & week/month &
 $4.34(5.82)\timesten^{-3}$  & $4.33(5.71)\timesten^{-3}$   & $\textit{1.60(1.96)}\timesten^{-3}$  &  $\bf 1.56(2.21)\timesten^{-3}$  \\
 S\&P 500 & 418 & week/month &
 $\textit{3.02(21.2)}\timesten^{-3}$ & $\textit{3.02(21.1)}\timesten^{-3}$ & $6.28(35.2)\timesten^{-3}$ & 
 $\bf 2.41(22.2)\timesten^{-3}$ \\
 Crypto & 100 & week/month &
 ${1.06(2.85)}\timesten^{-2}$ & $\textit{1.05(4.86)}\timesten^{-2}$ & ${1.34(3.44)}\timesten^{-2}$ & $\bf 8.57(26.4)\timesten^{-3}$ \\
\bottomrule
\end{tabular}
    \caption{MSE of different estimators in estimating maxima over longer time scales. Best and second best performances are marked in \textbf{bold} and \textit{italic} respectively.}
    \label{tab:real_data}
\end{table*}

\begin{figure}[h!]
    \centering
\begin{subfigure}{.24\textwidth}
  \centering
  \includegraphics[width=\linewidth]{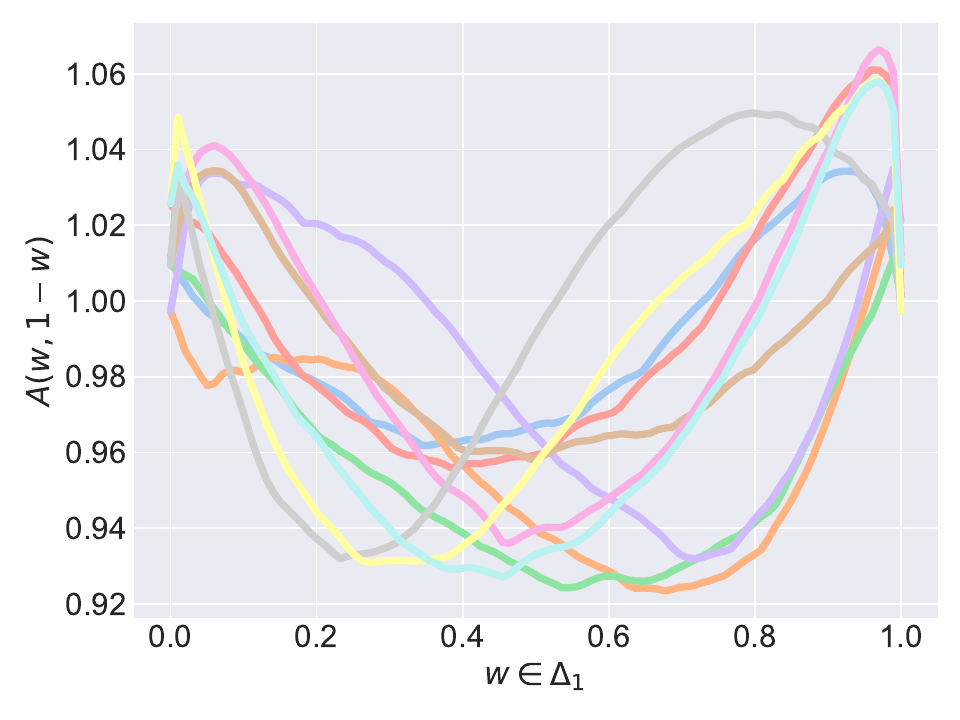}  
  \caption{Pickands 2d Margins}
  \label{fig:pick_marg}
\end{subfigure}
\begin{subfigure}{.24\textwidth}
  \centering
  \includegraphics[width=\linewidth]{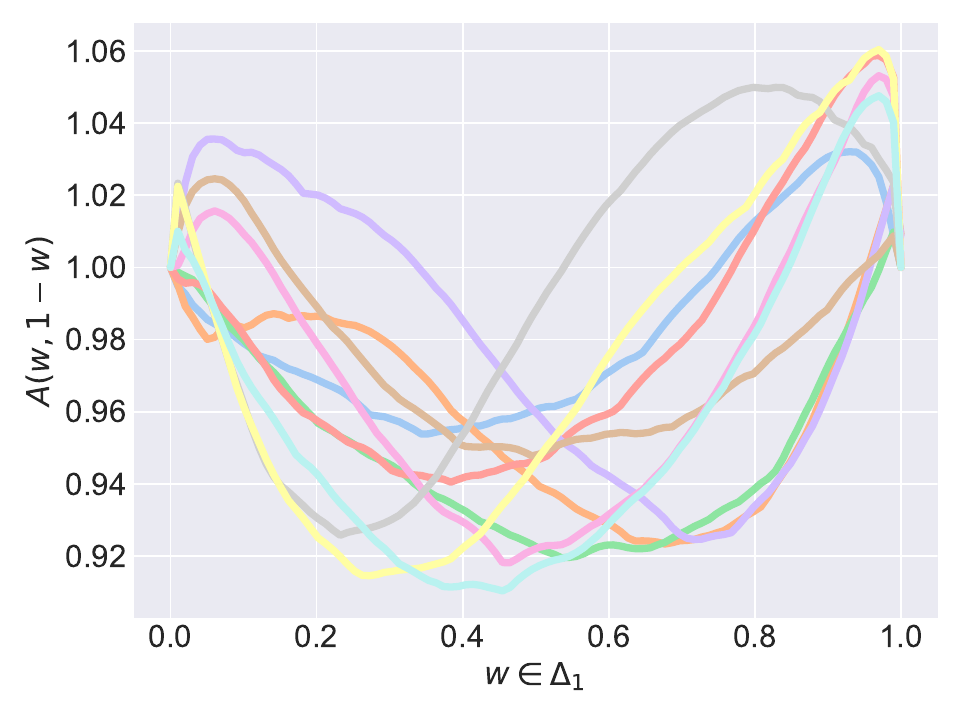}  
  \caption{CFG 2d Margins}
  \label{fig:cfg_marg}
\end{subfigure}
\begin{subfigure}{.24\textwidth}
  \centering
  \includegraphics[width=\linewidth]{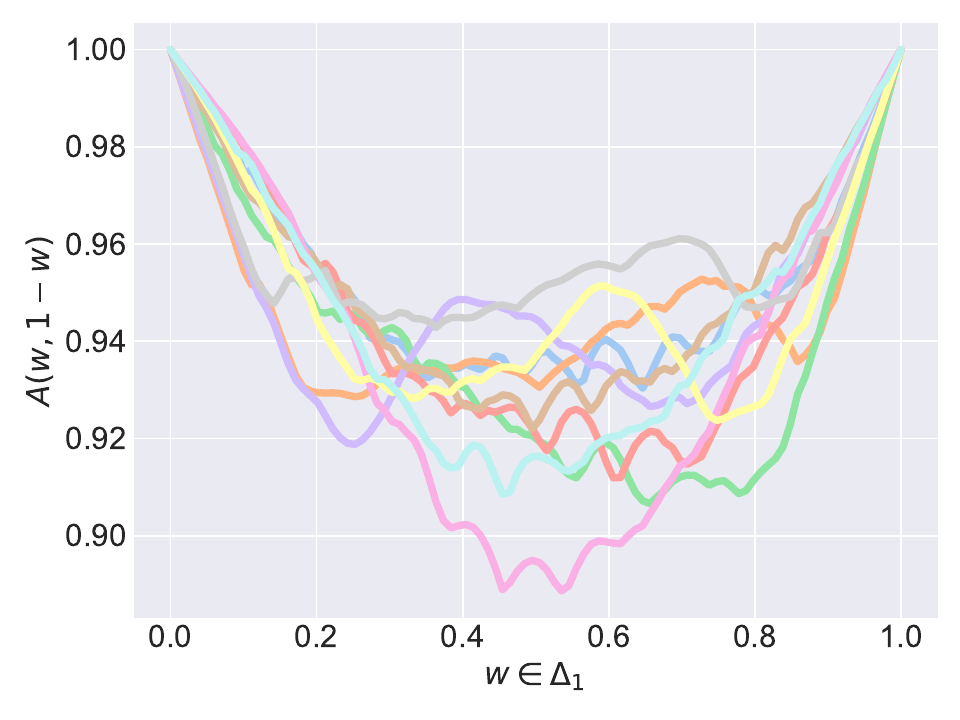}  
  \caption{BDV 2d Margins}
  \label{fig:bdv_marg}
\end{subfigure}
\begin{subfigure}{.24\textwidth}
  \centering
  \includegraphics[width=\linewidth]{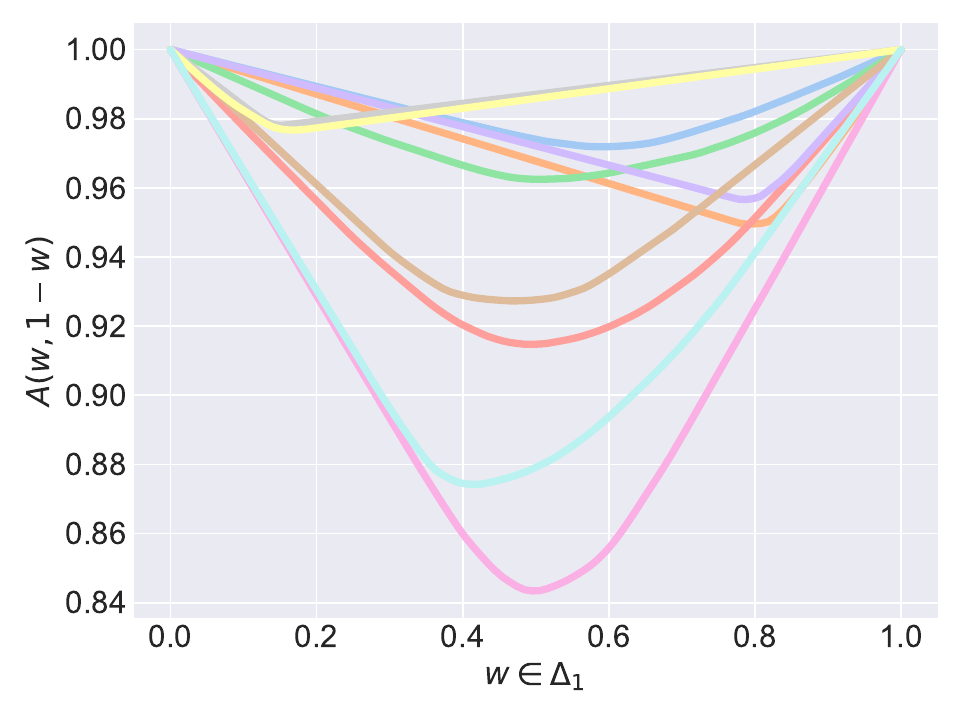}  
  \caption{$d$MNN 2d Margins}
  \label{fig:net_margins}
\end{subfigure}
\label{fig:marg}
\caption{Qualitative comparison of 10 out of 45 total 2d margins from learned 10d MEV for the California Winds dataset. The $d$MNN is the only method that retains margins that are valid Pickands dependence functions. See additional figures in Appendix~\ref{sec:large_figs} and~\ref{sec:more_experiments}. Figures~\ref{fig:net_marg_ozone}-~\ref{fig:net_marg_crypto}.}
\end{figure}
\textbf{Real data.}
We test the proposed estimator with real data on extreme ozone levels $(d=4)$, wind gusts $(d=10)$, commodity prices $(d=10)$, cryptocurrencies to USD conversion rates $(d=100)$ and S\&P 500 components with sufficient history $(d=418)$.
We provide details for each dataset in Appendix~\ref{sec:exp_details}.
For environmental datasets, we compute the maximum over the different sampling periods, while for the financial data we compute the maximum drawdown.
The maximum drawdown is defined as the difference between the minimum and maximum values over a time period normalized by the maximum value.
All margins were fitted with GEVs using the \verb|scipy| implementation.

The main challenge associated with real data is the lack of a ground truth for comparison purposes.
It is extremely difficult to accurately compare different estimators on real data because we can never observe the true distribution of extremes.
Since the purpose of EVT is to extrapolate to the tails from observations not necessarily in the tails, we consider extreme events on different time scales. 
If we fit based on extreme observations on shorter time scales and test on extreme observations on longer time scales, we will obtain an estimate of how well the different methods extrapolate to tail probabilities, since longer time scales will have more extreme events.  

We compute the accuracy of the different estimators with respect to the empirical estimate on held out data over longer time scales. 
Specifically, we choose a series of quantiles where we observe data and compute the difference between the estimated survival probabilities and the empirical estimate calculated from observed data. 
This is quantified as:
$
     \frac{1}{|Q|} \sum_{\mathbf{\gamma} \in Q}  \left[ \frac{1}{B} \sum_{b=1}^B \mathbbm{1}_{ \{ M_{n, b} \geq \mathbf{\gamma} \}} - P_\theta(M_n\geq \mathbf{\gamma}) \right]^2,
    \label{eqn:empirical_accuracy}
$
where $M_{n, b} = \left(M_{n, b}^{(1)}, \ldots, M_{n, b}^{(d)} \right)$ is the $d-$dimensional vector of point-wise maxima (or point-wise maximum drawdown over a period of interest),
$P_\theta$ is the estimated survival probability, and $Q$ is a set of thresholds to consider. 
\begin{figure}[tbh!]
    \centering
\begin{subfigure}{.24\textwidth}
  \centering
  \includegraphics[width=\linewidth]{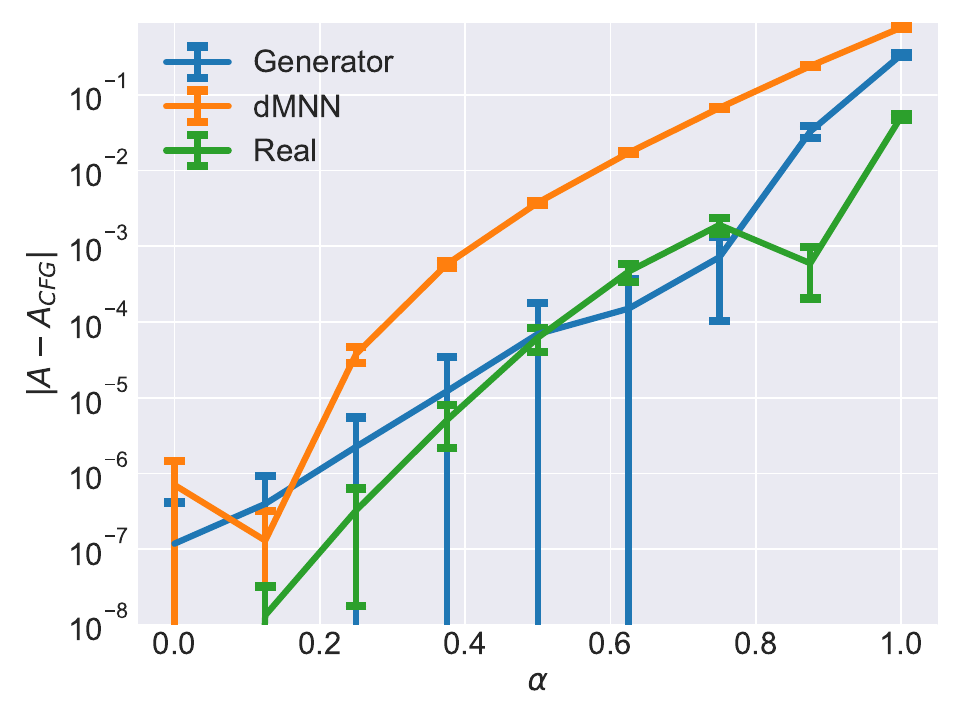}  

  \caption{SL CFG MSE $\Delta \alpha$}
  \label{fig:sl_mse_gen}
\end{subfigure}
\begin{subfigure}{.24\textwidth}
  \centering
    \includegraphics[width=\linewidth]{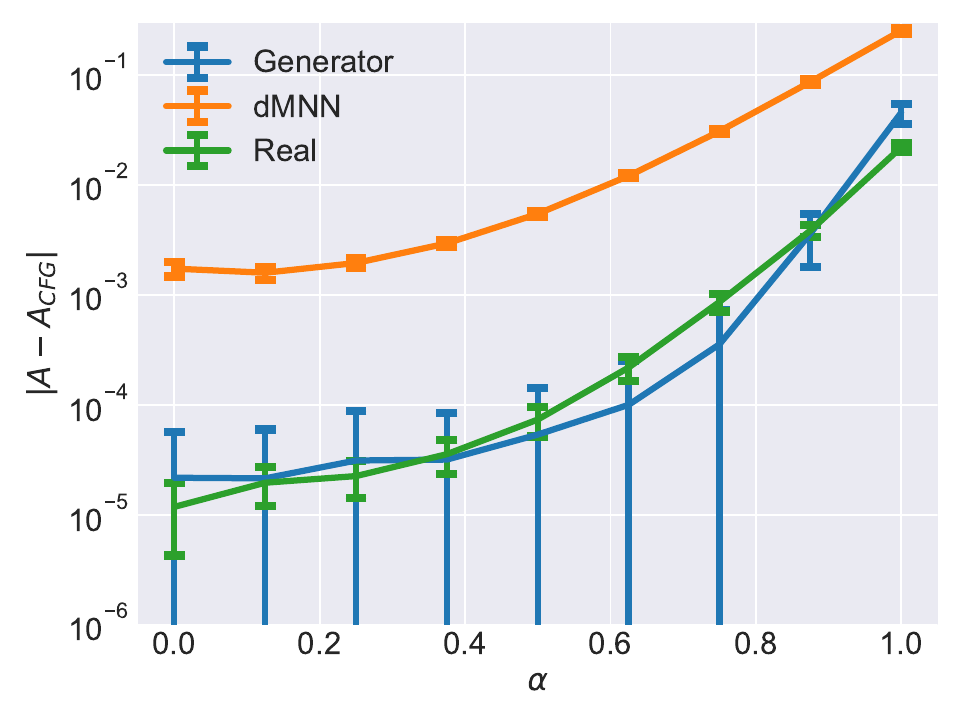}  
  \caption{ASL CFG MSE $\Delta \alpha$}
  \label{fig:asl_mse_gen}
\end{subfigure}
\begin{subfigure}{.24\textwidth}
  \centering
  \includegraphics[width=\linewidth]{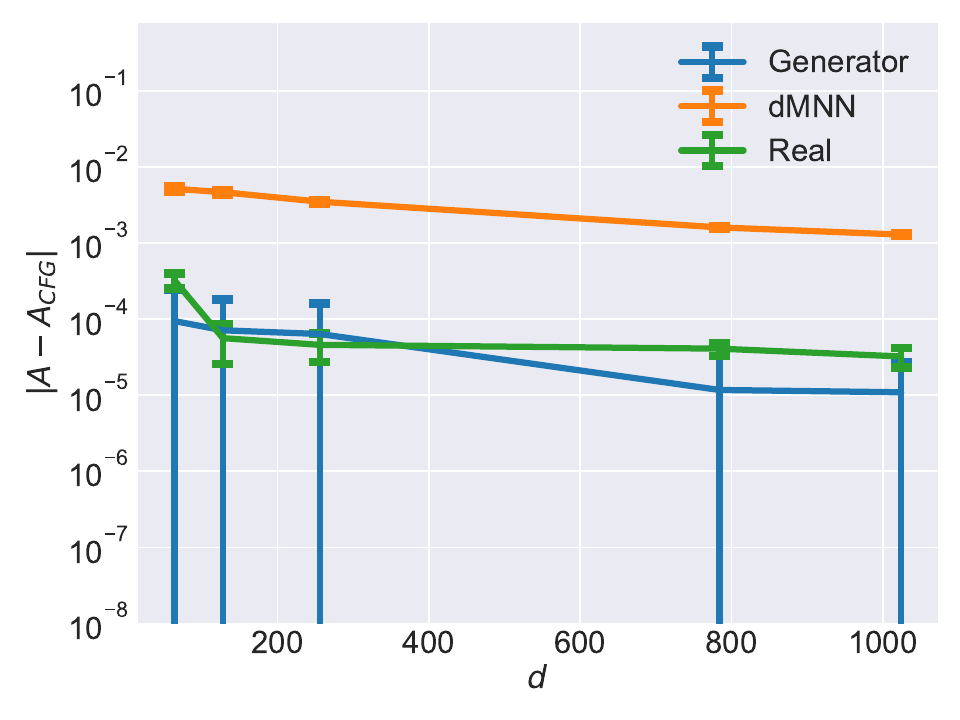}  
  \caption{SL CFG MSE $\Delta d$}
  \label{fig:sl_mse_gen_all_d}
\end{subfigure}
\begin{subfigure}{.24\textwidth}
  \centering
  \includegraphics[width=\linewidth]{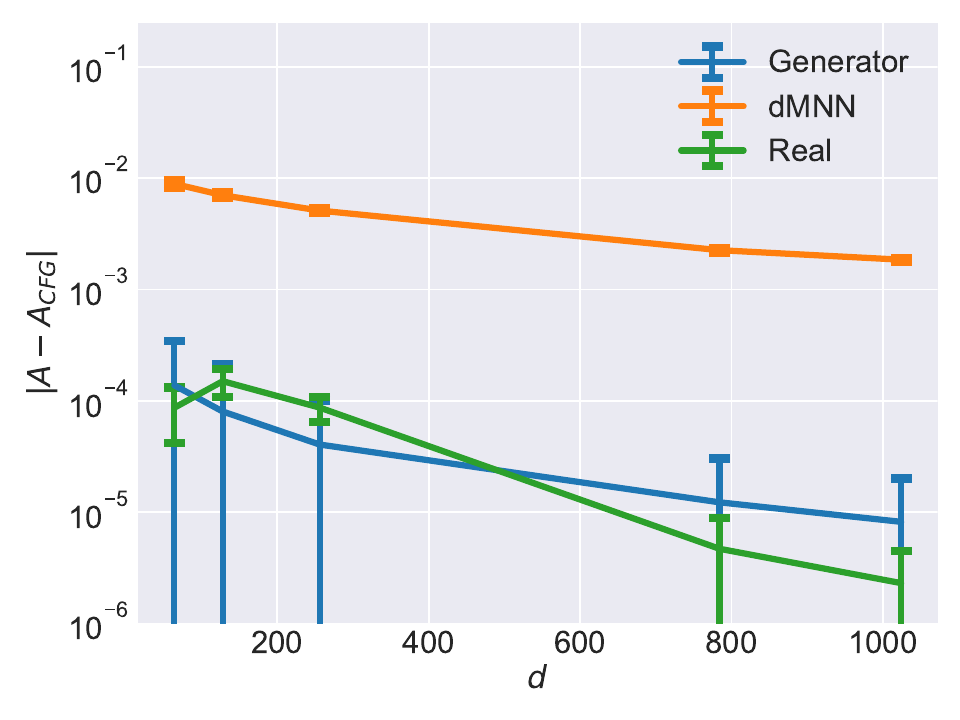}  
  \caption{ASL CFG MSE $\Delta d$}
  \label{fig:asl_mse_gen_all_d}
\end{subfigure}
\caption{MSE of CFG estimate for 1000 samples and 1000 simplex points for $d=225$ at various $\alpha \in (0,1)$ (\ref{fig:sl_mse_gen}, \ref{fig:asl_mse_gen}) and for $\alpha=0.5$ at various $d=\{64, 128, 256, 784, 1024\}$ (\ref{fig:sl_mse_gen_all_d}, \ref{fig:asl_mse_gen_all_d}). Data sampled from generative model (blue), $d$MNN (orange), and ground truth (green), where the distributions considered were $A_\text{SL}$ (\ref{fig:sl_mse_gen}) and  $A_\text{ASL}$ (\ref{fig:asl_mse_gen}). Both models were trained with 1000 data points.}
\end{figure}
We choose $Q$ to be all quantiles such that the empirical probability is greater than 0. 
This measures how well the proposed method can extrapolate to greater extremes over longer time scales.
The results are presented in Table~\ref{tab:real_data} and suggest that while most estimators perform similarly, the proposed method most consistently performs the best in terms of the evaluation metric.
We would like to emphasize that empirical evaluation on real data is very challenging, and the high variances prevent us from making meaningful statements on the efficacy of any of the methods.
However, from Figures~\ref{fig:net_marg3} and~\ref{fig:net_margins}, we see that our proposed estimator is the only one that satisfies the necessary properties of the Pickands function, which is the main purpose of the proposed method.
Additional figures in Appendix~\ref{sec:large_figs} showcase this property on additional datasets and Appendix~\ref{sec:more_experiments} Figures~\ref{fig:net_wind} to \ref{fig:net_crypto} compares these for different architectures. 
It is critical that these properties are satisfied so that downstream tasks such as conditional probabilities can be computed.
From the state of the art estimators, the properties are not satisfied and thus the applicability of the estimators is severely limited.

\textbf{Sampling from the copula.}
Finally, to determine the efficacy of sampling from an arbitrary Pickands copula, we consider two synthetic examples using the previously described MEV distributions in Table~\ref{tab:sl}. 
In this experiment, we train the generator $G(.; \bm \phi)$ in \eqref{eqn:segers_opt} based on 1000 samples from the target distribution. 
We represent the $G(\cdot;\bm \phi)$ as a 2 layer 256 width multi-layer perceptron with $\mathrm{ReLU}$ activation functions and set $\eta = 1$.
Since the Pickands function completely determines the dependency of the random variables, we compare the CFG estimate of the Pickands function from generated samples to the true Pickands function as a measure of sampling quality.
We use the CFG estimator due to its ubiquity in the literature and its highly regarded status as a standard estimator for the Pickands dependence function.
The results for generating 225 dimensional samples with varying dependence $\alpha \in [0, 1]$ are shown in Figures~\ref{fig:sl_mse_gen} and~\ref{fig:asl_mse_gen}.
The figures suggest that the generative model performs comparatively well for both distributions considered, with the worst performance occurring in the nearly independent cases ($\alpha = 1$). 
This is expected, since independence implies a spectral measure with delta functions on the corners of the simplex, which is difficult to learn (see the bottom row of Figure~\ref{fig:illustration} as an example).
The figures additionally suggest that sampling using the learned weights of the $d$MNN has lower variance (since the spectral measure in this is a finite discrete approximation) but does not perform as well in sampling as the generative model.
The error of the CFG estimate for the proposed sampling methods (blue and orange) and the exact sampling (green) follow very similar trends in errors, suggesting that both sampling methods are recovering the true spectral measure.
\section{Concluding Remarks}\label{sec:conclusion}
We introduced a new neural network architecture for modeling MEV distributions while enforcing all the properties of the distribution.
We additionally show that the architecture can approximate any Pickands function, which allows for precise representations of MEV distributions. 
Finally, we present a generative model for recovering the spectral representation. 
Numerical results are provided to empirically demonstrate the effectiveness of the methods in their respective tasks.
However, there are some limitations of the proposed methods.

\textbf{Limitations of Pickands-$d$MNNs and Generative Model.}
The main challenge associated with modeling using $d$MNNs are optimization and architectural choices. 
Choosing appropriate hyperparameters is a difficult and opaque task that requires additional care.
This is a case where non-parametric methods are advantageous, at the cost of being unable to guarantee the necessary properties of the function. 
Additional progress on understanding the training deep neural networks should improve the representational capabilities of the $d$MNNs, given its theoretical potential to approximate any Pickands functions to arbitrary precision. 
Optimization of the generative model suffers from the same issues.

\textbf{Future Work.}
The proposed methods have possible applications in a variety of modeling situations. 
One possibility is to use the $d$MNN to compute conditional probabilities, which is useful in a variety of classification tasks.
Another is in using the spectral measure for finding groups of variables that are extreme simultaneously, such as in \citep{engelke2021sparse}. 
Finally, applications of extremes are important in understanding robustness properties of neural networks \citep{weng2018evaluating}, and the proposed work provides foundation for high dimensional extensions. 

\bibliography{refs}
\section*{Acknowledgements}
Material in this paper is based upon work supported by the Air Force Office of Scientific Research under award number FA9550-20-1-0397.
AH was supported by NSF Graduate Research Fellowship.

\bibliographystyle{plainnat}

\onecolumn
\title{Appendix}
\maketitle
\appendix

\section{Fully $d$-max-decreasing Functions}
\label{sec:fdmd}
We use the definition given in \cite{hofmann2009characterization}.
A function $A(\mathbf{w}) : \mathbb{R}^d \to \mathbb{R}$ is fully d-max-decreasing if and only if for any $\mathbf{x} \leq \mathbf{y} \leq 0$ and any subset $E \subsetneq \{1, \ldots, d\}$:
$$
\sum_{\substack{\mathbf{m} = \{0,1\}^d,\\ m_j = 1 \text{ if } j \in E}} \left [(-1)^{d + 1 -\sum_{j \leq d} m_j} \left ( \sum_{j \leq d} -y_j^{m_j} x_j^{1-m_j}\right) A \left (\frac{y_1^{m_1} x_1^{1-m_1}}{\sum_{j \leq d} y_j^{m_j} x_j^{1-m_j}}, \cdots, \frac{y_d^{m_d} x_d^{1-m_d}}{\sum_{j \leq d} y_j^{m_j} x_j^{1-m_j}} \right)\right ] \geq 0, 
$$
and $A(\mathbf{e}_i) = 1$ where $\mathbf{e}_i$ is the canonical basis function. 
Moreover, from \citet[Theorem 3.1.1]{hofmann2009characterization}, the following three characterizations are equivalent:
\begin{enumerate}
    \item The function $$\exp\left(-\sum_{j=1,\ldots,d}x_j A\left(\frac{x_1}{\sum_{j=1,\ldots,d}x_j}, \cdots, \frac{x_d}{\sum_{j=1,\ldots,d}x_j}\right)\right)$$ defines a multivariate extreme value distribution; 
    \item There exists a spectral measure $\Lambda$ such that $$A(\mathbf{w}) = \int_{\Delta_{d-1}} \max_{k=1,\ldots,d} w_k s_i \, \mathrm{d} \Lambda(\mathbf{s}), \quad \mathbf{w} \in \Delta_{d-1};$$
    \item $A(\mathbf{w})$ is fully $d$-max-decreasing.
\end{enumerate}
Next, we note the nesting property of $d$-max-decreasing functions, given in \citet[Section 2.2]{hofert2018hierarchical}, that hierarchies of spectral measures define valid EVDs, i.e.
$$
A(\mathbf{w}) = \mathbb{E}\left [ s^{(2,1)}_1 \mathbb{E}\left[ \max_{k=1,\ldots,d} s^{(1,1)}_k w_k \right], \ldots,  s^{(2,1)}_d\mathbb{E}\left[ \max_{k=1,\ldots,d} s^{(1,d)}_k w_k \right] \right].
$$
We use part of this property in the next section to define the $d$-max neural network. 

\section{Proofs}
\label{sec:proofs}
\subsection{$d$-Max-Decreasing Neural Networks}
\label{sec:proof_arch}

We partition the proof into the 1-layer case and the $n$-layer case. 
We assume that all parameters $\theta \in [0,1]$ for the purposes of the proof. 

{\bf Background: $D-$norms.}
$D-$norms are norms defined as 
\begin{equation}
\| \mathbf{x} \|_D := d\; \mathbb{E}_{\bm{\theta}}\left[ \max_{k=1\ldots d}(|x_k| \theta_k)\right], \quad \mathbf{x} \in \mathcal{X} \subset \mathbb{R}^d,
\label{eq:dnorm}
\end{equation}
where $\bm{\theta} \in [0, 1]^{d}$ and $\mathbb{E}[\theta_k] = 1/d$, for $i=1\ldots d$. 
We note that the condition $\mathbb{E}[\theta_k] = 1/d$ is not necessary for the $d-\max$ decreasing property, though it leads to unit exponential margins for convenience during inference, see \citet[Section 2.1]{fougeres2013dense} or \citet[Definition 5.4.1]{hofmann2009characterization} for more on this property.
The key condition is that the expectation in \eqref{eq:dnorm} is taken with respect to the distribution of $\bm{\theta}$ which has support only on nonnegative real numbers. 
Taking $\mathcal{X}$ to be the unit simplex, we see that a $D-$norm defines a Pickands dependence function, and by the spectral representation of the Pickands function, all Pickands functions are $D-$norms.
The main property we will use throughout the proof is that compositions of $D-$norms are also $D-$norms. This property is well established in, for example, \citet{hofert2018hierarchical} and \citet{hofmann2009characterization}.
We finally note that all $D-$norms satisfy the fully $d-$max decreasing property defined as shown in~\citet[Theorem 3.1.1]{hofmann2009characterization} .

{\bf 1-Layer Case.}
\begin{proof}
Recall that the 1-layer $d$MNN is given by
\begin{align}
\label{eq:A1}
A^{(1)}_\theta(\mathbf{w}) &= \max\left(L^{(1)}(\mathbf{w}) + (1-L^{(1)}(\mathbf{e})^T\mathbf{w}), \max_{k=1\ldots d} w_k\right), \quad \mathbf{w} \in \Delta_{d-1}  \\
L^{(1)}(\mathbf{w}) &= \frac1{n_1}\sum_{j=1}^{n_1}\left(\max_{k=1\ldots d} w_k\theta_{kj} \right)_j
\label{eq:L1}
\end{align}
The expression \eqref{eq:L1}, corresponding to the first term in \eqref{eq:A1}, is a valid $D-$norm since it is the expectation with respect to a nonnegative spectral measure. The second term $(1-L^{(1)}(\mathbf{e})^T \mathbf{w}) = \sum_{i=1}^d (1-\mathbb{E}[\theta_i]) w_i$ is also a $D-$norm since $(1-\mathbb{E}[\theta_i])$ is positive, for $i=1\ldots d$, and thus it is also an expectation with respect to a nonnegative spectral measure. In fact, it is equivalent to $\|\text{diag} (1-\mathbb{E}[{\bm\theta}]) \mathbf{w}\|_1$. The combination $L^{(1)}(\mathbf{w}) + (1 - L^{(1)}(\mathbf{e})^T \mathbf{w})$ is then the sum of two $D-$norms, equivalent to a composition with the $\| \, \cdot \, \|_1$ norm, which is again a $D-$norm. 
Finally, the outer $\max$ with $\max_{i=1\dots d} w_i$, a $D-$norm corresponding to dependence, is yet another composition of $D-$norms. This results in a function that is fully $d-\max$ decreasing and concludes the proof for the single layer case. 
\end{proof}
{\bf $n$-Layer Case.} 
\begin{proof}
We first show the base case (the 2-layer case), then show that the general $n$-layer case follows.
We focus on the composition of intermediate layers, since the technique for proving the output layer is a $D-$norm follows from the $1$-layer case. 
Recall that the 2-layer $d$MNN is given by
\begin{align*}
    A^{(2)}_\theta(\mathbf{w}) &= \max\left(L^{(2)}(\mathbf{w}) +(1 - L^{(2)}(\mathbf{e})^T \mathbf{w}), \max_{k=1\ldots d} w_k\right), \quad \mathbf{w} \in \Delta_{d-1} \\
    L^{(2)}(\mathbf{w}) &= \frac1{n_2} \sum_{j=1}^{n_2}\left( \ell^{(2)}\left(\ell^{(1)}(\mathbf{w})\right)\right)_j
\end{align*}
Let $\ell^{(1)}(\mathbf{w})$ have width $n_1$. 
Then the output of $\ell^{(1)}$ is given by the following vector
\begin{equation}
\label{eq:first_layer}
\ell^{(1)}(\mathbf{w}) = \begin{pmatrix}
\mathbb{E}_{\bm{\theta}^{(1,1)} \sim \lambda^{(1,1)} } \left [ \max_{k=1\ldots d} \theta_k^{(1, 1)} w_k \right ] \\
\vdots \\
\mathbb{E}_{\bm{\theta}^{(1, n_1)} \sim \lambda^{(1,n_1)}} \left [ \max_{k=1\ldots d} \theta_k^{(1, n_1)} w_k \right ] 
\end{pmatrix}, \quad {\bm \theta} \in [0,1]^{d}.
\end{equation}
Each row in~\eqref{eq:first_layer} is a $D-$norm where the expectation is taken over a delta function centered at $\bm \theta$, i.e. $\lambda^{(1,j)}=\delta(\bm{\theta}^{(1,j)})$ for $j=1,\ldots,n_1$. 
Therefore, the property of $D-$norms is preserved for each row of~\eqref{eq:first_layer}. By analogy to $\ell^{(1)}$ in~\eqref{eq:first_layer}, the property of $D-$norms is preserved for $\ell^{(2)}$ in~\eqref{eq:second_layer}:
\begin{equation}
\label{eq:second_layer}
\ell^{(2)}(\mathbf{w}) = \begin{pmatrix}
\mathbb{E}_{\bm{\theta}^{(2,1)} \sim \lambda^{(2,1)} } \left [ \max_{k=1\ldots n_1} \theta_k^{(2, 1)} w_k \right ] \\
\vdots \\
\mathbb{E}_{\bm{\theta}^{(2, n_2)} \sim \lambda^{(2,n_2)}} \left [ \max_{k=1\ldots n_1} \theta_k^{(2, n_2)} w_k \right ] 
\end{pmatrix}, \quad {\bm \theta} \in [0,1]^{d}.
\end{equation}

We then use the nesting property of $D-$norms given in \citet{hofert2018hierarchical} such that $\ell^{(2)}(\ell^{(1)}(\bm{w}))$ is a $D-$norm, and by the same construction, $\ell^{(n)}(\ell^{(n-1)}(\cdots(\ell^{(1)}(\bm{w}))))$ is a $D-$norm and is thus fully $d-\max$-decreasing. Following the arguments in the $1$-layer case for the output layer then completes the proof.
\end{proof}

\subsection{Universal Approximation}
\label{sec:proof_pointwise}

Our proof that our architecture is an universal approximator of Pickand's copula functions is constructive.
Recall that every Pickands function has the form
\begin{equation}
\label{eq:pick_spec}
 A(\mathbf{w}) = \mathbb{E}_{\mathbf{s} \sim \lambda(\Delta_{d-1})}\left[\max_{k = 1 \ldots d} s_kw_k\right], \quad \mathbf{w} \in \Delta_{d-1},   
\end{equation} where $\lambda$ is a spectral measure with $\text{supp}(\lambda) = \Delta_{d-1}$. We now construct a single layer dMNN, with width $n$, by sampling $n$ independent and identically distributed (i.i.d.) samples $\mathbf{s}^{(1)}, \ldots, \mathbf{s}^{(n)} \sim \lambda(\Delta_{d-1})$, and setting
\begin{equation}
\tilde A_n (\mathbf{w}) = \frac1n \sum_{j=1}^n \max_{k=1\ldots d}(s_k^{(j)} w_k)
\end{equation}
Before showing that $\tilde A_n$ converges uniformly to $A$, we show that it converges point-wise. Although this intermediary result is not needed to show uniform converge, its proof provides intuition while being less technical.

{\it
The copula $\tilde A_n$ converges pointwise to $A$, almost surely.
}

\begin{proof}

Consider the discrete distribution $\mathbb{\Lambda}_n$ given by the $n$ i.i.d. samples $\mathbf{s}^{(1)}, \ldots, \mathbf{s}^{(n)} \sim \lambda(\Delta_{d-1})$:
$$
\mathbb{\Lambda}_n := \frac1n \sum_{i=1}^n \delta(\mathbf{s}^{(i)}),
$$
where $\delta(\mathbf{s})$ represents a Dirac measure at $\mathbf{s}$.
By the law of large numbers, for every $\mathbf{w} \in \Delta_{d-1}$,
\begin{align*}
\mathbb{E}_{\mathbf{s} \sim \mathbb{\Lambda}_n}\left[\max_{k=1\ldots d}s_kw_k\right] & \xrightarrow{a.s.} \mathbb{E}_{\mathbf{s} \sim \lambda}\left[\max_{k=1\ldots d}s_kw_k\right], \quad n \to \infty \\
\implies A_{\bm\theta}(\mathbf{w}) & \xrightarrow{a.s.} A(\mathbf{w}).\qedhere
\end{align*}
\end{proof}


We now state and prove the main result regarding uniform convergence. 

{\it
The empirical process $$\mathbb{G}_n = \sqrt{n} \left( \tilde A_n - A \right )$$ weakly converges to a zero-mean Gaussian process as $n \to \infty$ where $\tilde A_n$ is a single layer dMNN with width $n$.
}
\begin{proof}
Let $\lambda$ be the law given by the spectral measure $\lambda(\Delta_{d-1})$ and the discrete empirical spectral measure be given by $\mathbb{\Lambda}_n :=\frac1n \sum_{j=1}^n \delta(\mathbf{s}^{(j)})$ for $n$ i.i.d. samples $\mathbf{s}^{(1)},\ldots, \mathbf{s}^{(n)}$ from $\lambda$. 
We additionally write $\lambda f := \mathbb{E}_{\mathbf{s} \sim \lambda}[f(\mathbf{s})]$ as the expectation with respect to the measure $\lambda$.
The empirical process $\mathbb{G}_n$ is defined by
\begin{align*}
\mathbb{G}_n &= \sqrt{n} \left ( \tilde A_n - A \right ) \\
&= \sqrt{n} \left( \mathbb{\Lambda}_n - \lambda \right)f \\
&= \sqrt{n} \left ( \frac1n \sum_{j=1}^n \max_{k=1\ldots d}(s_k^{(j)} w_k) - \mathbb{E}_{\mathbf{s} \sim \lambda}\left[\max_{k=1\ldots d}(s_kw_k)\right] \right ),
\end{align*}
where $f \in \mathcal{F}$ and
$$
\mathcal{F} := \{ f_w(s) := \max_{k = 1 \ldots d} ( s_k w_k ) : \mathbf{w} \in \Delta_{d-1} \}.
$$
By the classical central limit theorem, for a given $\mathbf{w}$, $\sqrt{n} \left (\tilde A_n(\mathbf{w})  - A(\mathbf{w})\right)\xrightarrow{d} \mathcal{N}(0, \sigma^2)$, with $\sigma^2 \leq (1 - A(\mathbf{w}))(A(\mathbf{w}) - \frac1{d^2})$, since the random variable $\max_{k= 1\ldots d}(w_k s_k) \in [1/d^2, 1]$ is bounded and has finite variance.

Our claim is that $\mathbb{G}_n \leadsto \mathbb{G}$ where $\mathbb{G}$ is a zero-mean Gaussian process for establishing uniform convergence over $\mathbf{w}$.  
We will now show that the function class given by $\mathcal{F}$ is $\lambda-$Donsker. 
To show this, we will show that the bracketing integral given by
\begin{equation}
\label{eq:brack_int}
\mathcal{J}_{[\,]}(1,\mathcal{F}, L_2(\lambda)) = \int_0^1 \sqrt{\log N_{[\,]}(\epsilon, \mathcal{F} \cup 0, L_2(\lambda))} \, \mathrm{d} \epsilon 
\end{equation}
converges where the $L_2(\lambda)$ norm is defined as $\|f\|_{\lambda, 2} = (\int f^2 d \lambda)^{1/2}$.
A sufficient condition for convergence of \eqref{eq:brack_int} is to show that the logarithm of the bracketing number $N_{[\,]}$ grows at a rate slower than $O(\frac{1}{\epsilon^2})$.
The function class $\mathcal{F}$ is indexed by $\mathbf{w} \in \Delta_{d-1}$ and is Lipschitz on $\mathbf{w}$.
From \citet[Lemma 2.14]{sen2018gentle}, the bracketing number of $\mathcal{F}$ is thus bounded above by the covering number of $\Delta_{d-1}$, i.e.
$$
N_{[\,]}(2\epsilon, \mathcal{F}, L_2(\lambda)) \leq N(\epsilon, \Delta_{d-1}, \| \cdot \|_2),
$$
where the covering number of the unit simplex is asymptotically $O\left(\frac{1}{\epsilon^{d-1}}\right)$. The logarithm of the bracketing number then grows at a rate $\log N_{[\,]}\leq O\left ( \left (d - 1 \right ) \log \frac1\epsilon \right) < O(\frac1{\epsilon^2})$. This proves that $\mathcal{F}$ is $\lambda-$Donsker and thus $\mathbb{G}_n \leadsto \mathbb{G}$.
\end{proof}

\section{Survival Probability Estimation}
\label{sec:survival}

One particularly useful task is estimating multi-dimensional survival probabilities rather than cumulative probabilities.
More precisely, let $\left(\gamma_1, \cdots, \gamma_d \right) \in \mathbb{R}^d$ be a $d-$dimensional vector of thresholds, we are interested in calculating the following survival probability:
\begin{align}
\label{survival_general}
   \mathbb{P} \left[M_{n}^{(1)} > \gamma_1, \cdots, M_{n}^{(d)} > \gamma_d \right] 
    = \mathbb{P} \left[\bar{M}_{n}^{(1)} > \bar{\gamma}_1 , \cdots, \bar{M}_{n}^{(d)} > \bar{\gamma}_d \right],
\end{align}
where $\bar{\gamma}_k = \frac{\gamma_k - b_n^{(k)}}{a_n^{(k)}}$, $k \in \{1, \cdots, d\}$.

To calculate this, we simply use a change-of-variable technique which we present in the following proposition. 
This approach is well known, and we only provide the proposition for completeness. 
\begin{proposition}[Survival Probability Computation]
Let $G_k(x):= F_k^{-1}(1-F_k(x))$ for $k~\in~\{1,\dots,d\}$, then the random variables $G_k(\bar{M}_{n}^{(k)})$ and $\bar{M}_{n}^{(k)}$ have the same marginal CDF $F_k$, for $k\in\{1,\dots,d\}$, and
\begin{align}
    \mathbb{P} \left[\bar{M}_{n}^{(1)} > \bar{\gamma}_1 , \cdots, \bar{M}_{n}^{(d)} > \bar{\gamma}_d \right] = \mathbb{P} \left[G_1(\bar{M}_{n}^{(1)}) < G_1(\bar{\gamma}_1), \cdots, G_d(\bar{M}_{n}^{(d)}) < G_d(\bar{\gamma}_d) \right].
    \label{survivalchgvareq}
\end{align}
\end{proposition}
\begin{proof}
With the change-of-variable $G_k(x):= F_k^{-1}(1-F_k(x))$ for $k \in \{1, \cdots, d\}$, it first follows that the random variables $G_k(\bar{M}_{n}^{(k)}) \sim F_k$: 
\begin{equation}
    \mathbb{P}(G_k(\bar{M}_{n}^{(k)})\leq x) = \mathbb{P}(F_k^{-1}(1-F_k(\bar{M}_{n}^{(k)}))\leq x)=\mathbb{P}(1-F_k(\bar{M}_{n}^{(k)})\leq F_k(x))=F_k(x),
\end{equation}
since $F_k(\bar{M}_{n}^{(k)})$ and $1-F_k(\bar{M}_{n}^{(k)})$ follow the unit uniform distribution.

Moreover, the survival probability can be written as:
\begin{align*}
     \mathbb{P} \left[\bar{M}_{n}^{(1)} > \bar{\gamma}_1 , \cdots, \bar{M}_{n}^{(d)} > \bar{\gamma}_d \right]
  & = \mathbb{P}\left[1- F_1(\bar{M}_{n}^{(1)}) < 1-F_1(\bar{\gamma}_1) , \cdots, 1-F_d(\bar{M}_{n}^{(d)}) < 1 - F_d(\bar{\gamma}_d) \right] \\ 
  & = \mathbb{P}\left[G_1(\bar{M}_{n}^{(1)}) < G_1(\bar{\gamma}_1) , \cdots, G_d(\bar{M}_{n}^{(d)}) < G_d(\bar{\gamma}_d) \right] \\
  & = C \left( 1 - F_1(\bar{\gamma}_1), \cdots, 1 - F_d(\bar{\gamma}_d) \right),
\end{align*}
where $C$ is the copula of $\left(G_1(\bar{M}_{n}^{(1)}), \cdots, G_d(\bar{M}_{n}^{(d)}) \right)$.
\end{proof}
This proposition implies that the transformed variables $G_k(\bar{M}_{n}^{(k)})$ are samples from extreme value distributions. 
Then, we can fit Pickands dependence function to these transformed variables, and finally evaluate the corresponding extreme value copula on
$\left(1-F_1(\bar{\gamma}_1), \cdots, 1-F_d(\bar{\gamma}_d)\right)$. %
Details on how to estimate the survival probability in \eqref{survival_general} are given in Algorithm 2.


\section{Additional Experiments and Figures}
\label{sec:more_experiments}

\subsection{24 Width 3 Depth Architecture}

Here we repeat the experiments with a different architecture. 
All other hyperparameters are the same, the only difference is we increase the depth to 3 and use a width of 24 for each layer.
Most of the results remain similar for the synthetic data but we see a change in the results for the real data, specifically, the Wind and Commodities data show a deterioration in performance. 
However, the variances are still high for the real experiments and not much can be said regarding the efficacy of any single method. 

\begin{figure}[tbh!]
    \centering
\begin{subfigure}{.48\textwidth}
  \centering
  \includegraphics[width=\linewidth]{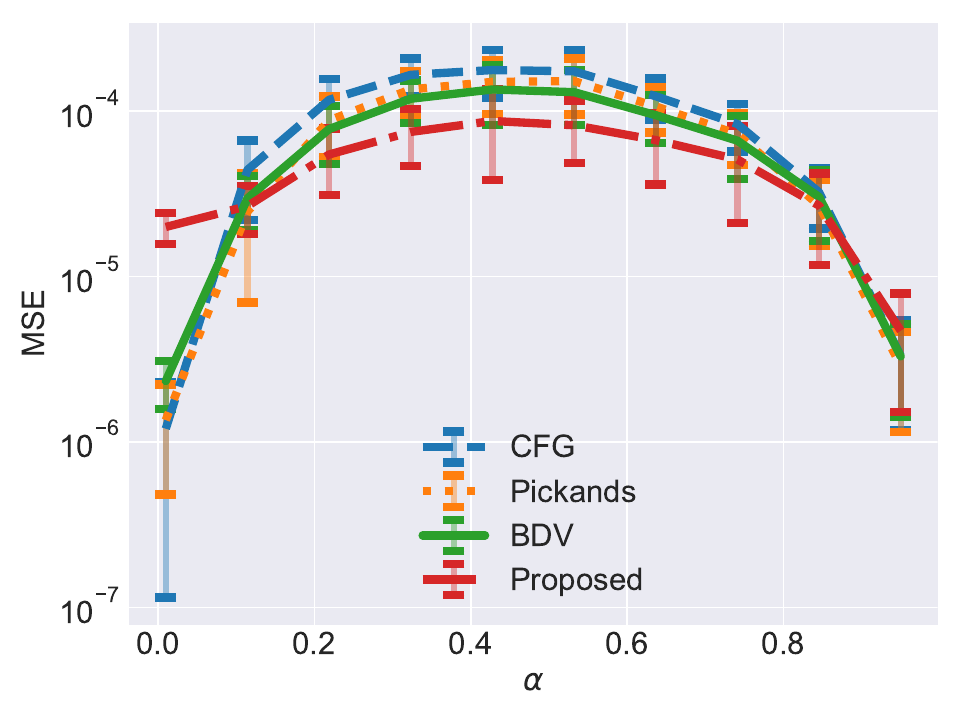}
  \caption{$A_\text{SL}$ MSE ($d=2$)}
  \label{fig:sl_survival32}
\end{subfigure}
\begin{subfigure}{.48\textwidth}
  \centering
  \includegraphics[width=\linewidth]{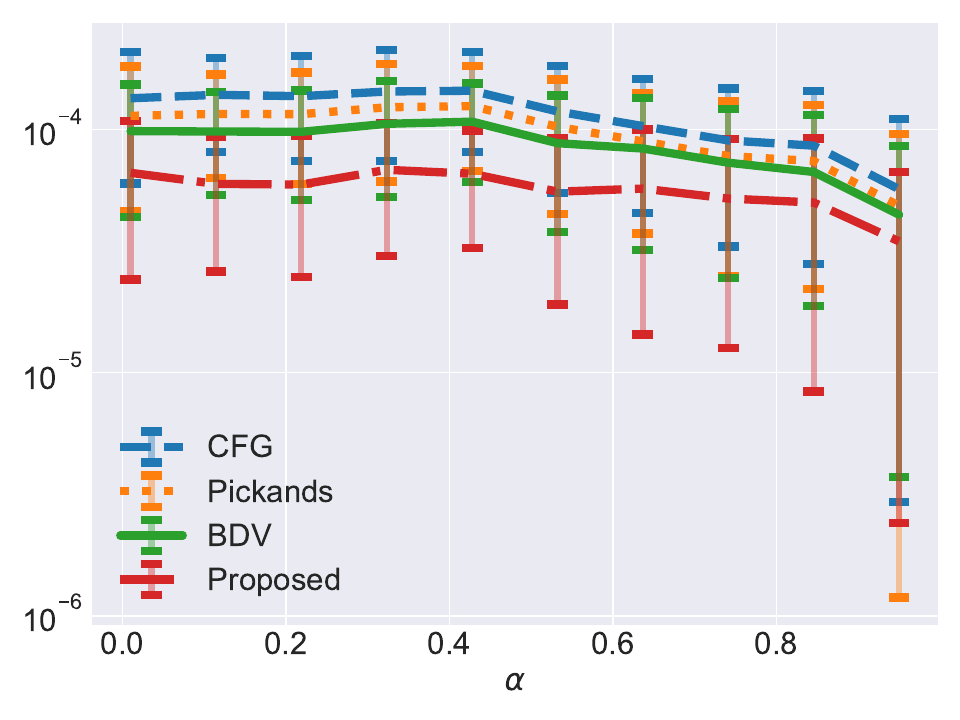}
  \caption{$A_\text{ASL}$ MSE ($d=2$)}
  \label{fig:asl_survival32}
\end{subfigure} 
\caption{Using 24 width 3 depth architecture: MSE of survival probabilities for $d=2$ with $100$ samples for $A_\text{SL}$ (\ref{fig:sl_survival32}) and $A_\text{ASL}$ (\ref{fig:asl_survival32}). Thresholds are above the $75$th percentile.}
\end{figure}

\begin{figure*}[h!]
    \centering
\begin{subfigure}{.48\textwidth}
  \centering
  \includegraphics[width=\linewidth]{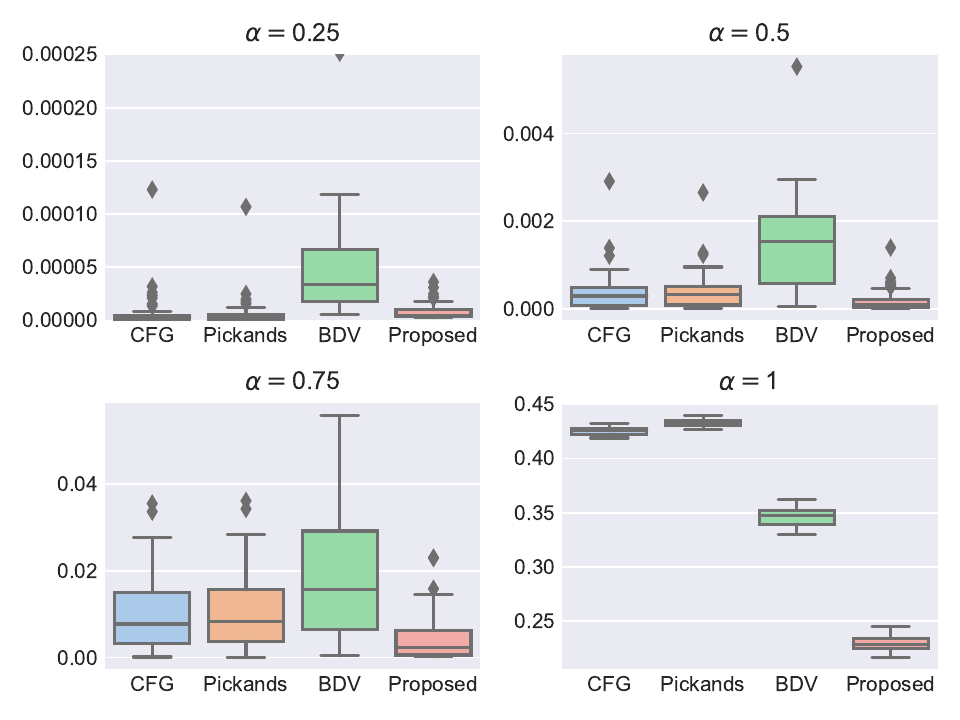} 
  \caption{$A_\text{SL}$ MSE ($d=256$)}
  \label{fig:sl_mse_est_all_a_a2}
\end{subfigure}
\begin{subfigure}{.48\textwidth}
  \centering
  \includegraphics[width=\linewidth]{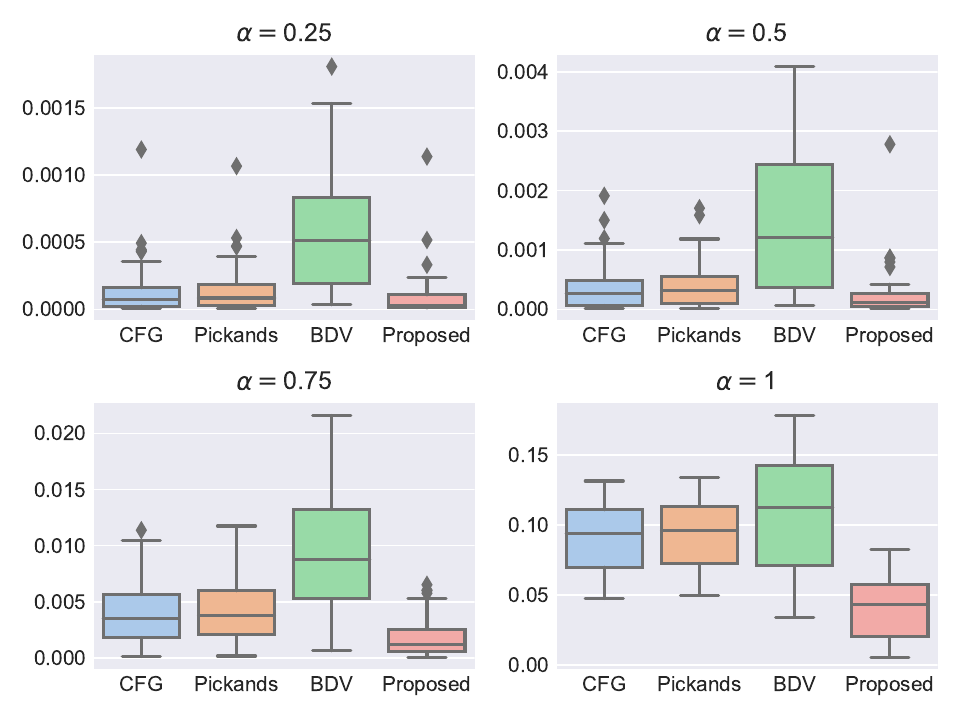}
  \caption{$A_\text{ASL}$ MSE ($d=256$)}
  \label{fig:asl_mse_est_all_a_a2}
\end{subfigure} 
\begin{subfigure}{.48\textwidth}
  \centering
\includegraphics[width=\linewidth]{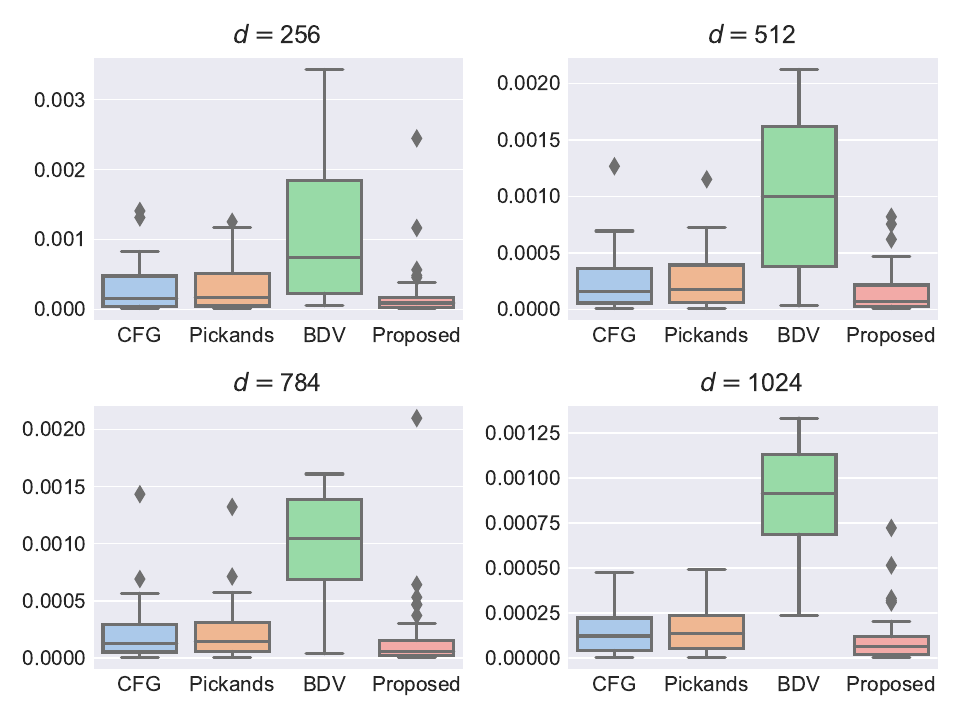}  
  \caption{$A_\text{SL}$ MSE ($\alpha=0.5$)}
  \label{fig:sl_mse_est_all_d_a2}
\end{subfigure}
\begin{subfigure}{.48\textwidth}
  \centering
    \includegraphics[width=\linewidth]{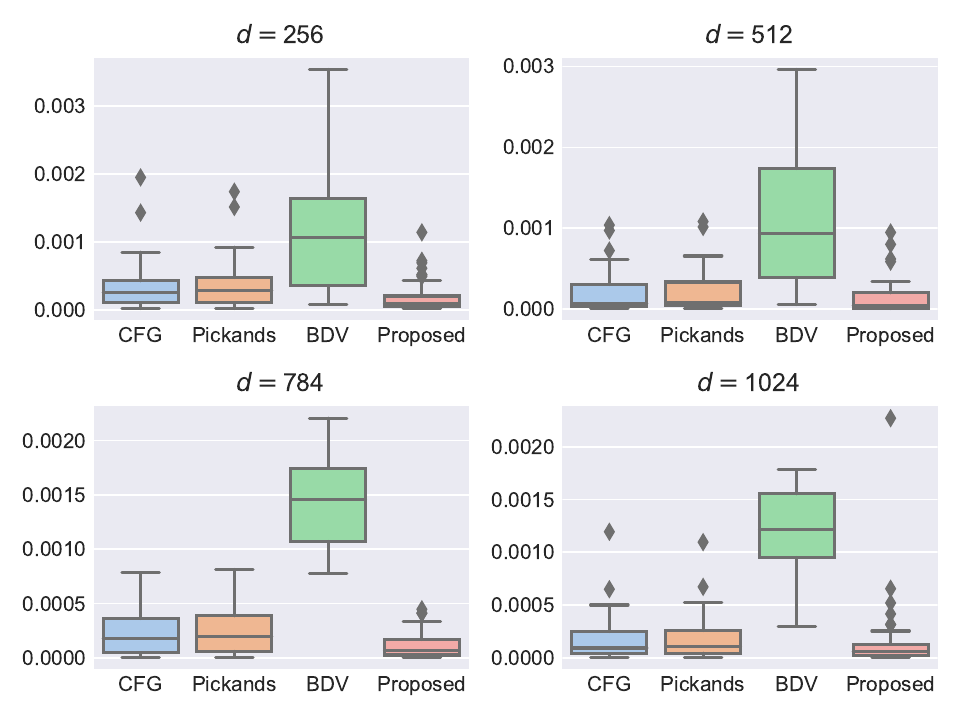}  
  \caption{$A_\text{ASL}$ MSE ($\alpha=0.5$)}
  \label{fig:asl_mse_est_all_d_a2}
\end{subfigure}  
\caption{Using 24 width 3 depth architecture: Comparison of $||\hat{A}(w) - A(w)||_2^2$ for different estimators $\hat{A}$ for different dependence $\alpha = \{0.25, 0.50, 0.75, 1.0\}$ with fixed $d=256$ (\ref{fig:sl_mse_est_all_a_a2}, \ref{fig:asl_mse_est_all_a_a2}) and for fixed  $\alpha=0.5$ with different $d = \{256, 512, 728, 1024\}$ (\ref{fig:sl_mse_est_all_d_a2}, \ref{fig:asl_mse_est_all_d_a2}). The truth models considered are $A_\text{SL}$ (\ref{fig:sl_mse_est_all_a_a2}, \ref{fig:sl_mse_est_all_d_a2}) and $A_\text{ASL}$ (\ref{fig:asl_mse_est_all_a_a2}, \ref{fig:asl_mse_est_all_d_a2}). Results are over 50 runs with 100 training samples for each run. } 
\end{figure*}
\begin{table*}[tbh!]
\newcommand{\timesten}{\text{\tiny $\times10$}}
\centering
\begin{tabular}{@{}lclllll@{}} 
\toprule
 & $d$ & Train/Test Length & \textsc{Pickands} & \textsc{CFG} & \textsc{BDV} & \textsc{Proposed}  \\ 
\midrule
 Wind & 10 & day/week &
 ${4.48(18.6)}\timesten^{-4}$ & $\textit{4.15(15.1)}\timesten^{-4}$ &   $\bf 4.10(16.3)\timesten^{-4}$ &   ${ 4.80(20.6)}\timesten^{-4}$ \\
 Ozone & 4 & day/week &
 $3.06(4.66)\timesten^{-2}$ &  $3.86(6.10)\timesten^{-2}$ & $\textit{2.86(4.46)}\timesten^{-2}$ & $\bf 2.82(4.38)\timesten^{-2}$ \\
 Commodities & 10 & week/month &
 $4.34(5.82)\timesten^{-3}$  & $4.33(5.71)\timesten^{-3}$   & $\bf 1.60(1.96)\timesten^{-3}$  &  $\textit{ 2.20(3.41)}\timesten^{-3}$  \\
 S\&P 500 & 418 & week/month &
 $\textit{3.02(21.2)}\timesten^{-3}$ & $\textit{3.02(21.1)}\timesten^{-3}$ & $6.28(35.2)\timesten^{-3}$ & 
 $\bf 2.41(22.2)\timesten^{-3}$ \\
 Crypto & 100 & week/month &
 ${1.06(2.85)}\timesten^{-2}$ & $\textit{1.05(4.86)}\timesten^{-2}$ & ${1.34(3.44)}\timesten^{-2}$ & $\bf 8.42(26.1)\timesten^{-3}$ \\
\bottomrule
\end{tabular}
    \caption{MSE of different estimators in estimating maxima over two time scales for 24 width 3 depth architecture. Best and second best performances are marked in \textbf{bold} and \textit{italic} respectively.}
    \label{tab:real_data_a2}
\end{table*}

\subsection{64 Width 4 Depth Architecture}

Here we repeat the experiments with a different architecture. 
All other hyperparameters are the same, the only difference is we increase the depth to 4 and use a width of 64 for each layer.
Most of the results remain similar for the synthetic data but we see a change in the results for the real data, specifically, the Wind and Commodities data show a deterioration in performance. 
However, the variances are still high for the real experiments and not much can be said regarding the efficacy of any single method. 

\begin{figure}[h!]
    \centering
\begin{subfigure}{.48\textwidth}
  \centering
  \includegraphics[width=\linewidth]{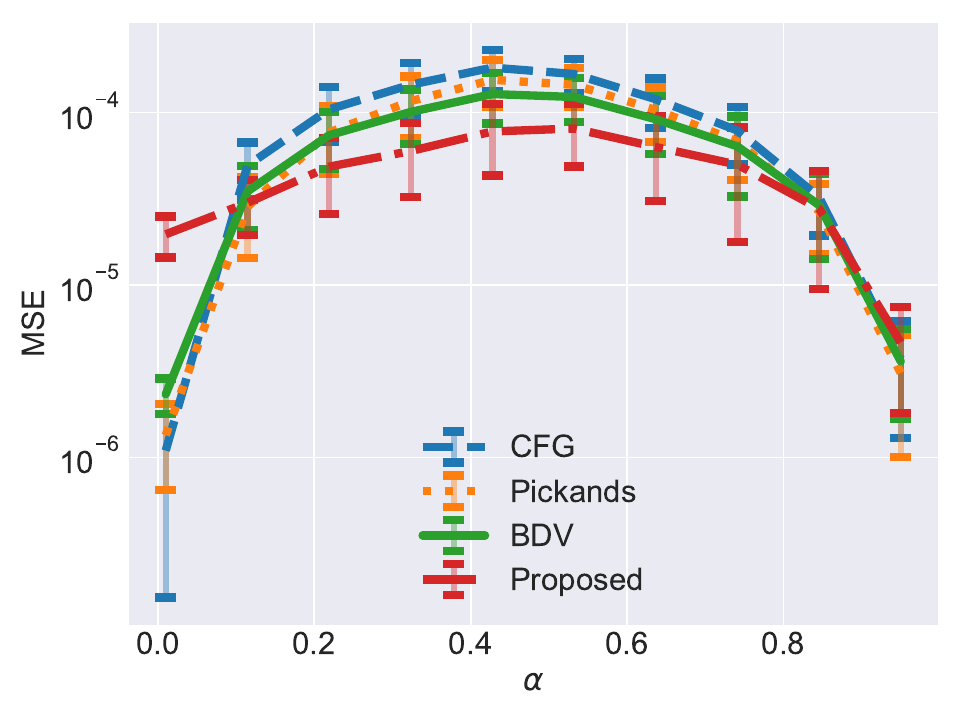}
  \caption{$A_\text{SL}$ MSE ($d=2$)}
  \label{fig:sl_survival64}
\end{subfigure}
\begin{subfigure}{.48\textwidth}
  \centering
  \includegraphics[width=\linewidth]{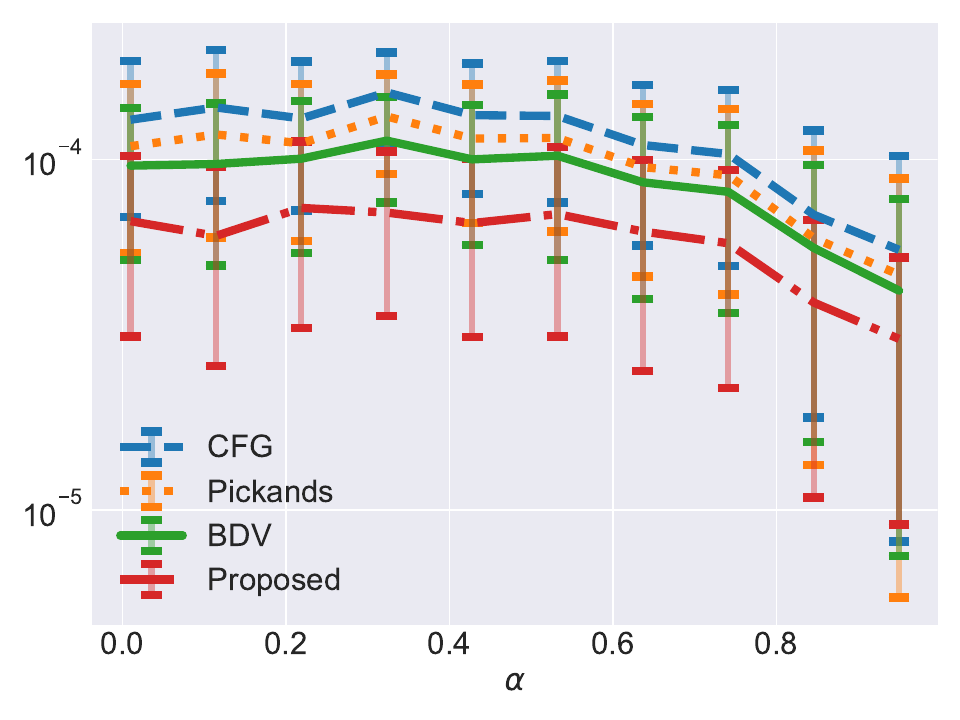}
  \caption{$A_\text{ASL}$ MSE ($d=2$)}
  \label{fig:asl_survival64}
\end{subfigure} 
\caption{Using 64 width 4 depth architecture: (\ref{fig:sl_survival64}, \ref{fig:asl_survival64}) MSE of survival probabilities for $d=2$ with $100$ samples for $A_\text{SL}$ (\ref{fig:sl_survival64}) and $A_\text{ASL}$ (\ref{fig:asl_survival64}). Thresholds are above the $75$th percentile.}
\end{figure}

\begin{figure}[tbh!]
    \centering
\begin{subfigure}{.48\textwidth}
  \centering
  \includegraphics[width=\linewidth]{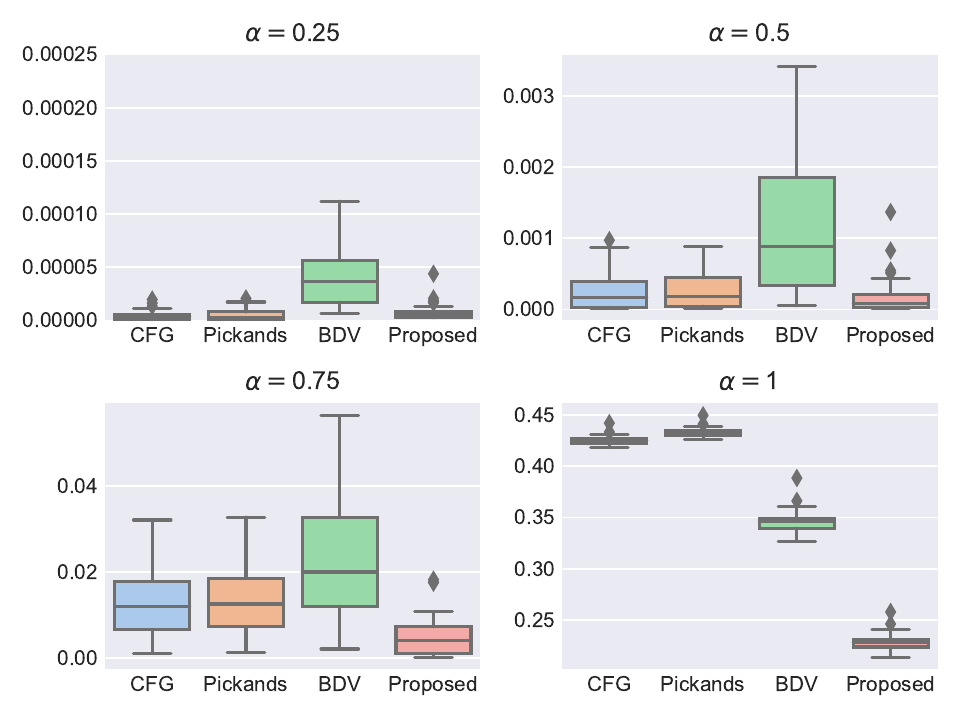} 
  \caption{$A_\text{SL}$ MSE ($d=256$)}
    \label{fig:sl_mse_est_all_a_a3}
\end{subfigure}
\begin{subfigure}{.48\textwidth}
  \centering
  \includegraphics[width=\linewidth]{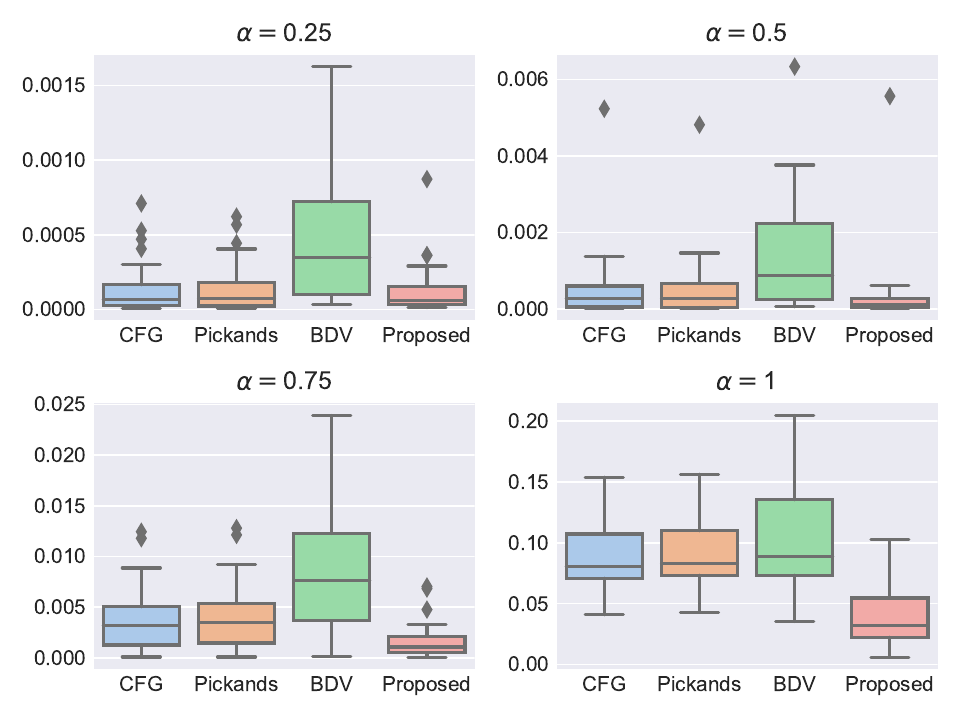}
  \caption{$A_\text{ASL}$ MSE ($d=256$)}
      \label{fig:asl_mse_est_all_a_a3}
\end{subfigure} 
\begin{subfigure}{.48\textwidth}
  \centering
\includegraphics[width=\linewidth]{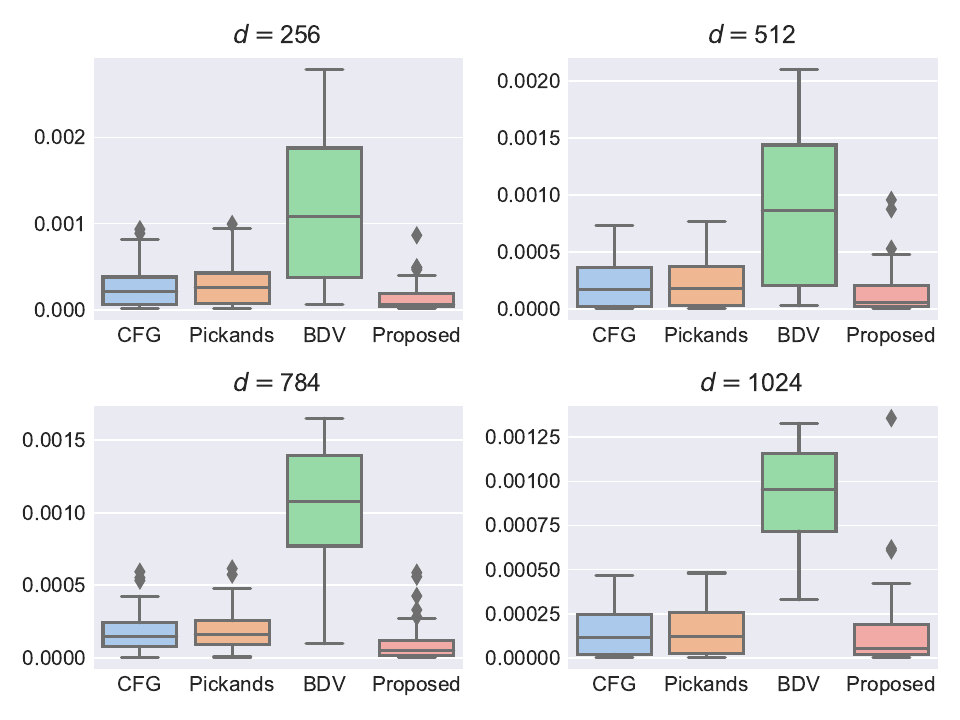}  
  \caption{$A_\text{SL}$ MSE ($\alpha=0.5$)}
    \label{fig:sl_mse_est_all_d_a3}
\end{subfigure}
\begin{subfigure}{.48\textwidth}
  \centering
    \includegraphics[width=\linewidth]{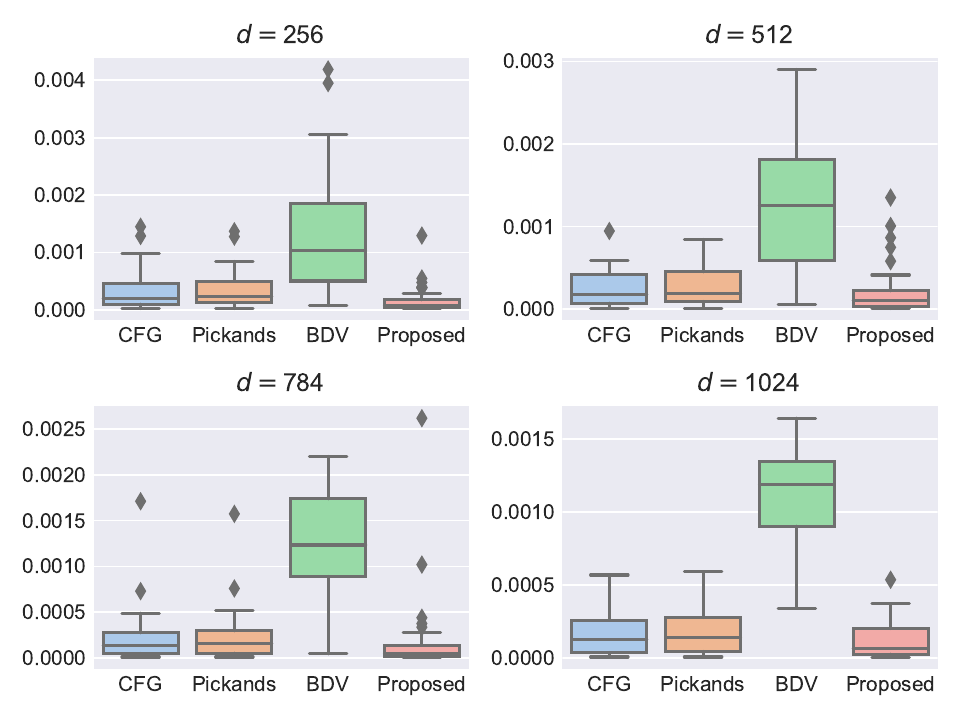}  
  \caption{$A_\text{ASL}$ MSE ($\alpha=0.5$)}
      \label{fig:asl_mse_est_all_d_a3}
\end{subfigure}  
\caption{Using 64 width 4 depth architecture: Comparison of $||\hat{A}(w) - A(w)||_2^2$ for different estimators $\hat{A}$ for different dependence $\alpha = \{0.25, 0.50, 0.75, 1.0\}$ with fixed $d=256$ (\ref{fig:sl_mse_est_all_a_a3}, \ref{fig:asl_mse_est_all_a_a3}) and for fixed  $\alpha=0.5$ with different $d = \{256, 512, 728, 1024\}$ (\ref{fig:sl_mse_est_all_d_a3}, \ref{fig:asl_mse_est_all_d_a3})  for $A_\text{SL}$ (\ref{fig:sl_mse_est_all_a_a3}, \ref{fig:sl_mse_est_all_d_a3}) and $A_\text{ASL}$ (\ref{fig:asl_mse_est_all_a_a3}, \ref{fig:asl_mse_est_all_d_a3}). Results are over 50 runs with 100 training samples for each run. } 
\end{figure}
\begin{table}[tbh!]
\newcommand{\timesten}{\text{\tiny $\times10$}}
\centering
\begin{tabular}{@{}lclllll@{}} 
\toprule
 & $d$ & Train/Test Length & \textsc{Pickands} & \textsc{CFG} & \textsc{BDV} & \textsc{Proposed}  \\ 
\midrule
 Wind & 10 & day/week &
 ${4.48(18.6)}\timesten^{-4}$ & $\textit{4.15(15.1)}\timesten^{-4}$ &   $\bf 4.10(16.3)\timesten^{-4}$ &   ${ 4.76(18.7)}\timesten^{-4}$ \\
 Ozone & 4 & day/week &
 $3.06(4.66)\timesten^{-2}$ &  $3.86(6.10)\timesten^{-2}$ & $\textit{2.86(4.46)}\timesten^{-2}$ & $\bf 2.73(4.25)\timesten^{-2}$ \\
 Commodities & 10 & week/month &
 $4.34(5.82)\timesten^{-3}$  & $4.33(5.71)\timesten^{-3}$  & $\bf 1.60(1.96)\timesten^{-3}$  &  $\textit{ 2.20(3.44)}\timesten^{-3}$  \\
 S\&P 500 & 418 & week/month &
 $\textit{3.02(21.2)}\timesten^{-3}$ & $\textit{3.02(21.1)}\timesten^{-3}$ & $6.28(35.2)\timesten^{-3}$ & 
 $\bf 2.39(22.1)\timesten^{-3}$ \\
 Crypto & 100 & week/month &
 ${1.06(2.85)}\timesten^{-2}$ & $\textit{1.05(4.86)}\timesten^{-2}$ & ${1.34(3.44)}\timesten^{-2}$ & $\bf 8.28(25.6)\timesten^{-3}$ \\
\bottomrule
\end{tabular}
    \caption{MSE of different estimators in estimating maxima over two time scales for 64 width 4 depth architecture. Best and second best performances are marked in \textbf{bold} and \textit{italic} respectively. }
    \label{tab:real_data_a3}
\end{table}

\subsection{Estimation Comparison}
We finally add a few figures comparing the learned dependence functions between different architectures.
We additionally provide a table comparing the results for different architectures on the real data experiments in Table~\ref{tab:ablation_real}.
\begin{figure}[tbh!]
    \centering
\begin{subfigure}{.30\textwidth}
  \centering
  \includegraphics[width=\linewidth]{imgs/margins/cali_extremal_MaxLinear_w512d1.pdf}  
  \caption{512 width 1 depth 2d Margins}
  \label{fig:net_512_wind}
\end{subfigure} 
\begin{subfigure}{.30\textwidth}
  \centering
  \includegraphics[width=\linewidth]{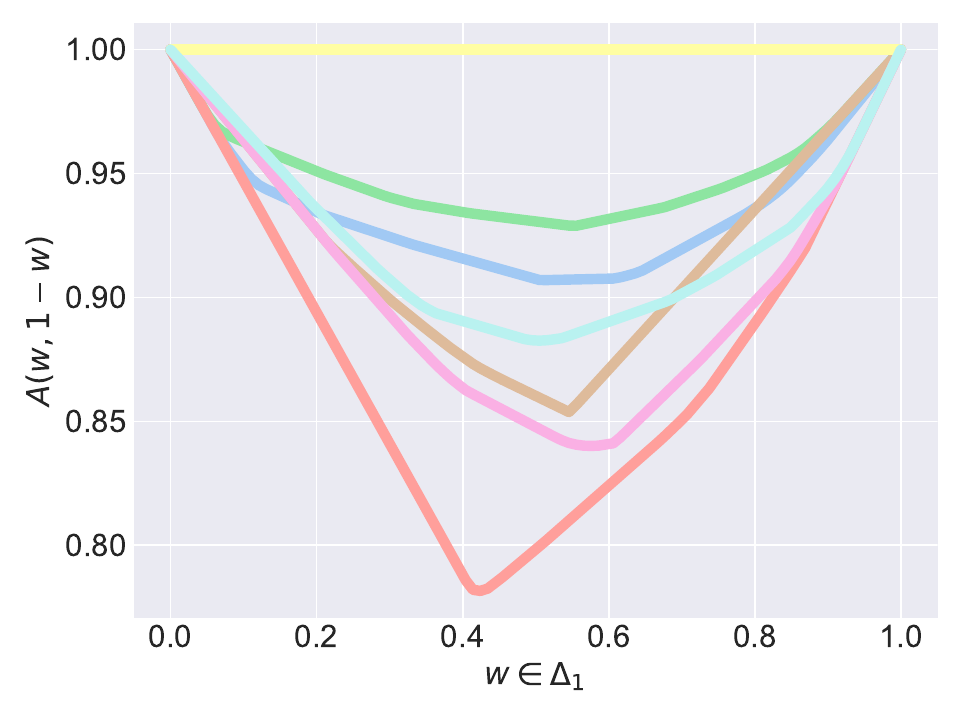}  
  \caption{24 width 3 depth 2d Margins}
  \label{fig:net_24_wind}
\end{subfigure}
\begin{subfigure}{.30\textwidth}
  \centering
  \includegraphics[width=\linewidth]{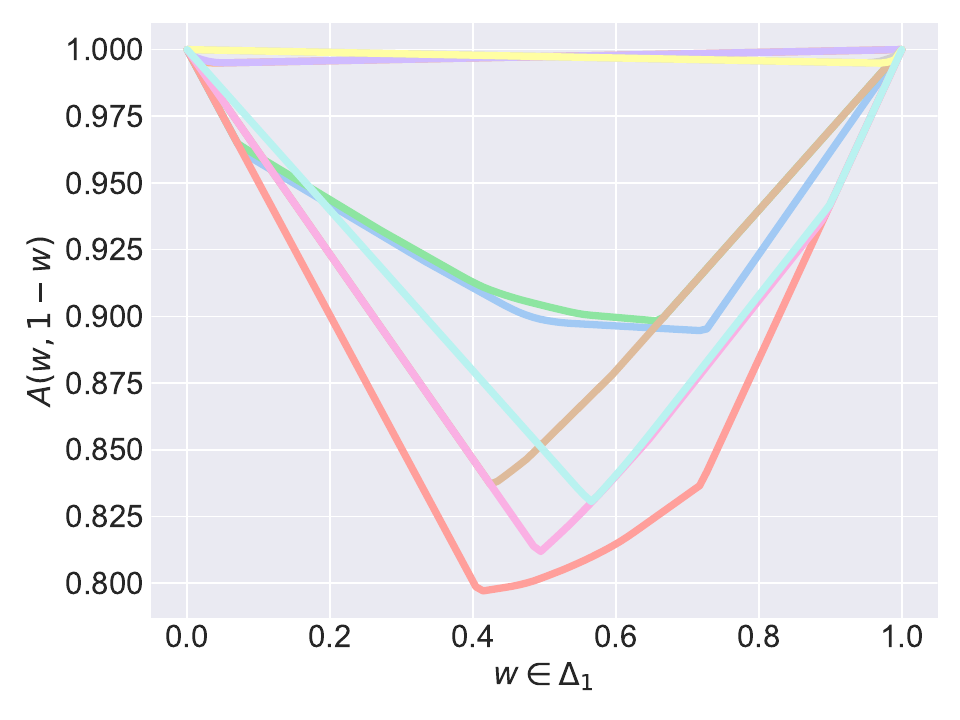}  
  \caption{64 width 4 depth 2d Margins}
  \label{fig:net_64_wind}
\end{subfigure}
\caption{Margin comparison for winds dataset. }
\label{fig:net_wind}
\end{figure}

\begin{figure}[tbh!]
    \centering
\begin{subfigure}{.30\textwidth}
  \centering
  \includegraphics[width=\linewidth]{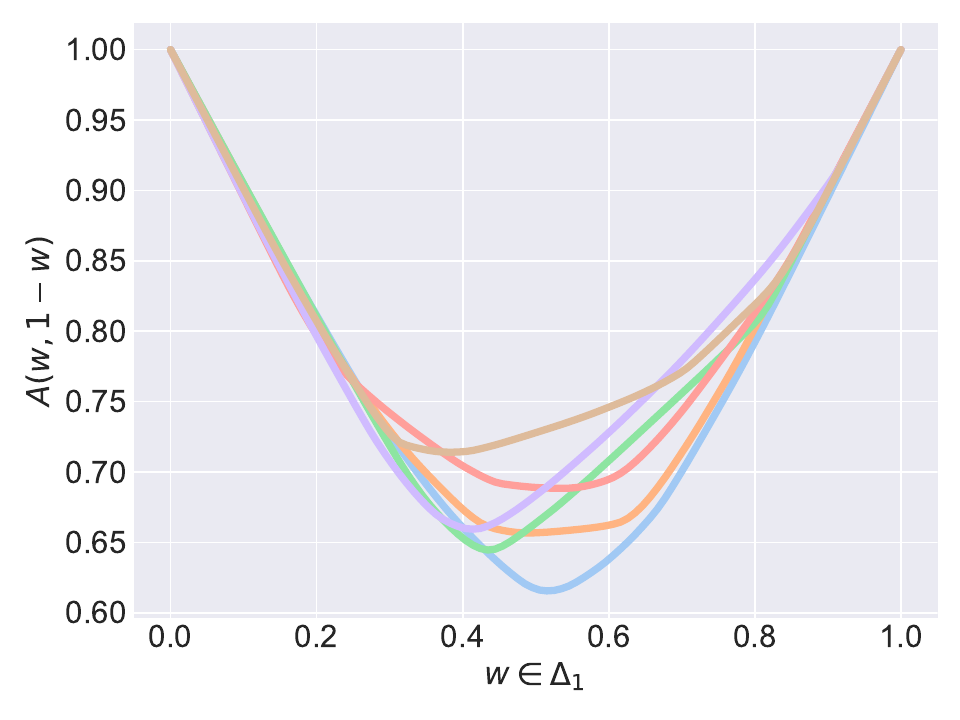}  
  \caption{512 width 1 depth 2d Margins}
  \label{fig:net_512_ozone}
\end{subfigure} 
\begin{subfigure}{.30\textwidth}
  \centering
  \includegraphics[width=\linewidth]{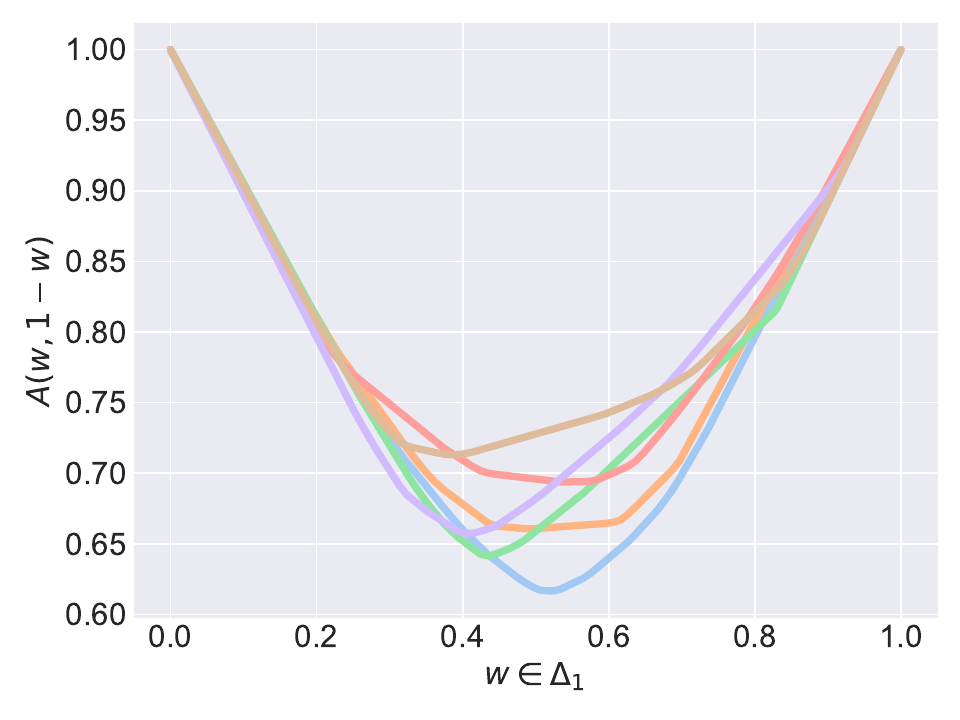}  
  \caption{24 width 3 depth 2d Margins}
  \label{fig:net_24_ozone}
\end{subfigure}
\begin{subfigure}{.30\textwidth}
  \centering
  \includegraphics[width=\linewidth]{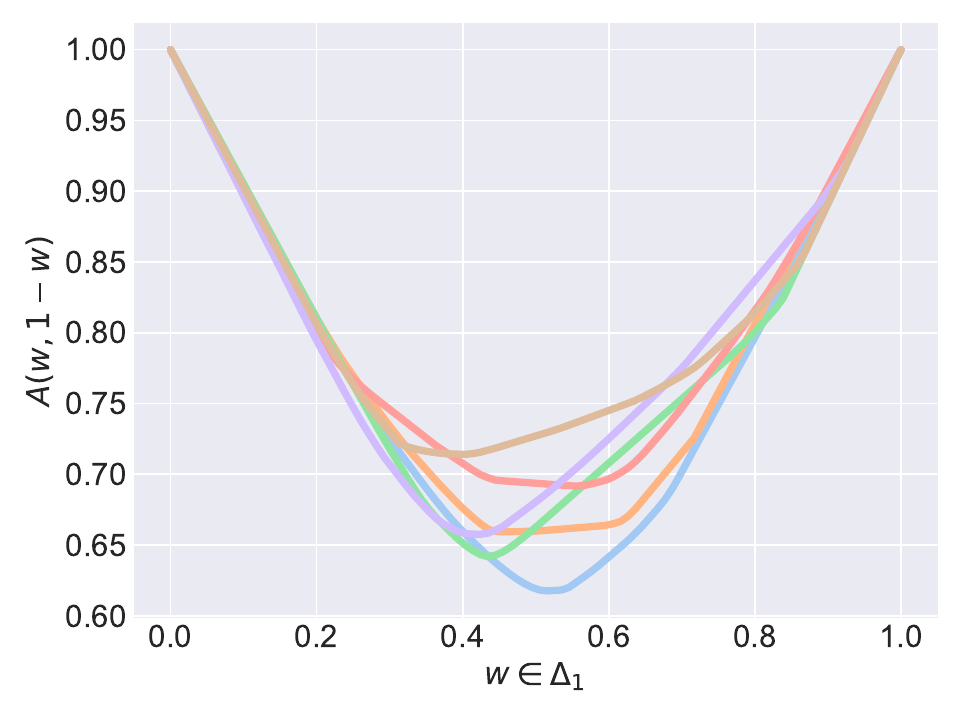}  
  \caption{64 width 4 depth 2d Margins}
  \label{fig:net_64_ozone}
\end{subfigure}
\caption{Margin comparison for ozone dataset. }
\end{figure}
\begin{figure}[tbh!]
    \centering
\begin{subfigure}{.30\textwidth}
  \centering
  \includegraphics[width=\linewidth]{imgs/margins/commodities_3d_MaxLinear_w512d1.pdf}  
  \caption{512 width 1 depth 3d Margins}
  \label{fig:net_512_comm}
\end{subfigure} 
\begin{subfigure}{.30\textwidth}
  \centering
  \includegraphics[width=\linewidth]{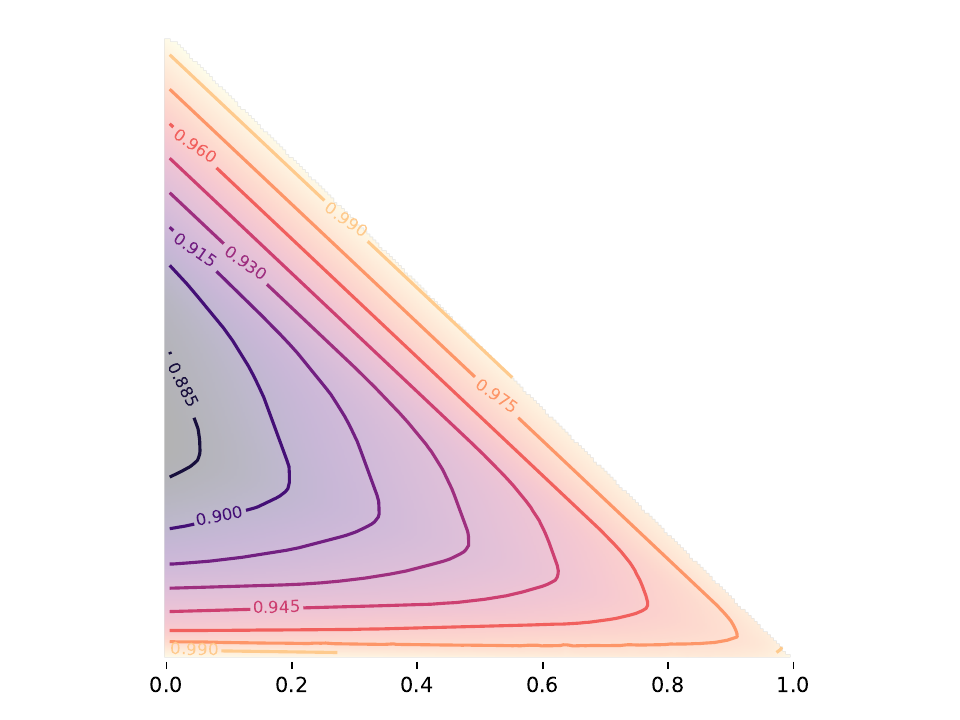}  
  \caption{24 width 3 depth 3d Margins}
  \label{fig:net_24_comm}
\end{subfigure}
\begin{subfigure}{.30\textwidth}
  \centering
  \includegraphics[width=\linewidth]{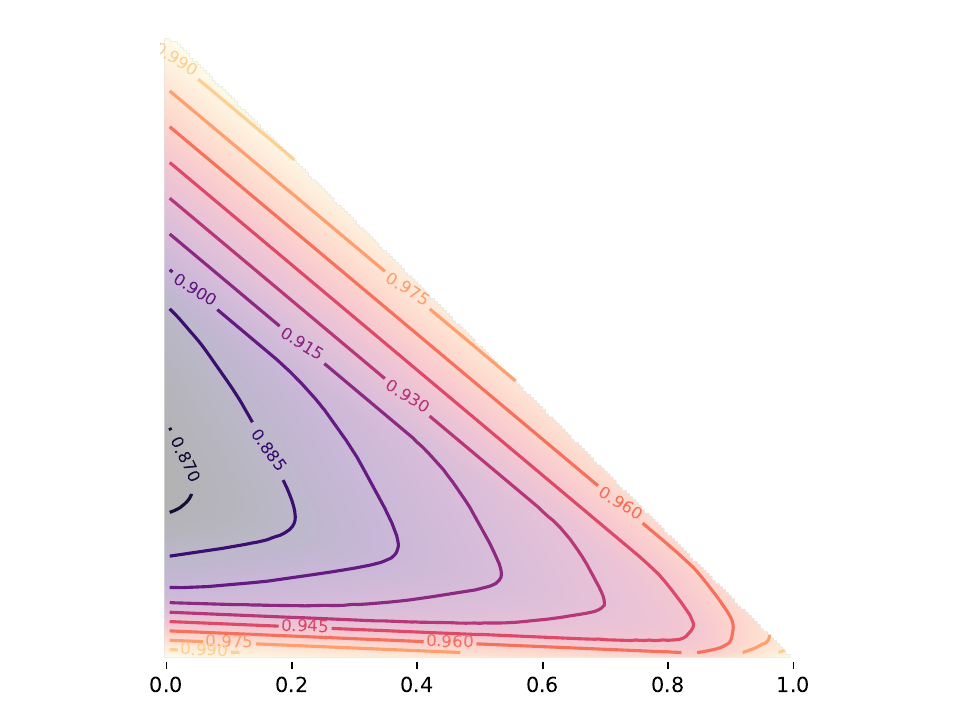}  
  \caption{64 width 4 depth 3d Margins}
  \label{fig:net_64_comm}
\end{subfigure}
\caption{Margin comparison for commodities dataset. }
\end{figure}

\begin{figure}[tbh!]
    \centering
\begin{subfigure}{.30\textwidth}
  \centering
  \includegraphics[width=\linewidth]{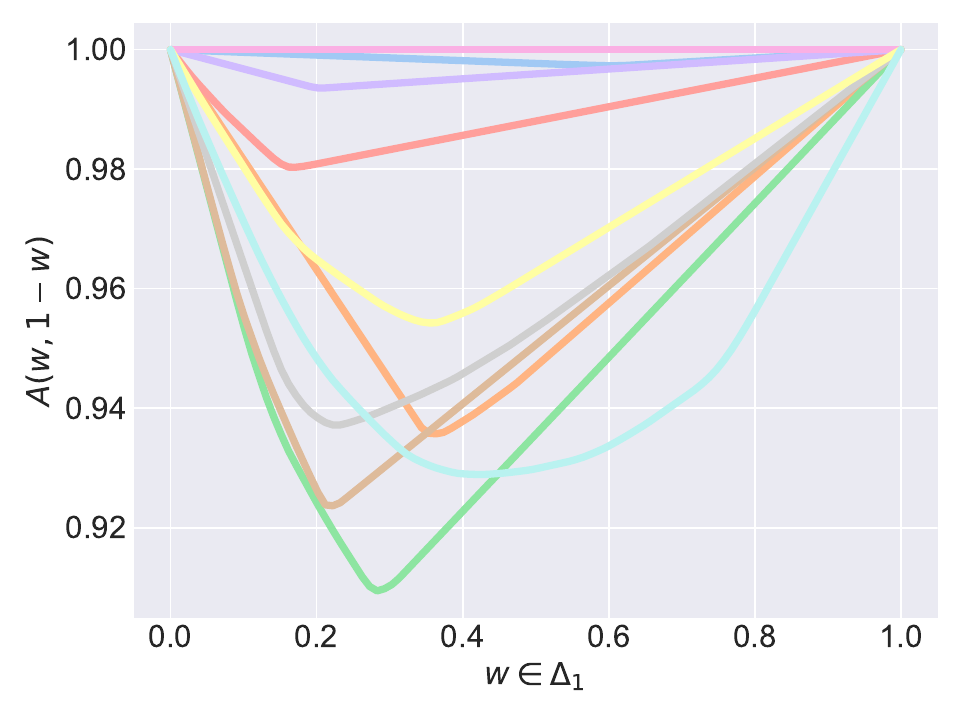}  
  \caption{512 width 1 depth 2d Margins}
  \label{fig:net_512_comm}
\end{subfigure} 
\begin{subfigure}{.30\textwidth}
  \centering
  \includegraphics[width=\linewidth]{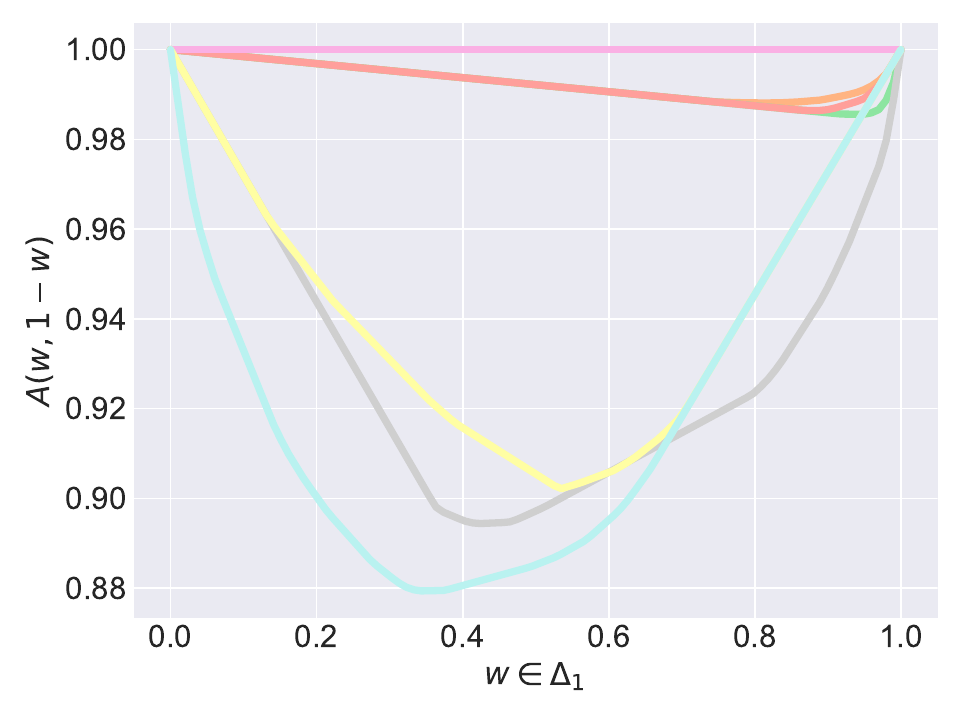}  
  \caption{24 width 3 depth 2d Margins}
  \label{fig:net_24_comm}
\end{subfigure}
\begin{subfigure}{.30\textwidth}
  \centering
  \includegraphics[width=\linewidth]{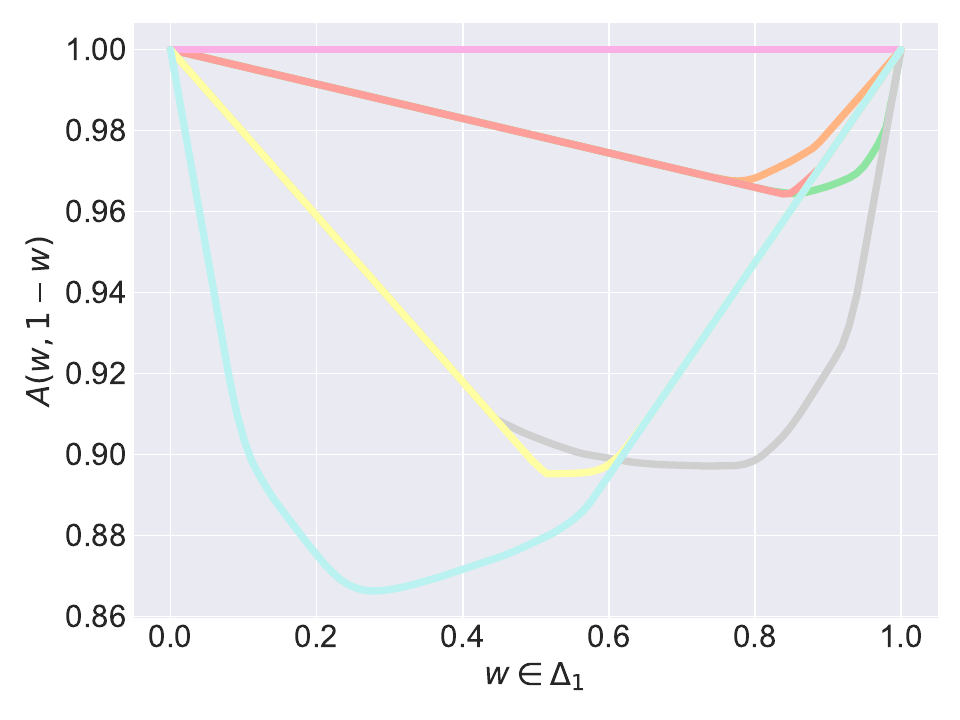}  
  \caption{64 width 4 depth 2d Margins}
  \label{fig:net_64_comm}
\end{subfigure}
\caption{Margin comparison for commodities dataset. }
\end{figure}
\begin{figure}[tbh!]
    \centering
\begin{subfigure}{.30\textwidth}
  \centering
  \includegraphics[width=\linewidth]{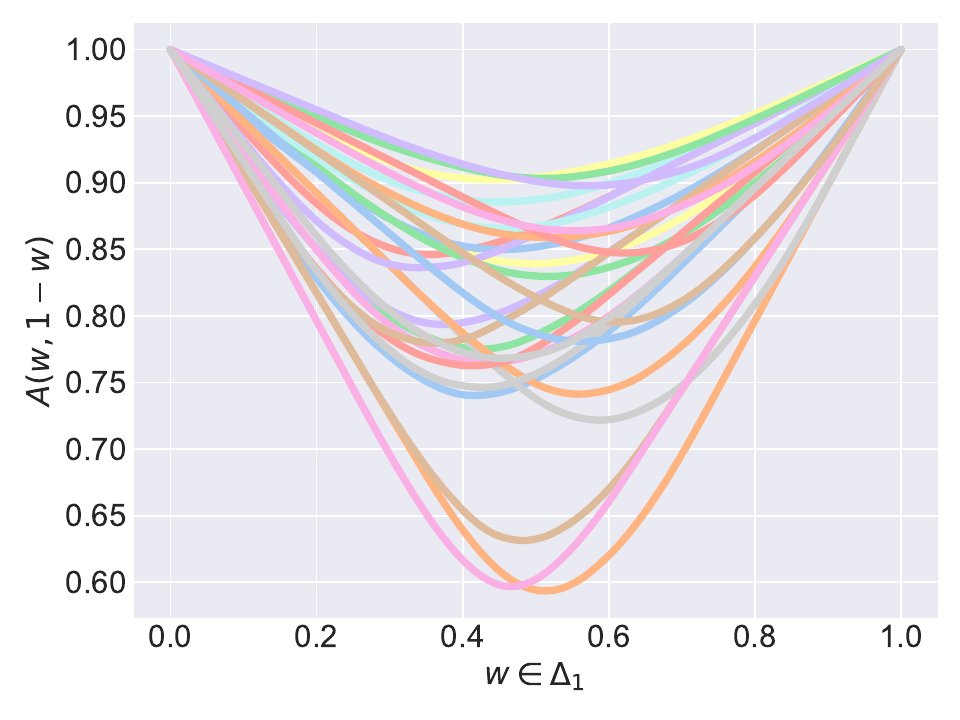}  
  \caption{512 width 1 depth 2d Margins}
  \label{fig:cfg_marg_spy}
\end{subfigure} 
\begin{subfigure}{.30\textwidth}
  \centering
  \includegraphics[width=\linewidth]{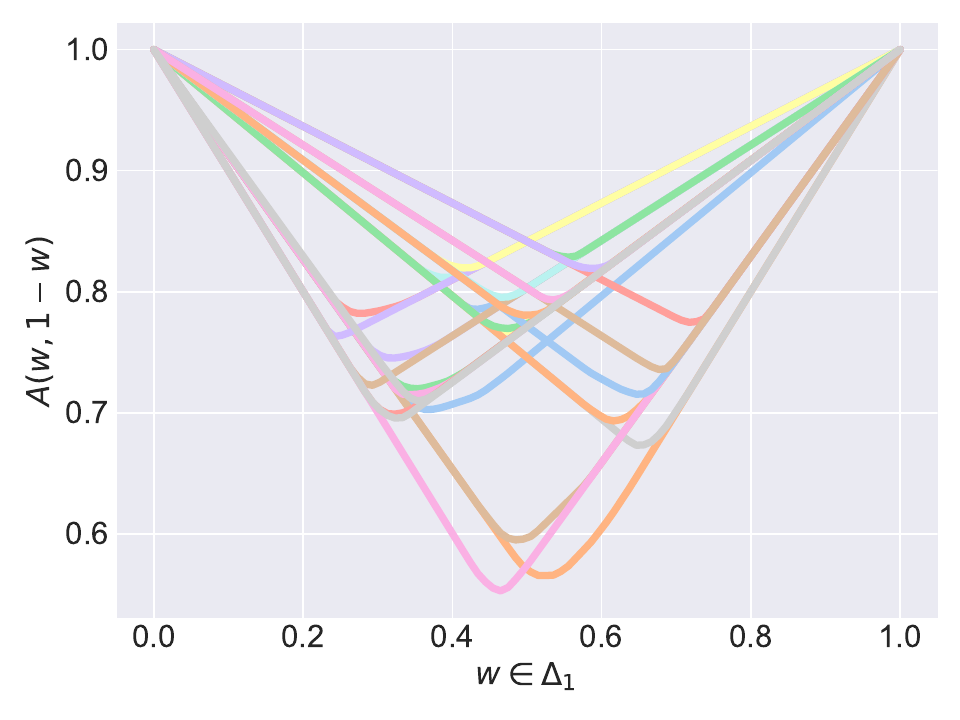}  
  \caption{24 width 3 depth 2d Margins}
  \label{fig:bdv_marg_spy}
\end{subfigure}
\begin{subfigure}{.30\textwidth}
  \centering
  \includegraphics[width=\linewidth]{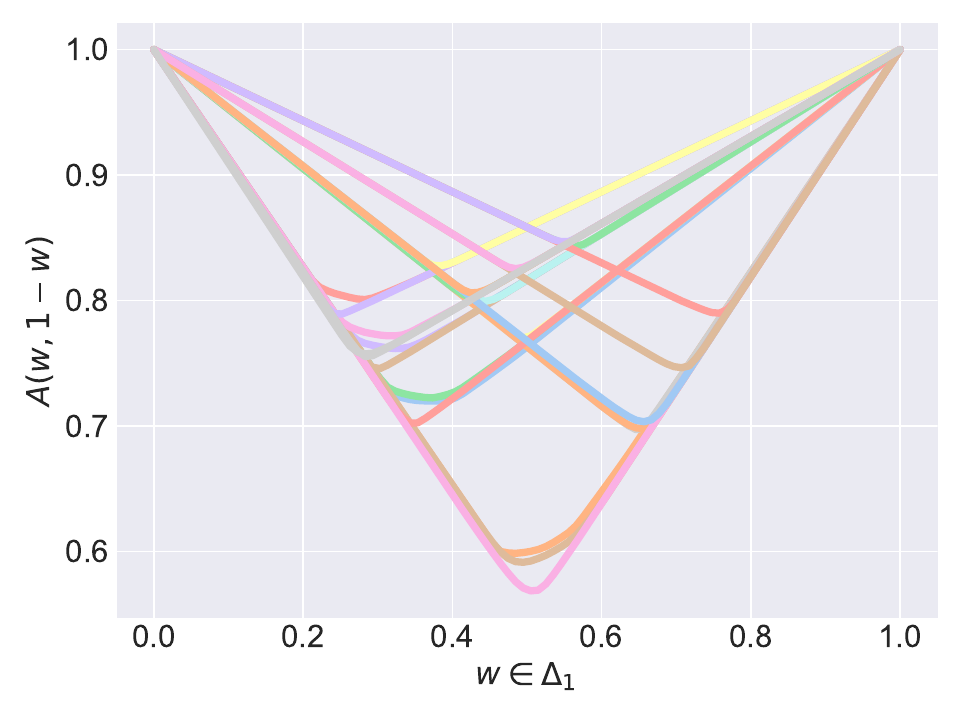}  
  \caption{64 width 4 depth 2d Margins}
  \label{fig:net_marg_spy}
\end{subfigure}
\caption{Margin comparison for S\&P 500 dataset. }
\end{figure}

\begin{figure}[tbh!]
    \centering
\begin{subfigure}{.30\textwidth}
  \centering
  \includegraphics[width=\linewidth]{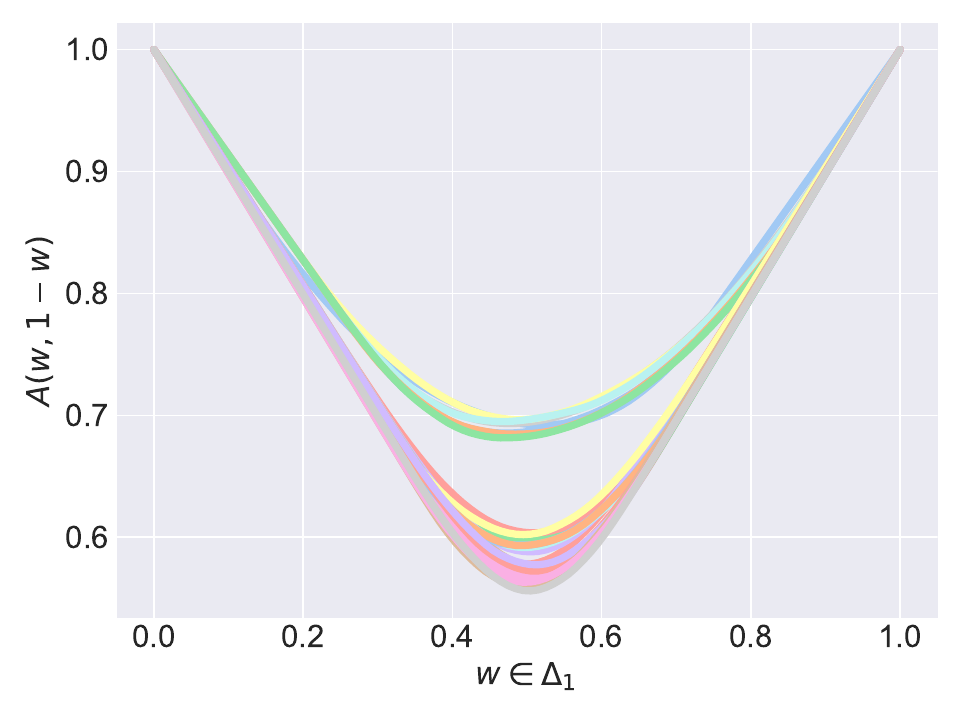}  
  \caption{512 width 1 depth 2d Margins}
  \label{fig:net_512_crypto}
\end{subfigure} 
\begin{subfigure}{.30\textwidth}
  \centering
  \includegraphics[width=\linewidth]{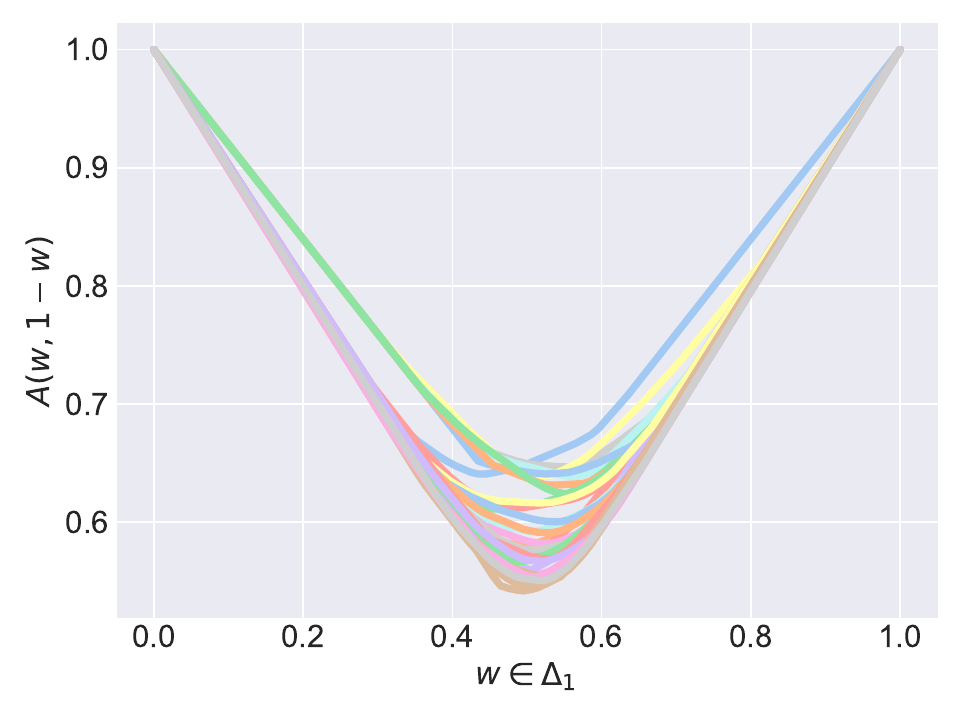}  
  \caption{24 width 3 depth 2d Margins}
  \label{fig:net_24_crypto}
\end{subfigure}
\begin{subfigure}{.30\textwidth}
  \centering
  \includegraphics[width=\linewidth]{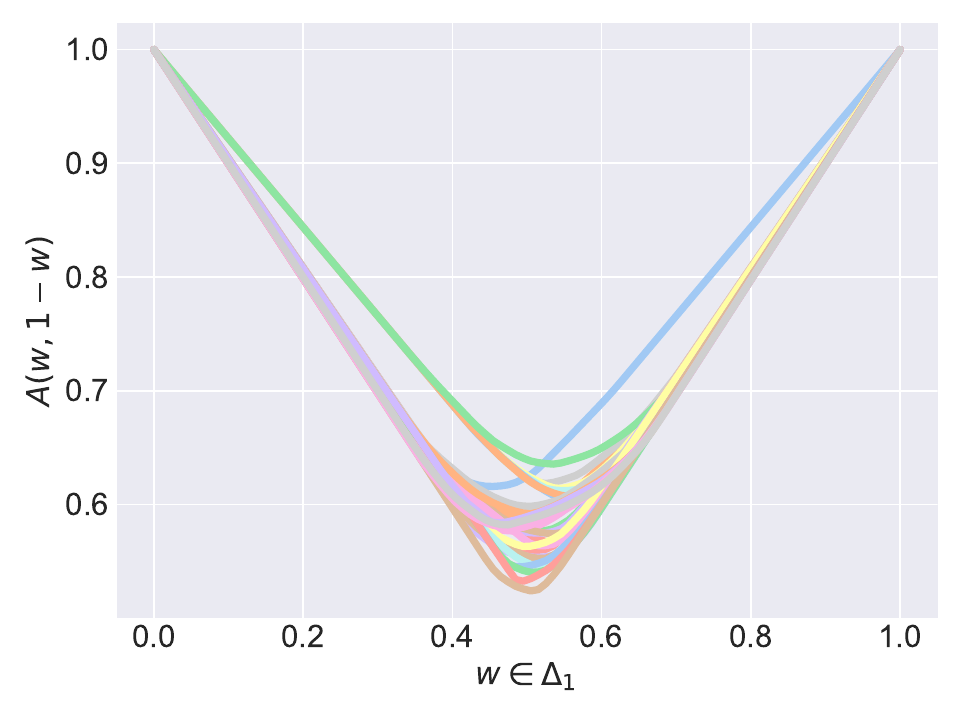}  
  \caption{64 width 4 depth 2d Margins}
  \label{fig:net_64_crypto}
\end{subfigure}
\caption{Margin comparison of different architectures for Cryptocurrencies dataset. }
\label{fig:net_crypto}
\end{figure}

\begin{table}[tbh!]
    \newcommand{\timesten}{\text{\tiny $\times10$}}
    \centering
    \begin{tabular}{cccc}
    \toprule
        & $512 \text{width} \times 1 \text{depth}$ & $24 \text{width} \times 3 \text{depth}$ & $64 \text{width} \times 4 \text{depth}$  \\ \midrule
         Wind & ${ 4.37(17.5)}\timesten^{-4}$ & ${ 4.80(20.6)}\timesten^{-4}$ & ${ 4.76(18.7)}\timesten^{-4}$ \\
         Ozone & $ 2.73(4.25)\timesten^{-2}$ & $2.82(4.38)\timesten^{-2}$ & $2.73(4.25)\timesten^{-2}$ \\
         Commodities & $1.56(2.21)\timesten^{-3}$ & ${ 2.20(3.41)}\timesten^{-3}$  & $2.20(3.44)\timesten^{-3}$ \\
         S \& P & $ 2.41(22.2)\timesten^{-3}$ & $2.41(22.2)\timesten^{-3}$ & $2.39(22.1)\timesten^{-3}$ \\
         Crypto & $8.57(26.4)\timesten^{-3}$ & $8.42(26.1)\timesten^{-3}$ & $8.28(25.6)\timesten^{-3}$ \\
         \bottomrule
    \end{tabular}
    \caption{Comparison of 3 different architectures on the real data experiments.}
    \label{tab:ablation_real}
\end{table}

\section{Data Description}
\label{sec:data}
\subsubsection*{Synthetic Data}
For the synthetic data experiments we consider samples of 100 points from each respective distribution. 
We use the full dataset for the batch size during training.
We additionally sample 1000 points from the simplex for each data point during training. 

\subsubsection*{Ozone Data}
We consider ozone levels measured at 4 different stations in Sequoia National Park from data that can be downloaded from the National Park Service website \footnote{\url{https://ard-request.air-resource.com/data.aspx}}. The 4 stations are located at Ash Mountain, Lower Kaweah, Grant Grove and Lookout Point. We train the different models on daily maxima of ozone levels at the 4 different stations for the period from January 1984 to December 1996. To reduce the effect of seasonality, we do not train over the whole period, but we train different models on a single month (training month, e.g. June of each year) and compute accuracy on the consecutive month (validation month e.g. July of the same year). We additionally only look at summer months due to the increase of extreme events during that time. The accuracy is averaged with the specific validation month of each year over the whole period. We train on daily maxima and test on weekly maxima. For the experiments, we consider the following pair of (training/test) months: (June/July), (July/August), and (August/September). 

\begin{figure}
    \centering
     \includegraphics[width=0.5\linewidth]{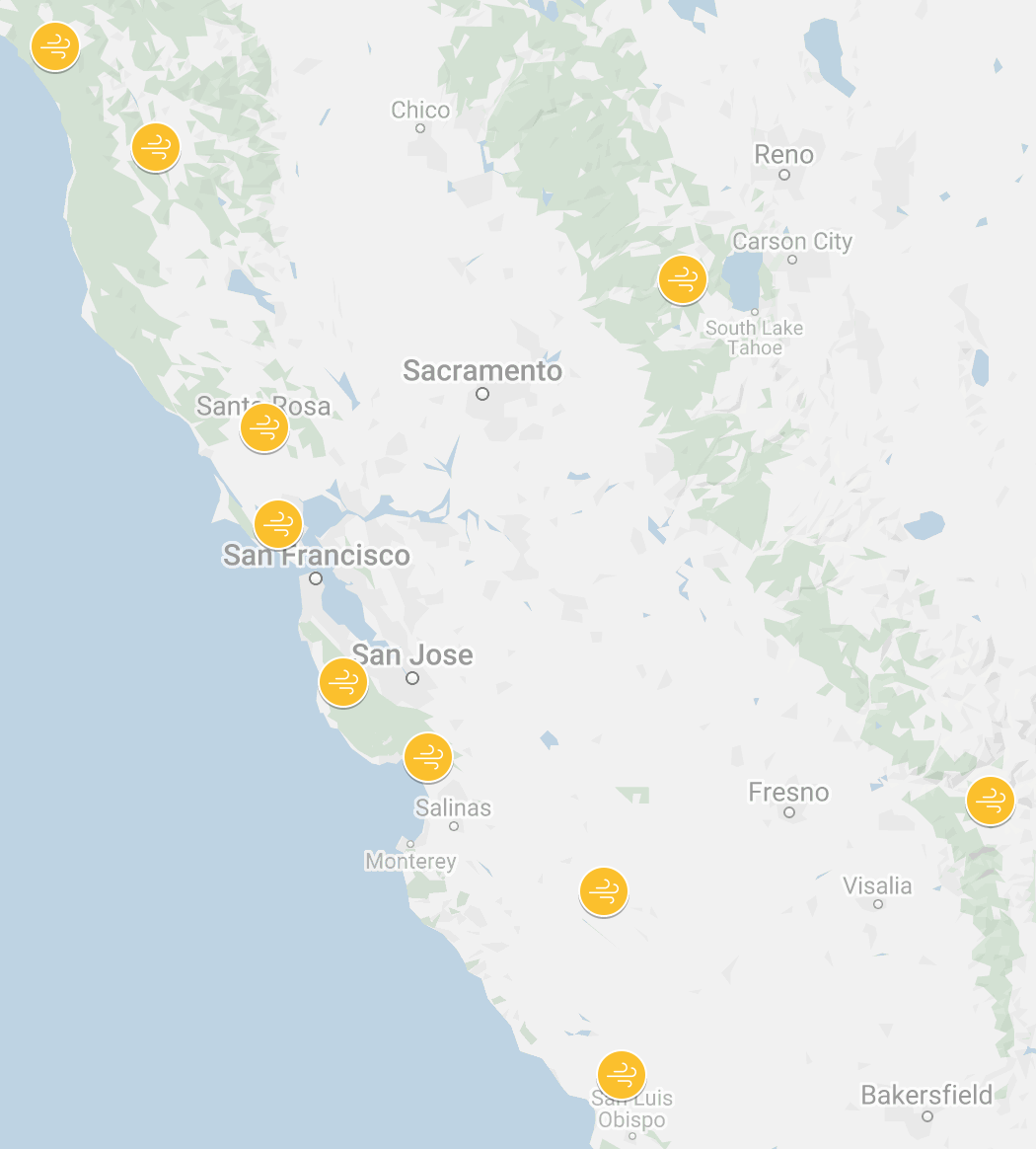}
     \caption{Locations (yellow circles) of weather stations sampled for the wind speed experiments. Figure generated via Google Maps.}
     \label{fig:wind_locations}
\end{figure}

\subsubsection*{California Wind Data}
We are interested in modeling the extremal relationship of wind gusts between different locations in California during the summer months. 
We consider 10 locations in California illustrated in \ref{fig:wind_locations}. 
We obtained the data from the Remote Automated Weather Station (RAWS) archive available at the online repository\footnote{\url{https://raws.dri.edu/index.html}}. 
The RAWS data are collected from various time intervals from December 1989 to December 2020. 
We consider only the time points that occur in the intersection of all the data collected and where all values are valid (i.e. not NaNs or missing) for the summer months.
Similarly to the ozone data, and in an effort to reduce seasonality, we consider the daily max wind gust for the different locations for a single month over all the years the data were collected.
To evaluate the proposed method, we train and test on data from consecutive months and repeat for multiple sets of months in our dataset.
Additionally, we train on daily max and test on monthly max using the following data splitting scheme (training/validation months): (June/July), (July/August), and (August/September). 

\subsubsection*{Commodities Data}
We consider the extreme dependency between different commodities such as Coffee, Copper, Corn, Crude Oil, Gold, Heating Oil, Natural Gas, Platinum, Silver and Wheat. We collect data of daily prices of the different commodities from January 2015 to December 2020 as published in \footnote{\url{https://www.investing.com/commodities/}}. For training, we consider weekly max drawdown over a year. We validate the performance by evaluating accuracy of monthly max drawdown over next three years. We consider the following pairs of ([training years],[validation years]): ([2015], [2016, 2017, 2018]), ([2016], [2017, 2018, 2019]), ([2017], [2018, 2019, 2020]).

\subsubsection*{S\&P 500 Data}
We obtain historical data from \url{https://www.alphavantage.co}\footnote{Alpha Vantage allows academic use as long as the website is cited.}. 
We choose the components of the S\&P 500 with sufficient history (resulting in 418 stocks). 
For training, we consider weekly max drawdown over a year. 
We validate the performance by evaluating accuracy of monthly max drawdown over next three years. We consider the following pairs of ([training years],[validation years]): ([2015], [2016, 2017, 2018]), ([2016], [2017, 2018, 2019]), ([2017], [2018, 2019, 2020]).
For the full list of stocks, see the \verb|sp_names.txt| file in the supplementary materials. 

\subsubsection*{Cryptocurrencies Data}
We obtain historical data from \url{https://coinmarketcap.com}\footnote{Coin Market Cap allows academic use as long as the website is cited (see FAQ page).} for 100 coins with the longest history. 
For training, we consider weekly max drawdown over a year. 
We validate the performance by evaluating accuracy of monthly max drawdown over next three years. We consider the following pairs of ([training years],[validation years]): ([2015], [2016, 2017, 2018]), ([2016], [2017, 2018, 2019]), ([2017], [2018, 2019, 2020]).
For the full list of coins, see the \verb|crypto_names.txt| file in the supplementary materials. 

\section{Pickands, CFG and BDV Estimators}
\label{sec:estimators}
\subsubsection*{Pickands Estimator}
The Pickands estimator \cite{pickands1981multivariate} is built following the transformations (6) and (7) in the paper. The estimator is obtained by exactly maximizing the likelihood (Equation (8) in the paper) resulting in the following non-parametric estimate: 
\begin{align}
    \label{pickands_est}
    \widehat{A}_{\text{Pickands}}(\mathbf{w}) = \left( \frac{1}{B}\sum_{i=1}^B Z_{w, i} \right)^{-1}.
\end{align}

\subsubsection*{CFG Estimator}
The CFG estimator \cite{caperaa1997nonparametric} is constructed following the observation: 
\begin{align*}
    \mathbb{E} \log Z_w = -\log A(\mathbf{w}) - \gamma,
\end{align*}
where $\gamma = -\int_{0}^{\infty} \log x e^{-x} dx$ denotes the Euler's contant. The CFG estimator is thus given by:
\begin{align}
    \label{cfg_est}
    \widehat{A}_{\text{CFG}}(\mathbf{w}) = \exp \left[ -\gamma - \frac{1}{B}\sum_{i=1}^B \log Z_{w, i}\right].
\end{align}
In our main submission we use a similar estimator, with the correction term presented in \cite{gudendorf2011nonparametric}:
\begin{equation}
    \widehat{A}_{\text{CFG,C}}(\mathbf{w})= \exp\left( \log \widehat{A}_{\text{CFG}}(\mathbf{w}) - \sum_{k=1}^d w_k\log\left(\widehat{A}_{\text{CFG}}(\mathbf{e}_k)\right)\right), 
\end{equation}
where $\mathbf{e}_k$ is the $k$-th canonical basis vector.
\subsubsection*{BDV Estimator}
We propose an $d$-dimensional extension to the bivariate estimator described in \cite{bucher2011new}.
We begin by defining the minimum distance estimator between the true CDF, $C(\mathbf{u})$ and the one estimated by the Pickands function $A(\mathbf{w})$.
\begin{align}
&\int_{[0,1]^d} \left[\log C(\mathbf u)  - \sum_{k=1}^d \log u_k A\left(\frac{\log(\mathbf u)}{\sum_k \log u_k}\right)\right]^2 \,  d\mathbf{u} \\
&= \int_{\Delta_{d-1}}\int_{0}^1 (\log C(y^{w_1}, \dots, y^{w_{d}})  - \log (y) A(\mathbf{w}))^2 (-\log(y))^{d-1}\, dy\, d\mathbf{w}.
\end{align}
We have
\begin{align}
\hat C(y^{w_1}, \dots, y^{w_d}) 
&= \frac{1}{B} \sum_{i=1}^B \mathbf{1}( F_{1}(\bar{M}_{n, i}^{(1)})\le y^{w_1}, \cdots, F_{d}(\bar{M}_{n, i}^{(d)})\le y^{w_d})\\
&= \frac{1}{B} \sum_{i=1}^B \mathbf{1}(F_{1}(\bar{M}_{n, i}^{(1)})^{\frac1{w_1}}\le y, \cdots, F_{d}(\bar{M}_{n, i}^{(d)})^{\frac1{w_d}}\le y)\\
&= \frac{1}{B} \sum_{i=1}^B \mathbf{1}(\max_{1 \leq k \leq d} F_{k}(\bar{M}_{n, i}^{(k)})^{\frac1{w_k}}\le y)\\
&= \frac{1}{B} \sum_{i=1}^B \mathbf{1}\left(\Gamma_{w, i} \leq y \right),
\end{align}
where $\Gamma_{w, i} = \exp(-Z_{w,i})$. Now, if we reorder these so that $\Gamma_{w, 1}\le \cdots \le \Gamma_{w, B}$, we have that
\begin{equation}
\hat C(y^{w_1}, \dots, y^{w_d})=\left\{\begin{aligned}
0&\;\:\text{if}\;\: y < \Gamma_{w, 1},\\
\frac{i}{B}&\;\:\text{if}\;\: \Gamma_{w, i}\le y < \Gamma_{w, i+1},\, i\in \{1,\dots,B-1\},\\
1&\;\:\text{if}\;\: \Gamma_{w, B}\le y.
\end{aligned}\right.
\end{equation}
Because $\log \hat C(\cdots)$ is not defined if $y < \Gamma_{w, 1}$, the following modified estimator is considered in \cite{bucher2011new}.
\begin{equation}
\tilde C(y^{w_1}, \dots, y^{w_d}):= \max\left\{C(y^{w_1}, \dots, y^{w_d}), B^{-\gamma}\right\},
\end{equation}
where $\gamma$ is any positive real greater or equal than $\frac12$. For convenience, we choose $\gamma=1$ so that:
\begin{equation}
\tilde C(y^{w_1}, \dots, y^{w_d})=\left\{\begin{aligned}
\frac{1}{B}&\;\:\text{if}\;\: y < \Gamma_{w, 2},\\
\frac{i}{B}&\;\:\text{if}\;\: \Gamma_{w, i}\le y < \Gamma_{w, i+1},\, i\in \{2,\dots,B-1\},\\
1&\;\:\text{if}\;\: \Gamma_{w, B}\le y.
\end{aligned}\right.
\end{equation}
Finally, as in \cite{bucher2011new}, for any positive weight function $h:(0,1)\to \mathbb{R}_0^+$, let $h^*(y):= h(y) (\log y)^2$,
\begin{equation}
B_h:= \int_{0}^1 h^*(y)\,dy\quad\text{and}\quad g(x) := - B_h^{-1} \int_{0}^x \frac{h^*(y)}{\log y}\,dy.
\end{equation}
Then, letting $\Gamma_{w,0}= 0$, $\Gamma_{w,B+1}= 1$, we define the BDV estimator $\widehat{A}_{\text{BDV,}h}$ as follows
\begin{align}
\widehat{A}_{\text{BDV,}h}(\mathbf{w}) &= B_h^{-1} \int_{0}^1 \frac{\log \tilde C(y^{w_1}, \dots, y^{w_d})}{\log y} h^*(y)\, dy \\
&= B_h^{-1} \sum_{i=0}^B \int_{\Gamma_{w,i}}^{\Gamma_{w,i+1}} \frac{\log \tilde C(y^{w_1}, \dots, y^{w_d})}{\log y} h^*(y)\, dy\\
&= -\log\frac{1}n g(\Gamma_{w, 2}) -\sum_{i=2}^n \log\frac{i}n \left(g(\Gamma_{w, i+1})-g(\Gamma_{w, i})\right)\\
&= -\sum_{i=2}^n \log\frac{i-1}n g(\Gamma_{w, i}) + \sum_{i=2}^n \log\frac{i}n g(\Gamma_{w, i})\\
&= \sum_{i=2}^n \log\left(1+\frac1{i-1}\right) g(\Gamma_{w, i})
\end{align}
In our main submission, we use a slightly modified estimator, which proved to have superior performance in our experiments. Recall that, if $A$ is a Pickand's dependence function, we have $\max(\mathbf{w})\le A(\mathbf{w})\le 1$, which implies the true copula verifies:
\begin{equation}\label{iusethisjustbelow}
\max(\mathbf{w}) \le \frac{\log C(y^{w_1}, \dots, y^{w_d})}{\log y} = A(\mathbf{w}) \le 1.
\end{equation}
Accordingly, we let
\begin{equation}
    \clamp_{a,b}(x):=
    \left\{\begin{array}{rl}
    a&\text{if }x\le a,\\
    x&\text{if }a<x<b,\\
    b&\text{if }x\ge b,
    \end{array}\right.
\end{equation}
and define
\begin{equation}\breve{C}(y^{w_1}, \dots, y^{w_d}) = \exp\left(\clamp_{\log y, \max(\mathbf{w}) \log y}\log\hat C(y^{w_1}, \dots, y^{w_d})\right),
\end{equation}
and
\begin{align}
\widehat{A}_{\text{BDV,MM,}h}(\mathbf{w}) &= B_h^{-1} \int_{0}^1 \frac{\log \breve C(y^{w_1}, \dots, y^{w_d})}{\log y} h^*(y)\, dy \\
&= B_h^{-1} \sum_{i=0}^B \int_{\Gamma_{w,i}}^{\Gamma_{w,i+1}} \frac{\log \breve C(y^{w_1}, \dots, y^{w_d})}{\log y} h^*(y)\, dy
\end{align}
Letting $\Gamma_{w,i}^{(\ell)} = \clamp_{\Gamma_{w,i},\Gamma_{w,i+1}}\left(\left(\frac{i}{n}\right)^{\frac1{\max(\mathbf{w})}}\right)$, $\Gamma_{w,i}^{(u)} = \clamp_{\Gamma_{w,i},\Gamma_{w,i+1}}\left(\frac{i}{n}\right)$ for $i\in \{0,\dots,B\}$ and $\eta(x)= B_h^{-1}\int_0^x h^*(y)\,dy$, we have
\begin{align*}
\widehat{A}_{\text{BDV,MM,}h}(\mathbf{w}) &= B_h^{-1} \sum_{i=0}^B \int_{\Gamma_{w,i}}^{\Gamma_{w,i+1}} \frac{\log \breve C(y^{w_1}, \dots, y^{w_d})}{\log y} h^*(y)\, dy\\
&= B_h^{-1} \sum_{i=0}^B \int_{\Gamma_{w,i}}^{\Gamma_{w,i}^{(\ell)}} \max(\mathbf{w}) h^*(y)\, dy + \int_{\Gamma_{w,i}^{(\ell)}}^{\Gamma_{w,i}^{(u)}} \log \frac{i}{n} \frac{h^*(y)}{\log y}\, dy +  \int_{\Gamma_{w,i}^{(u)}}^{\Gamma_{w,i+1}} h^*(y)\, dy \\
&= \sum_{i=0}^B \max(\mathbf{w})\left(\eta(\Gamma_{w,i}^{(\ell)}) - \eta(\Gamma_{w,i})\right) - \log\frac{i}n \left(g(\Gamma_{w,i}^{(u)})-g(\Gamma_{w,i}^{(\ell)})\right) + \eta(\Gamma_{w,i+1}) \\ & - \eta(\Gamma_{w,i}^{(u)})
\end{align*}
In our main submission, we use $\widehat{A}_{\text{BDV,MM,}h}$  with $h(y)=\frac{1}{\log(y)}$.

\section{Further Details on Experiments}
\label{sec:exp_details}
\subsubsection*{Architecture Details}
For learning the Pickands dependence function, in all experiments in the manuscript we used 512 width and 1 depth $d$MNNs.
Only the input layer was changed according to the input dimension.
In order to force the weights to be positive, we use a weight clipping during training.

For the generative model experiments, we model $p_z$ as a 128 d Gaussian random variable. 
The generator is a basic multi layer perceptron (MLP) with ReLU activations and batch norm. 
For all experiments, we use a width 256 and depth 2 MLP for the generator. 
The output is ensured to be positive through a final ReLU operation.

\subsubsection*{Hyperparameter Tuning}
For learning the Pickands dependence experiments, we used the Adam \cite{adam} optimizer for optimizing all parameters with learning rate $1 \times 10^{-2}$ with a decay according to the \texttt{ReduceLROnPlateau} decay algorithm with a patience of 100 epochs. 
Each model was trained for 2000 epochs for the survival experiments and 4000 for the sampling experiments.
For the sampling experiments, the generator was trained using Adam with learning rate $1 \times 10^{-3}$, $\beta_1 = 0.5$ and $\beta_2 = 0.99$ with exponential decay on the learning rate of $0.99998$.
Models for the generator were trained for 4000 epochs

\subsubsection*{Computational Resources}
All experiments were run on an Nvidia RTX Titan GPU with an Intel Core i9-7900X CPU @ 3.30GHz and 64 GB of RAM. 

\section{Larger Figures}
\label{sec:large_figs}

\begin{figure}[h!]
    \centering
\begin{subfigure}{.24\textwidth}
  \centering
  \includegraphics[width=\linewidth]{imgs/margins/commodities_3d_NaiveEstimator.pdf}  
  \caption{Pickands 3d Margins}
\end{subfigure}
\begin{subfigure}{.24\textwidth}
  \centering
  \includegraphics[width=\linewidth]{imgs/margins/commodities_3d_CFGEstimator.pdf}  
  \caption{CFG 3d Margins}
\end{subfigure}
\begin{subfigure}{.24\textwidth}
  \centering
  \includegraphics[width=\linewidth, trim=2pt 0pt 2pt 0pt]{imgs/margins/commodities_3d_BDVEstimatorMM.pdf}  
  \caption{BDV 3d Margins}
\end{subfigure}
\begin{subfigure}{.24\textwidth}
  \centering
  \includegraphics[width=\linewidth]{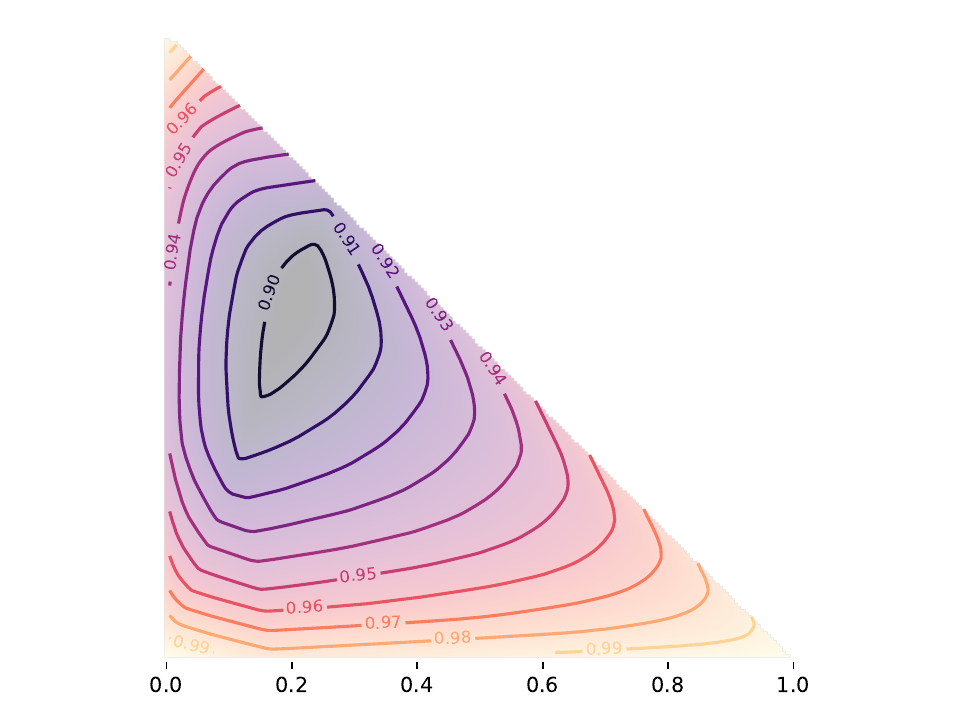}  
  \caption{$d$MNN 3d Margins}
\end{subfigure}
\caption{(Larger figures from main text) Qualitative comparison of 3d margins from learned 10d MEV for the commodities dataset. The $d$MNN is the method that retains margins that are valid Pickands dependence functions as the others are non-convex and outside the required bounds. Contours plotted with solid line.}
\end{figure}

\begin{figure}[h!]
    \centering
\begin{subfigure}{.48\textwidth}
  \centering
  \includegraphics[width=\linewidth]{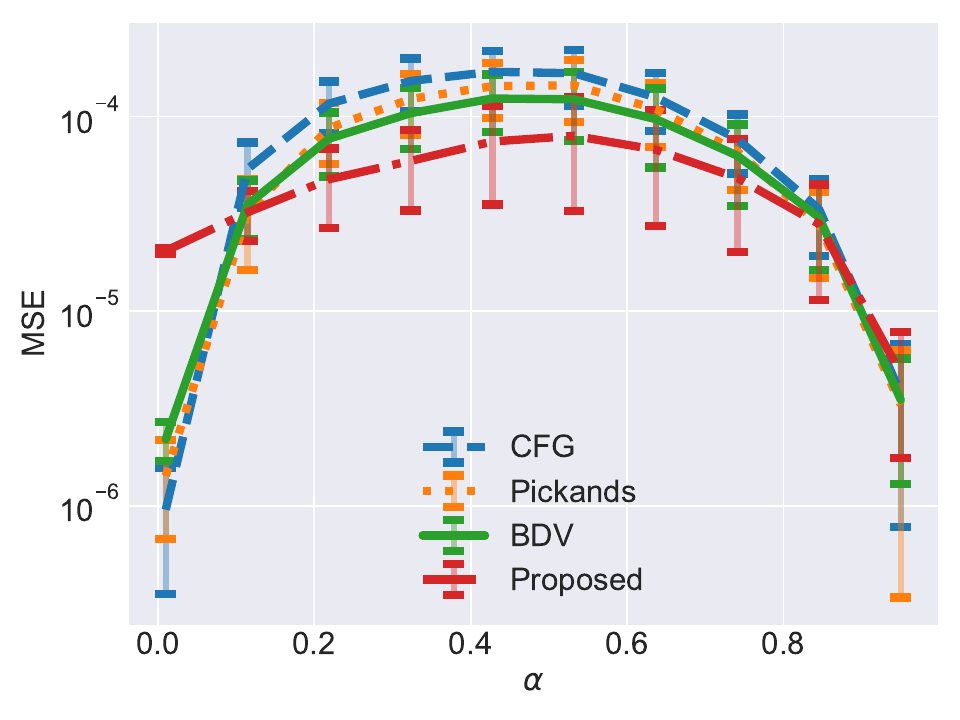} 
  \caption{$A_\text{SL}$ MSE ($d=2$)}
\end{subfigure}
\begin{subfigure}{.48\textwidth}
  \centering
  \includegraphics[width=\linewidth]{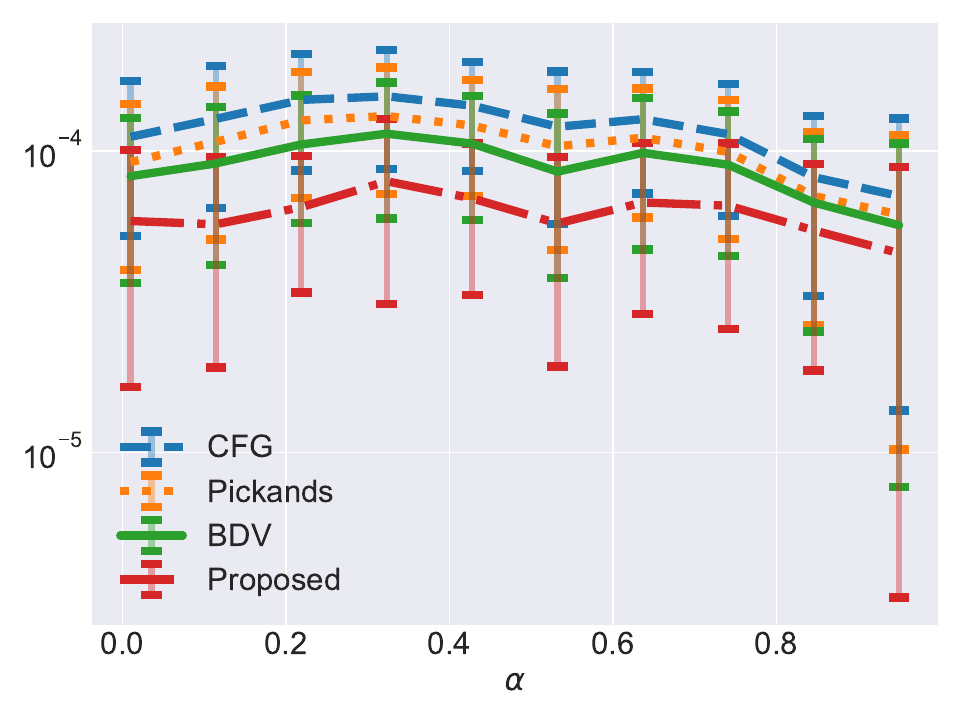}
  \caption{$A_\text{ASL}$ MSE ($d=2$)}
\end{subfigure} 
\caption{(Larger figures from main text) (\ref{fig:sl_survival}, \ref{fig:asl_survival}) MSE of survival probabilities for $d=2$ with $100$ samples for $A_\text{SL}$ (\ref{fig:sl_survival}) and $A_\text{ASL}$ (\ref{fig:asl_survival}). Thresholds are above the $75$th percentile.}
\vspace{-10pt}
\end{figure}

\begin{figure}[h!]
    \centering
\begin{subfigure}{.4\textwidth}
  \centering
  \includegraphics[width=\linewidth]{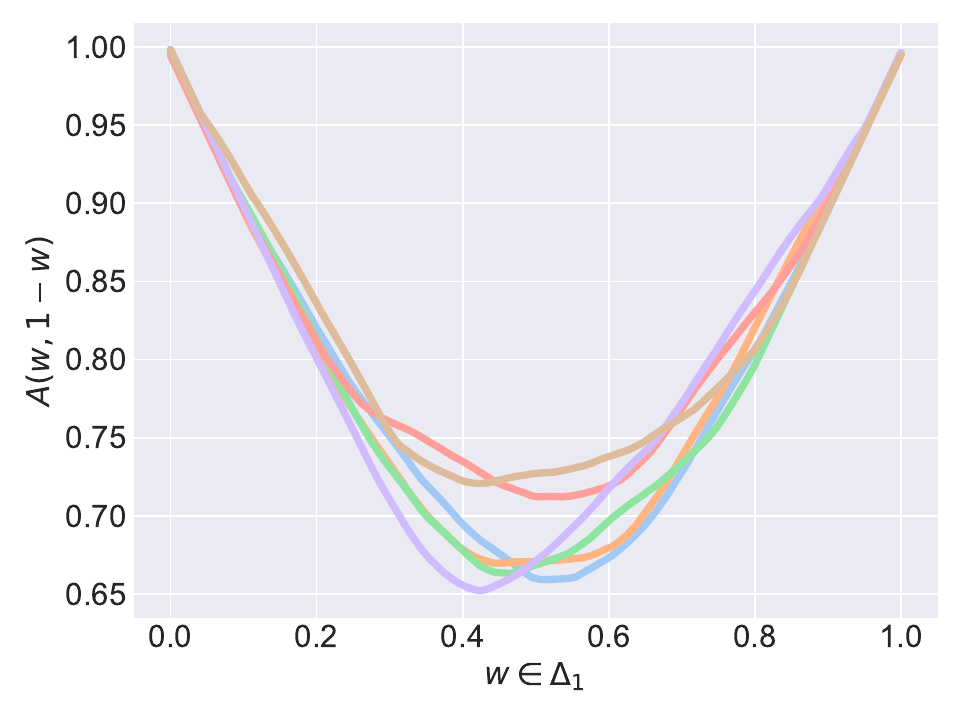}  
  \caption{Pickands 2d Margins}
\end{subfigure}
\begin{subfigure}{.4\textwidth}
  \centering
  \includegraphics[width=\linewidth]{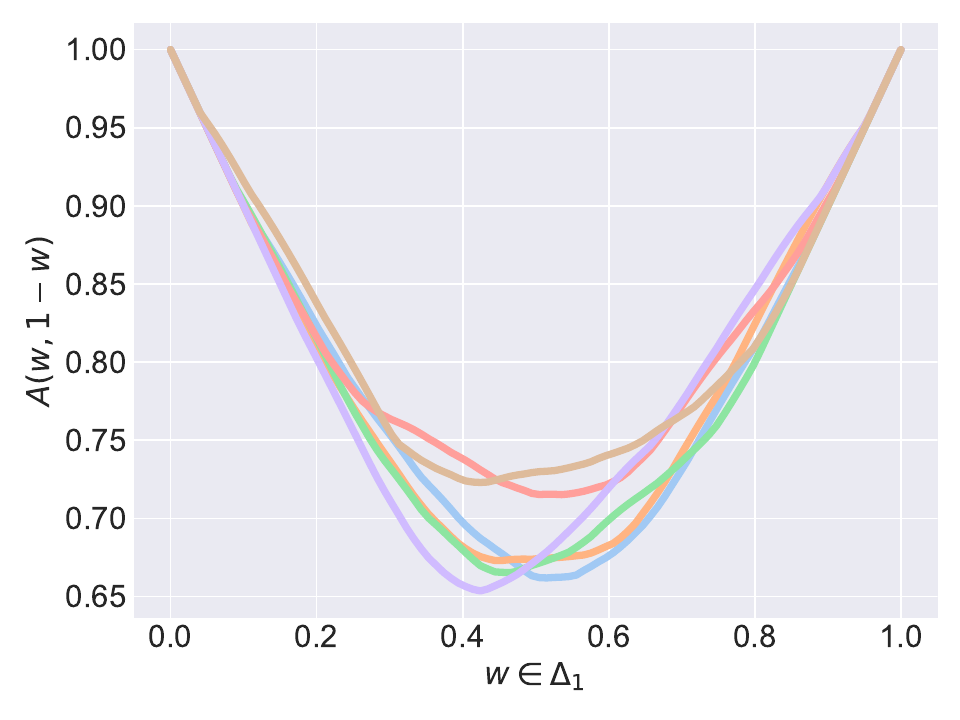}  
  \caption{CFG 2d Margins}
\end{subfigure} 
\begin{subfigure}{.4\textwidth}
  \centering
  \includegraphics[width=\linewidth, trim=2pt 0pt 2pt 0pt]{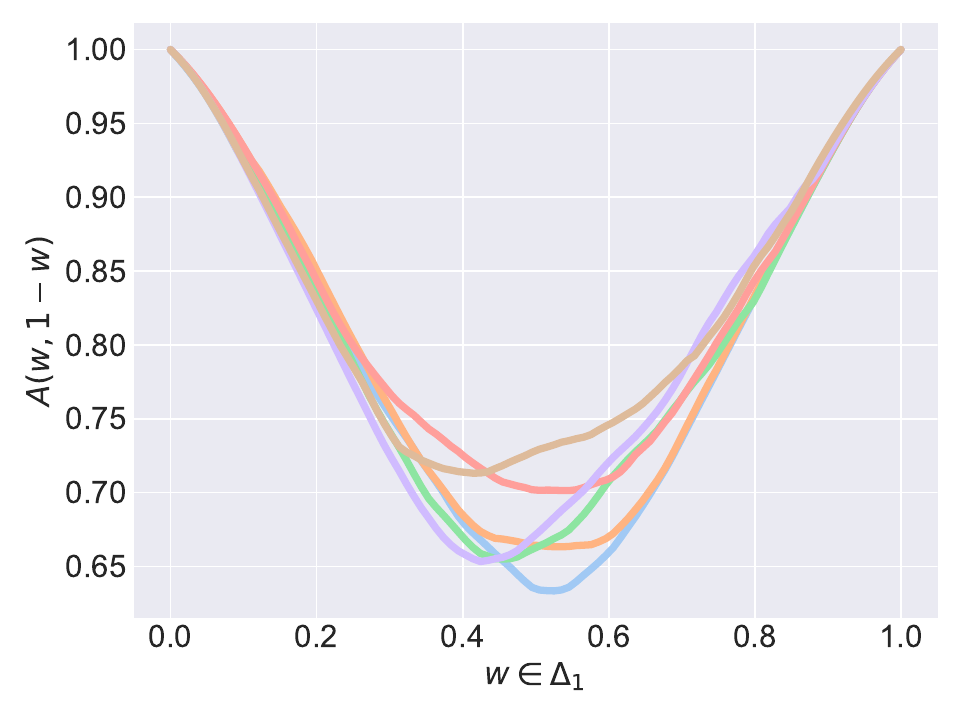}  
  \caption{BDV 2d Margins}
\end{subfigure}
\begin{subfigure}{.4\textwidth}
  \centering
  \includegraphics[width=\linewidth]{imgs/margins/ozone_extremal_MaxLinear_w512d1.pdf}  
  \caption{$d$MNN 2d Margins}
    \label{fig:net_marg_ozone}
\end{subfigure}
\label{fig:marg_ozone}
\caption{(Additional figure.) Qualitative comparison of 6 2d margins from learned 4d MEV for the Ozone dataset. The $d$MNN is the method that retains margins that are valid Pickands dependence functions as the others are non-convex and outside the required bounds.}
\end{figure}

\begin{figure}[h!]
    \centering
\begin{subfigure}{.4\textwidth}
  \centering
  \includegraphics[width=\linewidth]{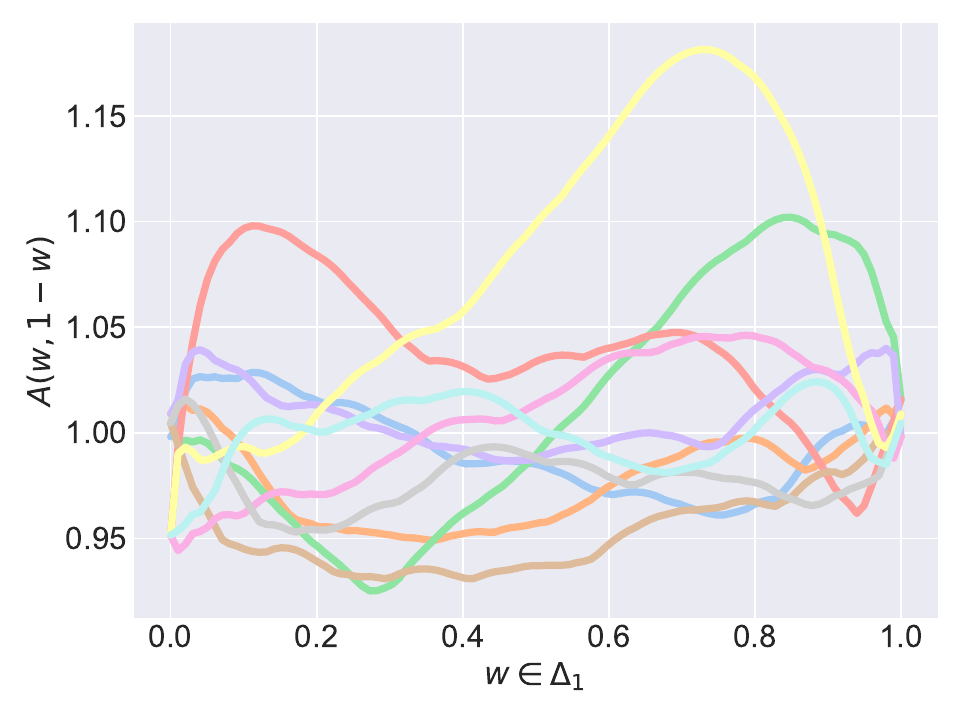}  
  \caption{Pickands 2d Margins}
\end{subfigure}
\begin{subfigure}{.4\textwidth}
  \centering
  \includegraphics[width=\linewidth]{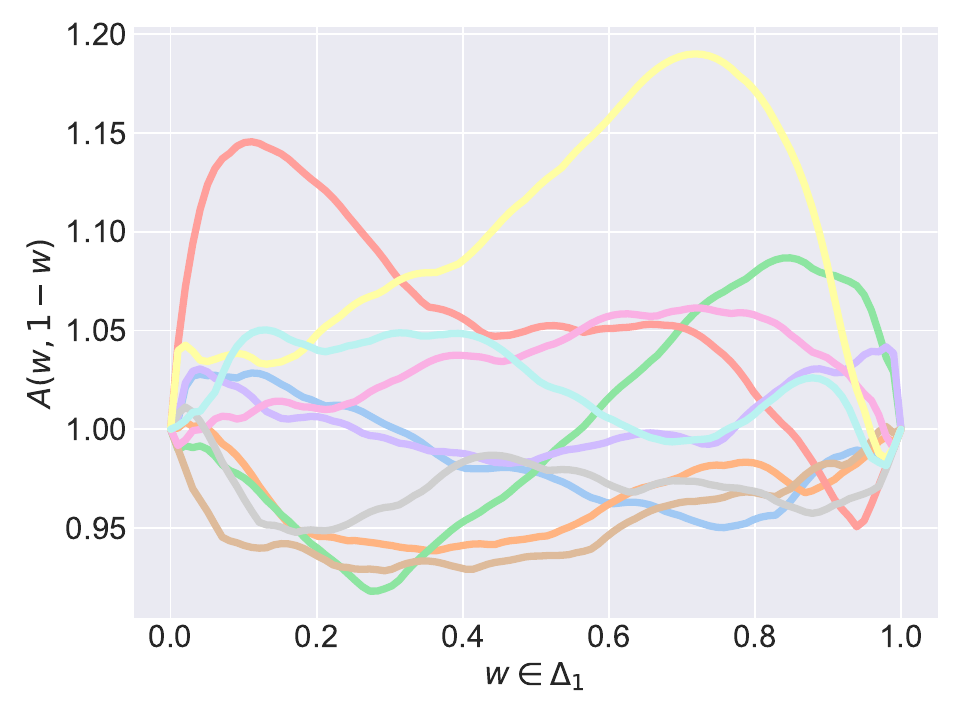}  
  \caption{CFG 2d Margins}
\end{subfigure} 
\begin{subfigure}{.4\textwidth}
  \centering
  \includegraphics[width=\linewidth, trim=2pt 0pt 2pt 0pt]{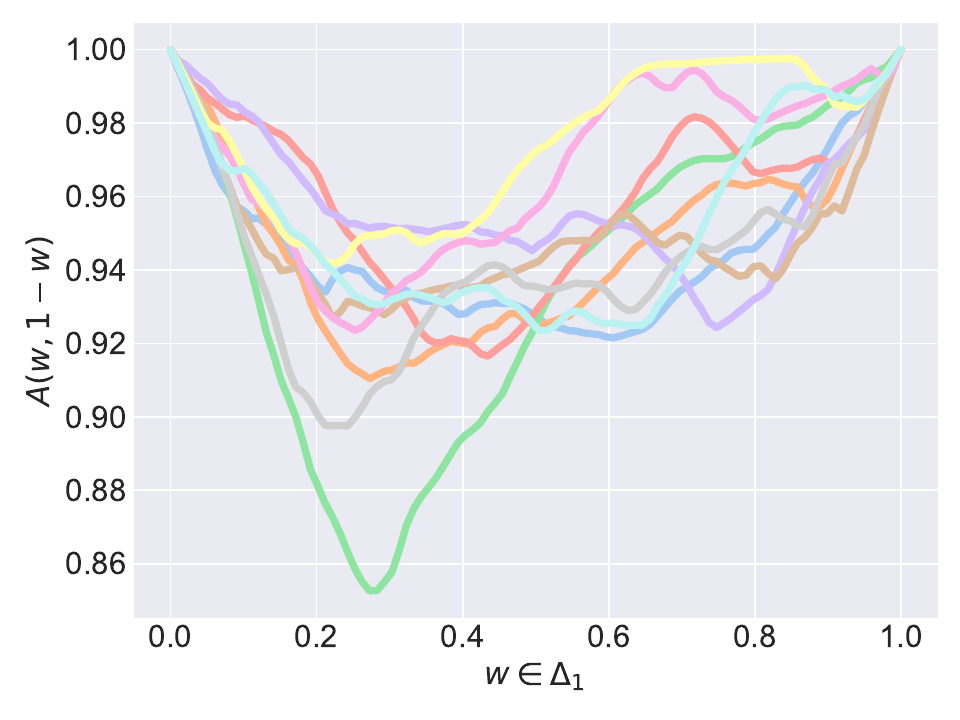}  
  \caption{BDV 2d Margins}
\end{subfigure}
\begin{subfigure}{.4\textwidth}
  \centering
  \includegraphics[width=\linewidth]{imgs/margins/comm_extremal_MaxLinear_w512d1.pdf}  
  \caption{$d$MNN 2d Margins}
\end{subfigure}
\label{fig:marg_comm}
\caption{(Additional figure.) Qualitative comparison of 10 2d margins from learned 10d MEV for the commodities. The $d$MNN is the method that retains margins that are valid Pickands dependence functions as the others are non-convex and outside the required bounds.}
\end{figure}
\begin{figure}[h!]
    \centering
\begin{subfigure}{.4\textwidth}
  \centering
  \includegraphics[width=\linewidth]{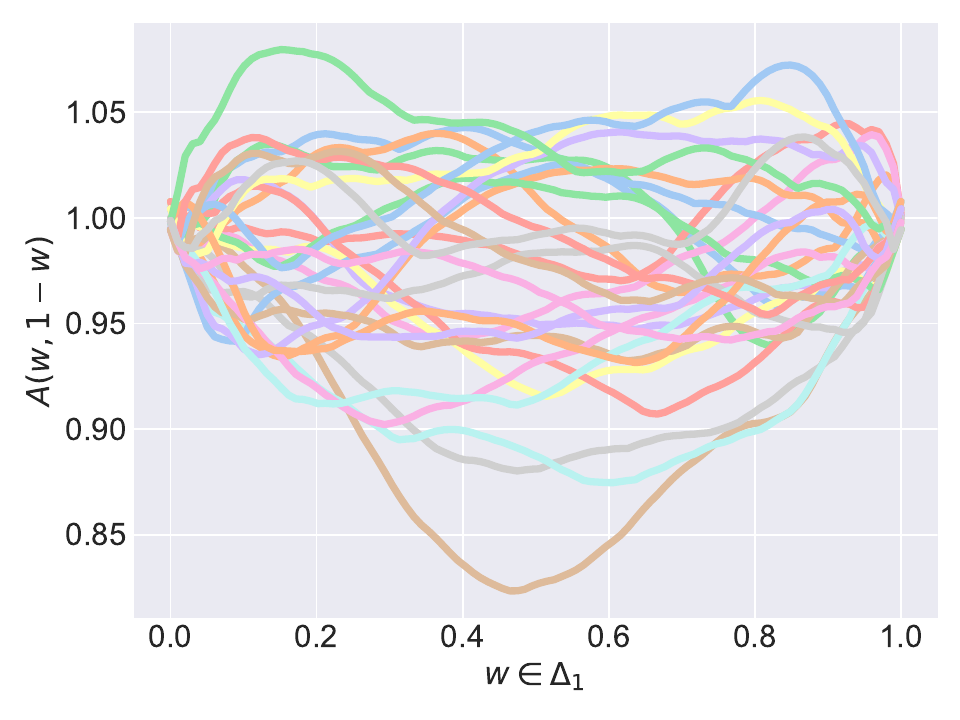}  
  \caption{Pickands 2d Margins}
\end{subfigure}
\begin{subfigure}{.4\textwidth}
  \centering
  \includegraphics[width=\linewidth]{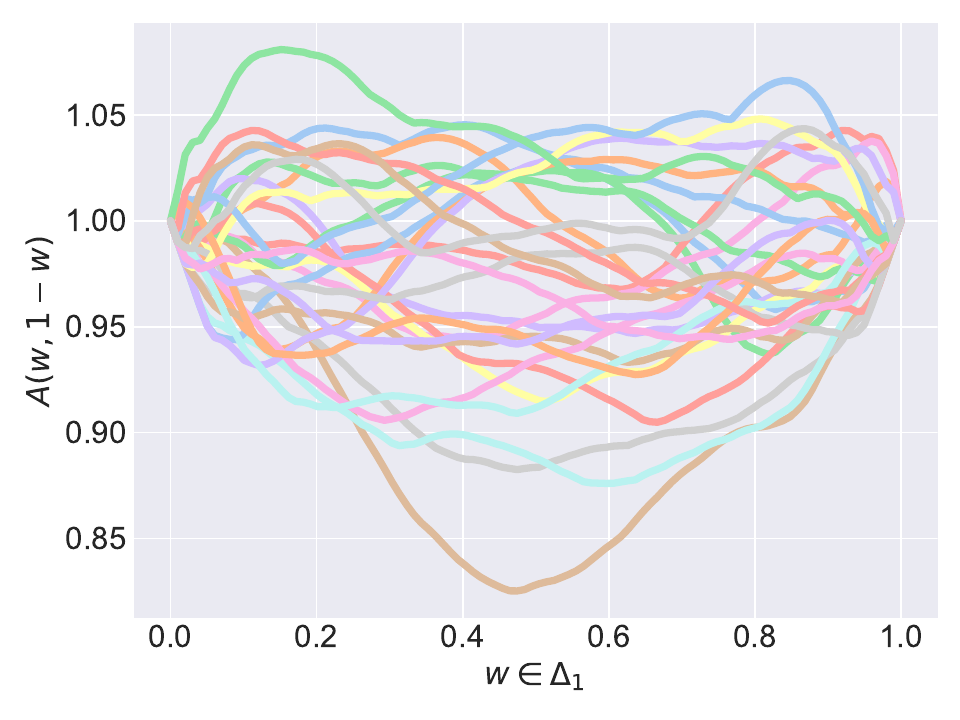}  
  \caption{CFG 2d Margins}
\end{subfigure} 
\begin{subfigure}{.4\textwidth}
  \centering
  \includegraphics[width=\linewidth, trim=2pt 0pt 2pt 0pt]{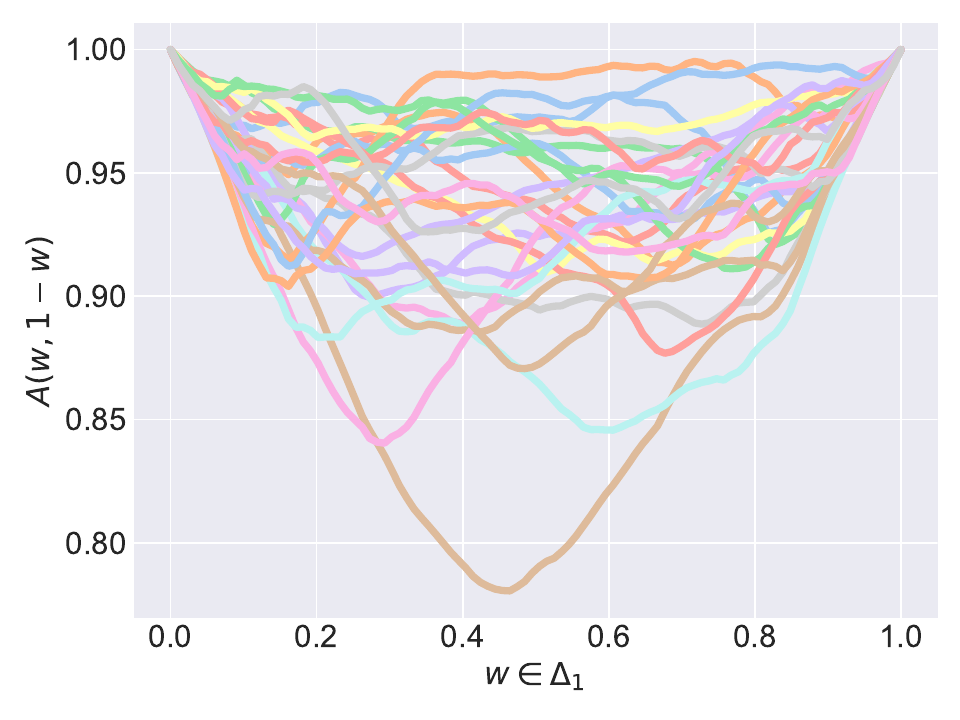}  
  \caption{BDV 2d Margins}
\end{subfigure}
\begin{subfigure}{.4\textwidth}
  \centering
  \includegraphics[width=\linewidth]{imgs/margins/spy_extremal_MaxLinear_w512d1.pdf}  
  \caption{$d$MNN 2d Margins}
    \label{fig:net_marg_comm}
\end{subfigure}
\label{fig:marg_spy}
\caption{(Additional figure.) Qualitative comparison of 28 2d margins from learned 418d MEV for the S\&P dataset. The $d$MNN is the method that retains margins that are valid Pickands dependence functions as the others are non-convex and outside the required bounds.}
\end{figure}

\begin{figure}[h!]
    \centering
\begin{subfigure}{.4\textwidth}
  \centering
  \includegraphics[width=\linewidth]{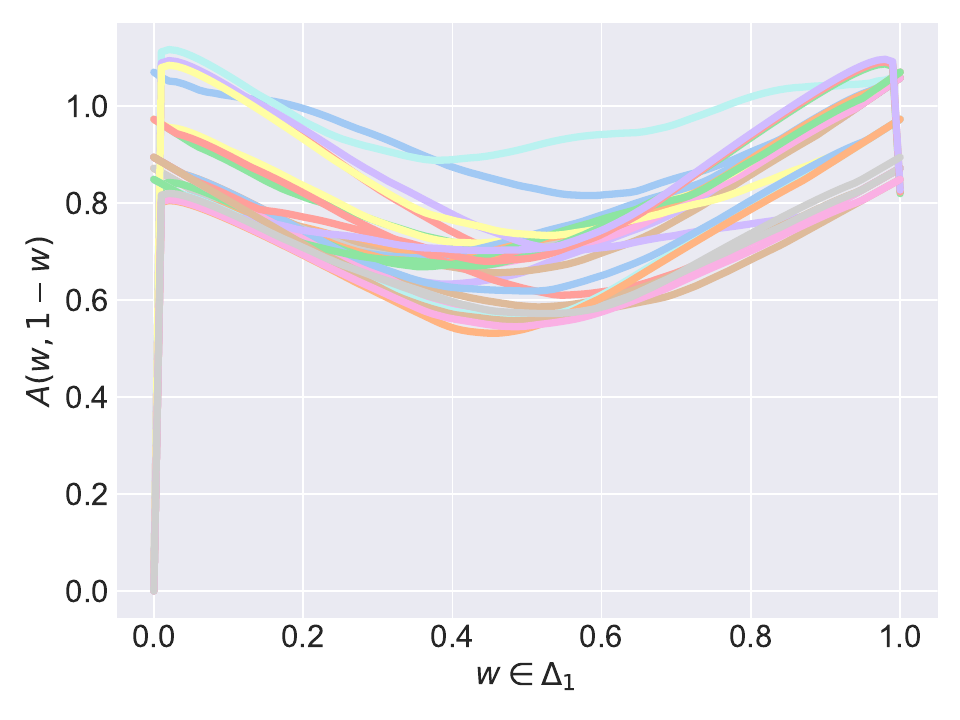}  
  \caption{Pickands 2d Margins}
\end{subfigure}
\begin{subfigure}{.4\textwidth}
  \centering
  \includegraphics[width=\linewidth]{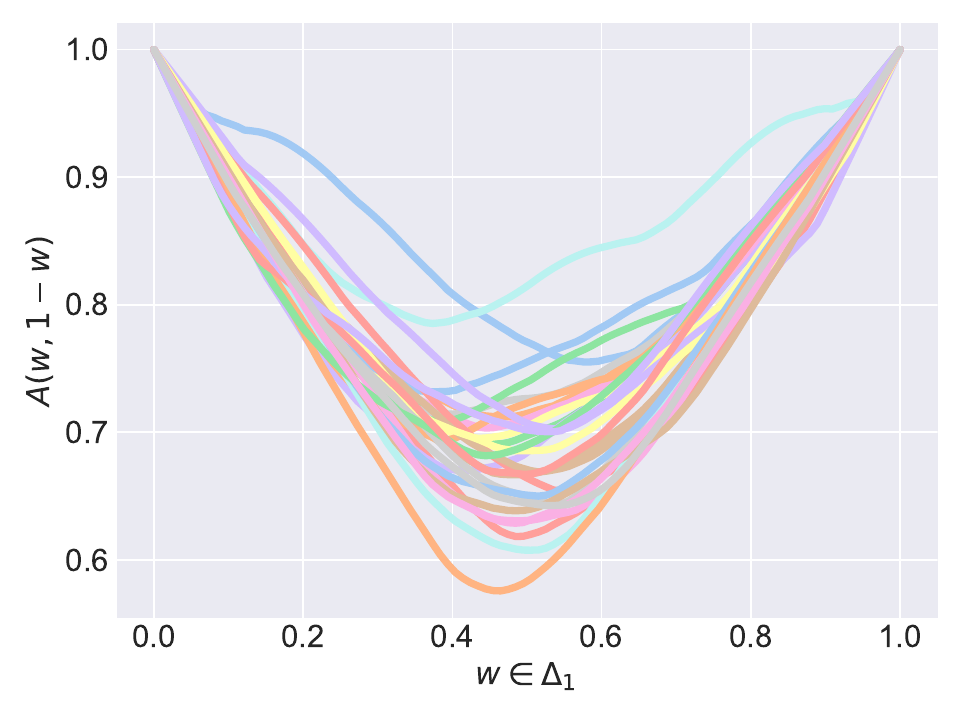}  
  \caption{CFG 2d Margins}
\end{subfigure} 
\begin{subfigure}{.4\textwidth}
  \centering
  \includegraphics[width=\linewidth, trim=2pt 0pt 2pt 0pt]{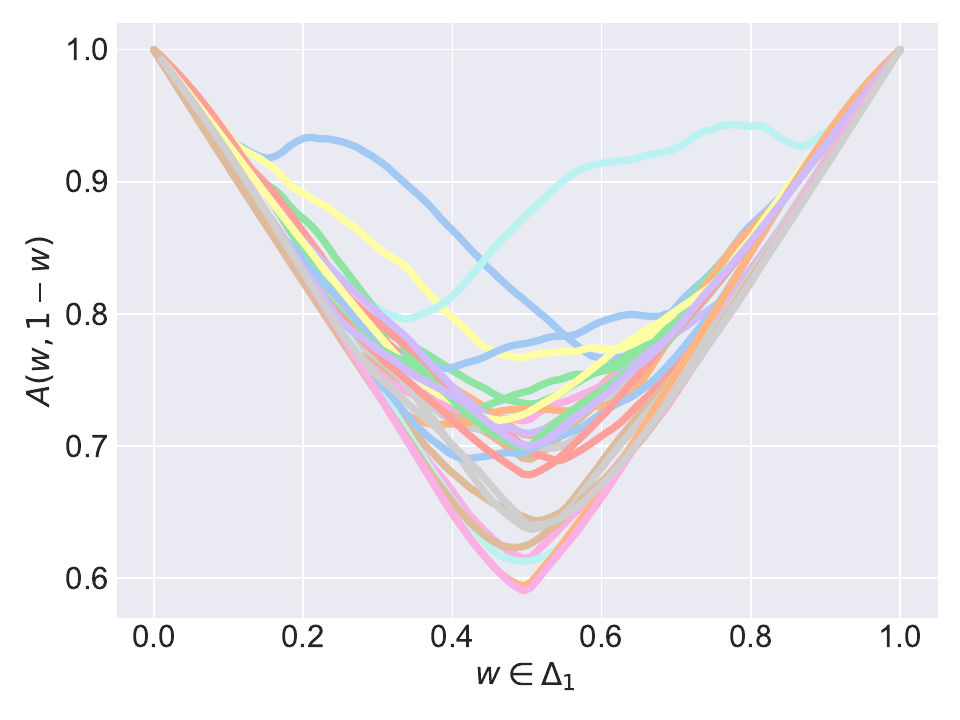}  
  \caption{BDV 2d Margins}
\end{subfigure}
\begin{subfigure}{.4\textwidth}
  \centering
  \includegraphics[width=\linewidth]{imgs/margins/crypto_extremal_MaxLinear_w512d1.pdf}  
  \caption{$d$MNN 2d Margins}
    \label{fig:net_marg_crypto}
\end{subfigure}
\label{fig:marg_crypto}
\caption{(Additional figure.) Qualitative comparison of 28 2d margins from learned 100d MEV for the Crypto dataset. The $d$MNN is the method that retains margins that are valid Pickands dependence functions as the others are non-convex and outside the required bounds. }
\end{figure}
\begin{figure}[h!]
    \centering
\begin{subfigure}{.4\textwidth}
  \centering
  \includegraphics[width=\linewidth]{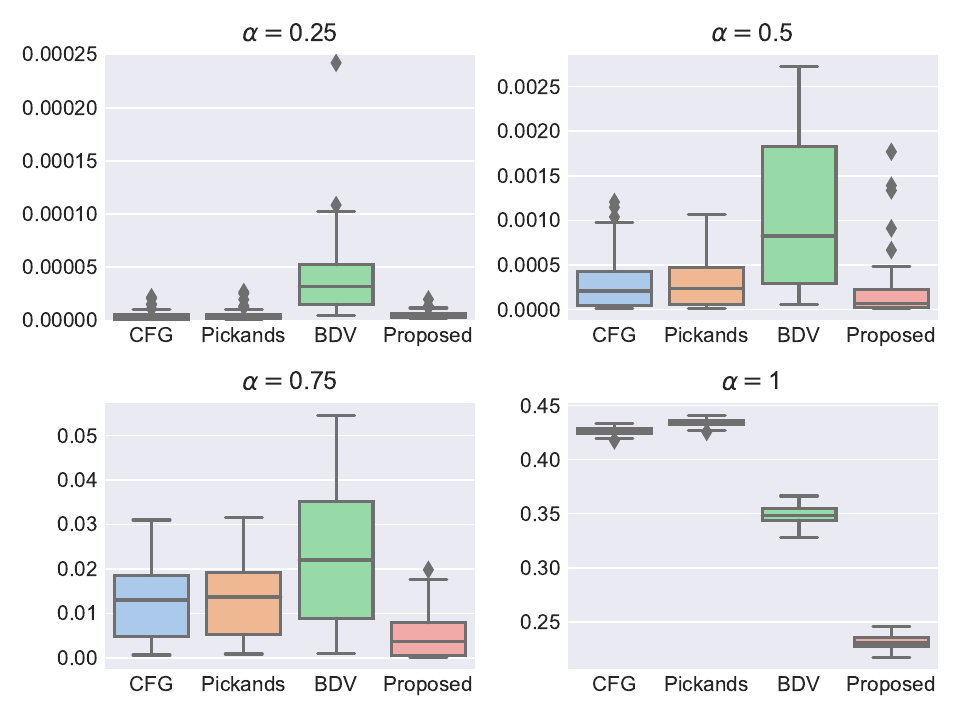} 
  \caption{$A_\text{SL}$ MSE ($d=256$)}
\end{subfigure}
\begin{subfigure}{.4\textwidth}
  \centering
  \includegraphics[width=\linewidth]{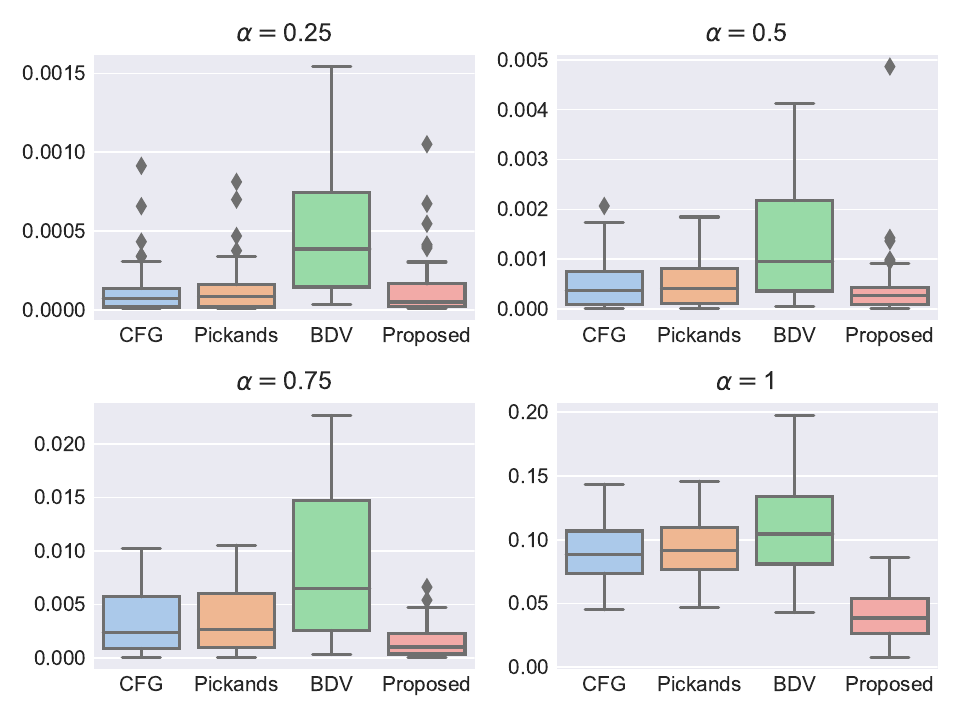}
  \caption{$A_\text{ASL}$ MSE ($d=256$)}
\end{subfigure} 
\begin{subfigure}{.4\textwidth}
  \centering
\includegraphics[width=\linewidth]{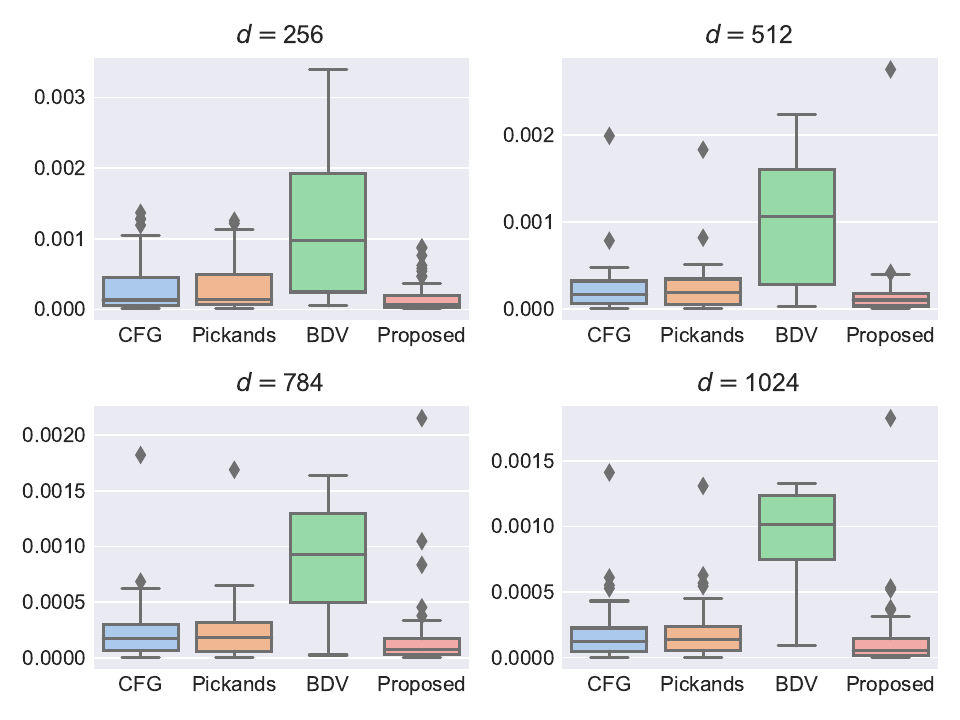}  
  \caption{$A_\text{SL}$ MSE ($\alpha=0.5$)}
\end{subfigure}
\begin{subfigure}{.4\textwidth}
  \centering
    \includegraphics[width=\linewidth]{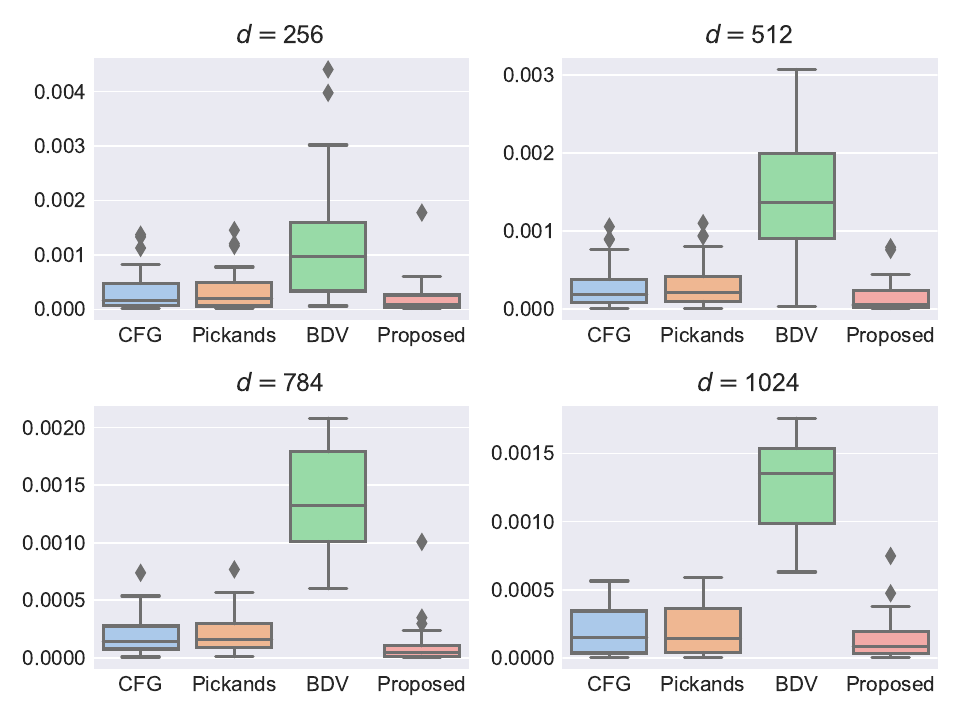}  
  \caption{$A_\text{ASL}$ MSE ($\alpha=0.5$)}
\end{subfigure}  

\caption{(Larger figures from main text) Comparison of $||\hat{A}(\mathbf{w}) - A(\mathbf{w})||_2^2$ for different estimators $\hat{A}$ for different dependence $\alpha = \{0.25, 0.50, 0.75, 1.0\}$ and $d=256$ (\ref{fig:sl_mse_est_all_a}, \ref{fig:asl_mse_est_all_a}) and for fixed  $\alpha=0.5$ for $d = \{256, 512, 728, 1024\}$ (\ref{fig:sl_mse_est_all_d}, \ref{fig:asl_mse_est_all_d})  for $A_\text{SL}$ (\ref{fig:sl_mse_est_all_a}, \ref{fig:sl_mse_est_all_d}) and $A_\text{ASL}$ (\ref{fig:asl_mse_est_all_a}, \ref{fig:asl_mse_est_all_d}). Results are over 50 runs with 100 training samples for each run.} 
    \vspace{-10pt}
\end{figure}
\begin{figure}[h!]
    \centering
\begin{subfigure}{.4\textwidth}
  \centering
  \includegraphics[width=\linewidth]{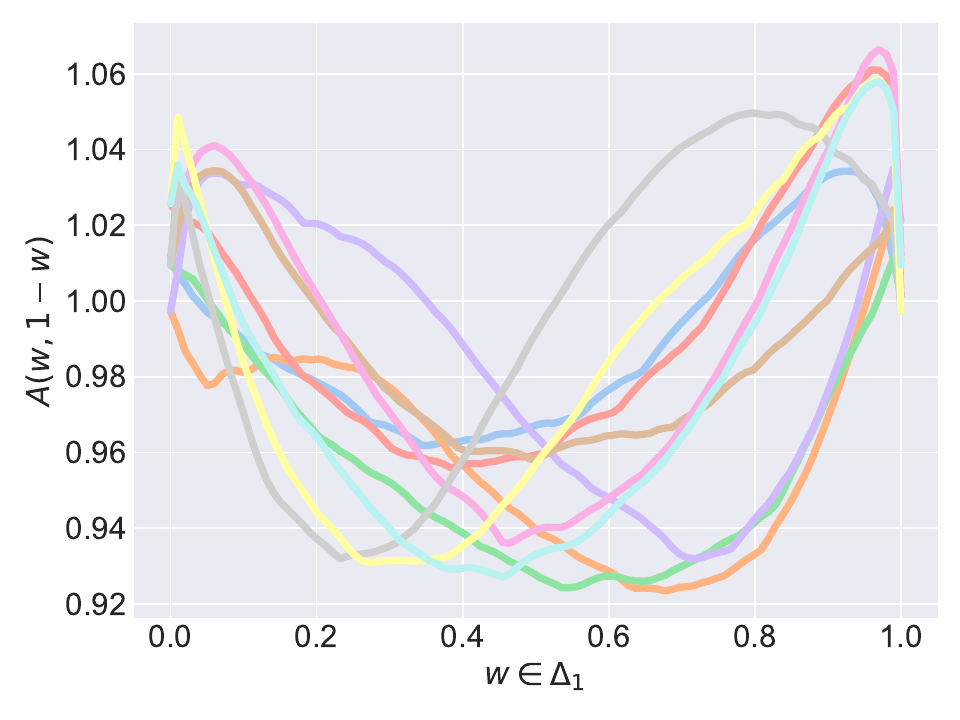}  
  \caption{Pickands 2d Margins}
\end{subfigure}
\begin{subfigure}{.4\textwidth}
  \centering
  \includegraphics[width=\linewidth]{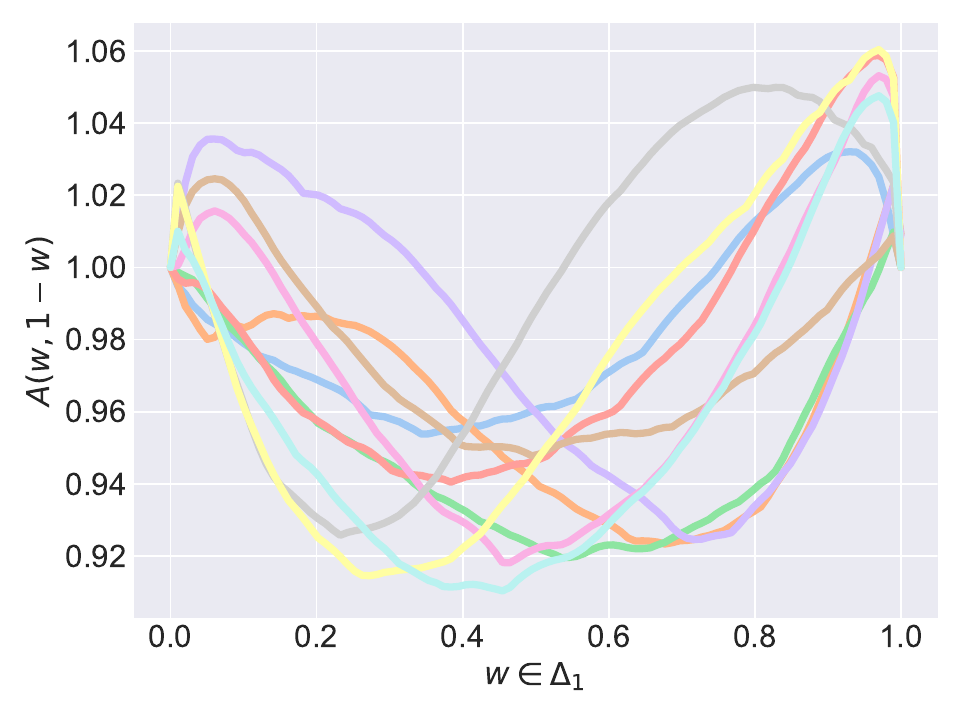}  
  \caption{CFG 2d Margins}
\end{subfigure}
\begin{subfigure}{.4\textwidth}
  \centering
  \includegraphics[width=\linewidth]{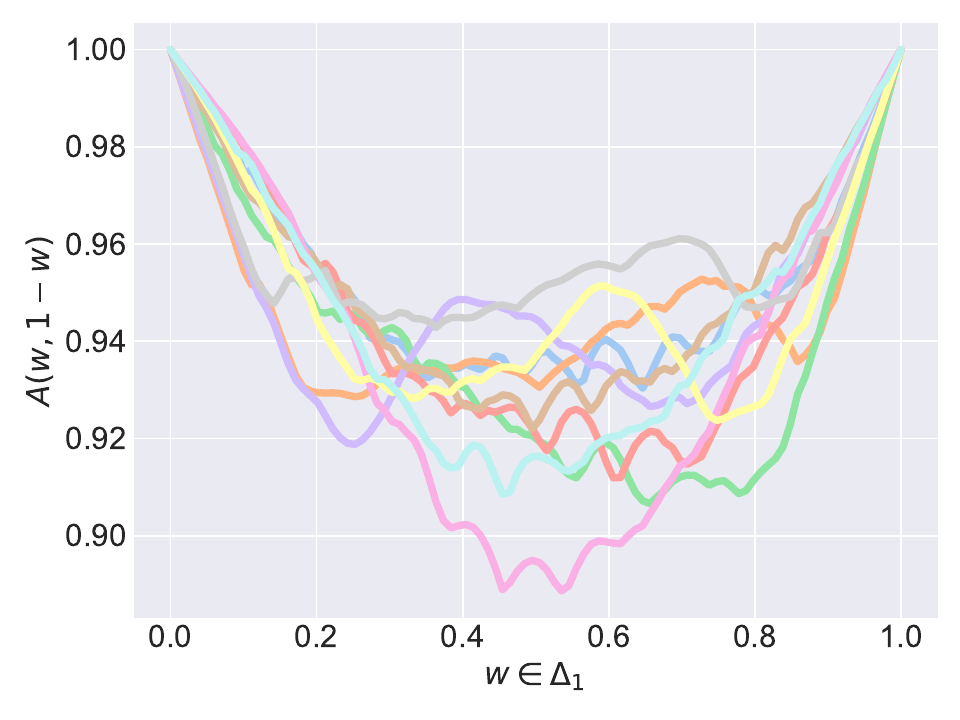}  
  \caption{BDV 2d Margins}
\end{subfigure}
\begin{subfigure}{.4\textwidth}
  \centering
  \includegraphics[width=\linewidth]{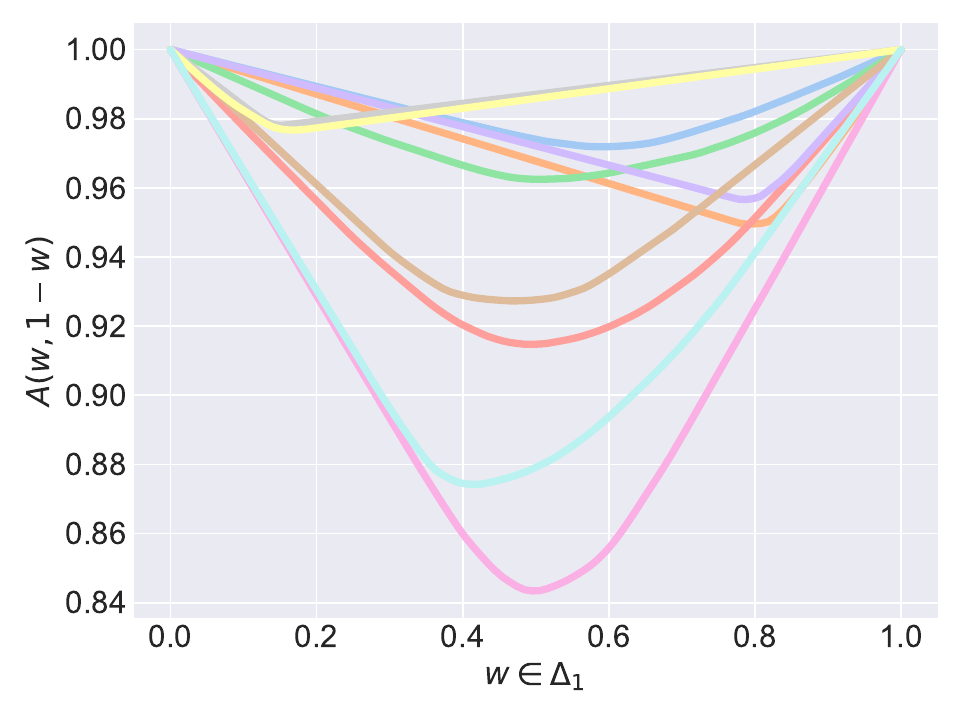}  
  \caption{$d$MNN 2d Margins}
\end{subfigure}
\caption{(Larger figures from main text) Qualitative comparison of 10 out of 45 total 2d margins from learned 10d MEV for the California Winds dataset. The $d$MNN is the only method that retains margins that are valid Pickands dependence functions. }
\vspace{-10pt}
\end{figure}

\begin{figure}[h!]
    \centering
\begin{subfigure}{.4\textwidth}
  \centering
  \includegraphics[width=\linewidth]{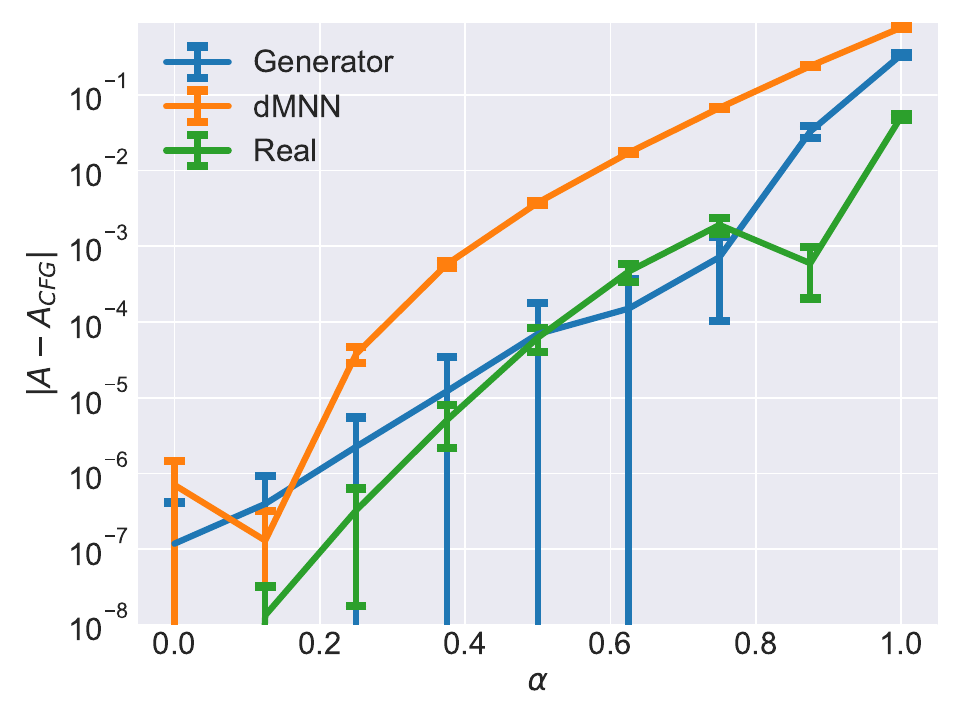}  

  \caption{SL CFG MSE $\Delta \alpha$}
\end{subfigure}
\begin{subfigure}{.4\textwidth}
  \centering
    \includegraphics[width=\linewidth]{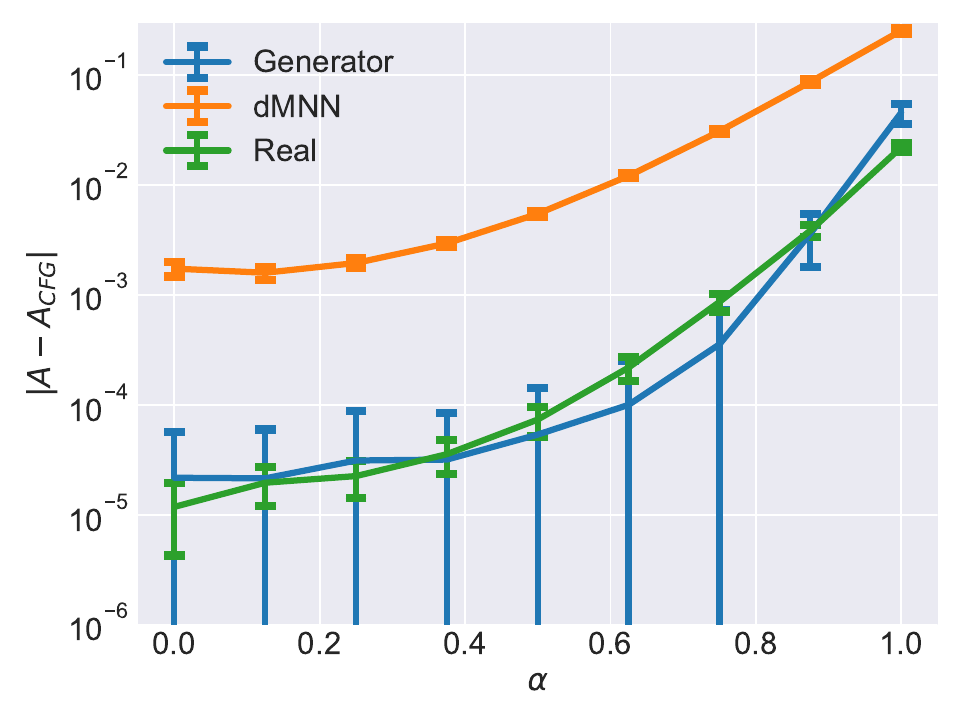}  
  \caption{ASL CFG MSE $\Delta \alpha$}
\end{subfigure}
\begin{subfigure}{.4\textwidth}
  \centering
  \includegraphics[width=\linewidth]{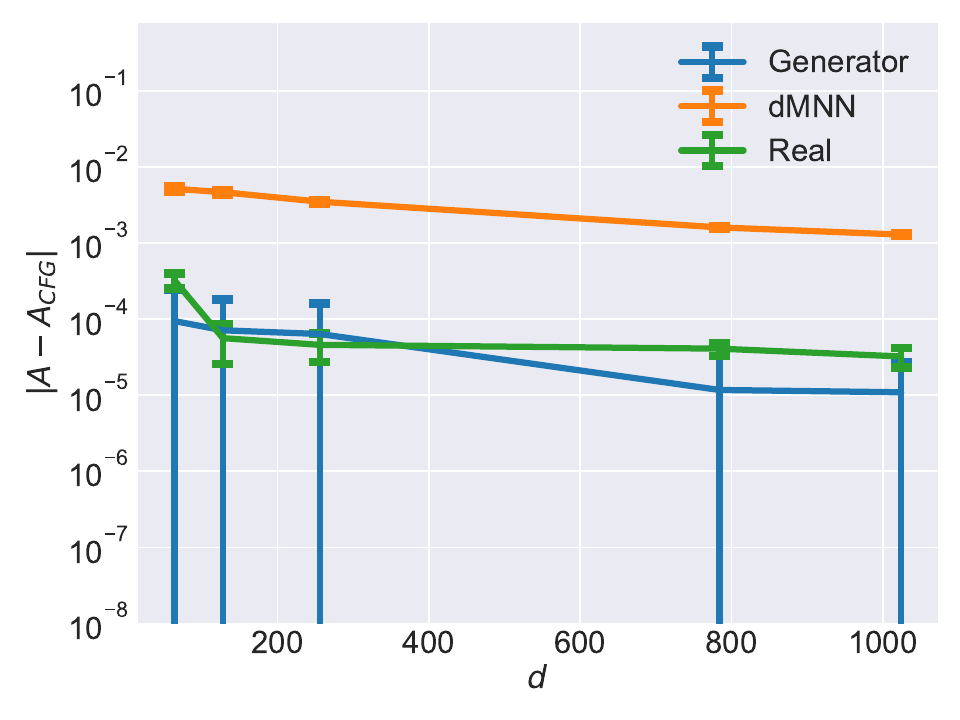}  
  \caption{SL CFG MSE $\Delta d$}
\end{subfigure}
\begin{subfigure}{.4\textwidth}
  \centering
  \includegraphics[width=\linewidth]{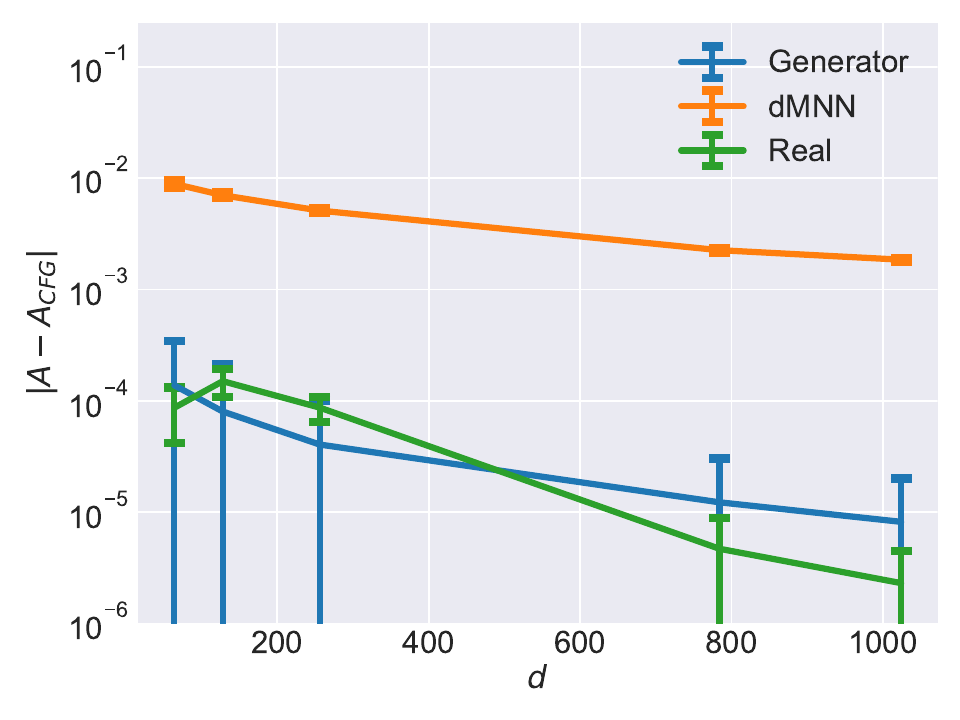}  
  \caption{ASL CFG MSE $\Delta d$}
\end{subfigure}
\caption{(Larger figures from main text) MSE of CFG estimate for 1000 samples and 1000 simplex points for $d=225$ (\ref{fig:sl_mse_gen}, \ref{fig:asl_mse_gen}) at various $\alpha \in (0,1)$ and $\alpha=0.5$ (\ref{fig:sl_mse_gen_all_d}, \ref{fig:asl_mse_gen_all_d}) at $d=\{64, 128, 256, 784, 1024\}$ for $A_\text{SL}$ (\ref{fig:sl_mse_gen}) and  $A_\text{ASL}$ (\ref{fig:asl_mse_gen}) for data sampled from generative model (blue), $d$MNN (orange), and exact sampled (green). Both models were trained with 1000 data points.}
\vspace{-5pt}
\end{figure}
\newpage
\section{Algorithms}
\label{sec:algs}
Here we provide algorithms for the estimation and sampling presented in the main content. 

\begin{algorithm}[h!]
	\caption{Fitting the Pickands-$d$MNN to Data} 
	\begin{algorithmic}[1]
	\STATE \textbf{Input:} $\left \{ \left(X_1^{(i)}, \ldots, X_d^{(i)} \right) \right \}_{i=1}^N$, $N=B \times n$ samples of i.i.d. random vectors where $B$ is the number of blocks of data and $n$ is the size of each block.
	\STATE Take component-wise maxima over each block: $\left \{ \left(M_{1}^{(n,b)}, \ldots, M_{d}^{(n,b)} \right)\right \}_{b=1}^B$ where $M_{k}^{(n,b)}=\max _{i=(b-1)n+1,...,bn}X_k^{(i)}$, $(k, b) \in \{1, \ldots, d\} \times \{1, \ldots, B\}$. 
	\STATE Fit a GEV to each component-wise maxima $\{ M_{k}^{(n,b)} \}_{b=1}^B$, obtain $\{\bar{M}_{k}^{(n,b)} \}_{b=1}^B$, then estimate marginals $F_k$ for each $k \in \{1, \ldots, d \}$.
	\STATE \textbf{Initialize} the parameters ${\bm \theta} \geq 0$ of the $d$MNN \\ 
    \textbf{Repeat}: 
    \STATE Randomly sample a minibatch of training data $\{\bar{M}_{k}^{(n,b)} \}_{b \in \text{batch}}$
    and uniformly sample $\mathbf{w} \in \Delta_{d-1}$.
    \STATE  Transform samples according to Equations \eqref{transform1} and \eqref{transform2} to obtain transformed samples $\{Z_{w, b} \}_{b \in \text{batch}}$. 
    \STATE Compute gradient
     $\nabla_{{\bm\theta}}  \sum_{b \in \text{batch}} \mathcal{L}\left(Z_{w, b}; {\bm \theta} \right)$.\\
    \STATE Update $\bm \theta$ with Adam \citep{adam} \\ 
    \textbf{Until} convergence \\ 
    \textbf{Output:} $A^{\star}_{\bm \theta}(\mathbf{w})$. 
	\end{algorithmic} 
\end{algorithm}

\begin{algorithm}[h!]
	\caption{Estimating survival probabilities with the Pickands dependence function} 
	 \label{alg_survival}
	\begin{algorithmic}[1]
	\STATE \textbf{Input:} $\{\bar{M}_{n, b}^{(k)} \}_{b=1}^B$, thresholds: $\left(\gamma_1, \cdots, \gamma_d \right)$. \\ 
	\STATE Train a model $A(\mathbf{w}; \theta)$ with the transformed variables $\{(G_1(\bar{M}_{n, b}^{(1)},\dots,G_d(\bar{M}_{n, b}^{(d)})) \}_{b=1}^B$ using Algorithm \ref{alg_train} and obtain  $A(\mathbf{w}; \theta_*)$.
	\STATE Evaluate the Pickands copula:
	\begin{align*}
	    C\left(1- F_1(\bar{\gamma}_1), \cdots,  1- F_d(\bar{\gamma}_d) \right),
	\end{align*}
	where $C$ is calculated as in Equation \eqref{eq:pickands_copula} with $A = A(\mathbf{w}; \theta_*)$. 
	\end{algorithmic}  
\end{algorithm}

\begin{algorithm}[h!]
	\caption{Training a Generator for a Pickands Copula} 
	 \label{alg:train_gen}
	\begin{algorithmic}[1]
	\STATE \textbf{Input:} $A(\mathbf{w})$, $p_z$, tolerance parameter $\epsilon$ \\ 
	\STATE Initialize parameters $\phi$ of generator $G( \cdot ; \phi)$ \\
	\STATE Sample $\{ \mathbf{w}^{(j)}\}_{j=1}^{N_\text{simplex}}$ samples uniformly over $\Delta_{d-1}$.\\
	\WHILE{$ \sum_{j =1}^{N_\text{simplex}} \mathcal{L}(\mathbf{w}^{(j)}; \phi) > \epsilon$}
	\STATE Sample $\{\mathbf{w}^{(j)}\}_{j=1}^{N_\text{simplex}}$ samples uniformly over $\Delta_{d-1}$. \\
	\STATE Sample $\{\mathbf{y}^{(i)}\}_{i=1}^{N_\text{gen}}$ where $\mathbf{y}^{(i)} = G(\mathbf{z}^{(i)}; \phi), \mathbf{z}^{(i)} \sim p_z$ for $1 \leq i \leq N_{\text{gen}}$. \\
	\STATE Define $\eta(\mathbf{w}, \mathbf{y}) = \max \{ \mathbf{w} \odot \mathbf{y} \}$ with $\odot$ denoting the point-wise multiplication. \\
	\STATE Compute gradient w.r.t $\phi$ of $\sum_{j=1}^{N_{\text{simplex}}}\mathcal{L}(\mathbf{w}^{(j)}; \phi)$ where:
	\vspace{-10pt}
	\begin{equation*}
	    \mathcal{L}(\mathbf{w}^{(j)}; \phi) = \left \|A(\mathbf{w}^{(j)}) - \frac{1}{N_\text{gen}} \sum_{i=1}^{N_\text{gen}} \eta(\mathbf{w}^{(j)}, \mathbf{y}^{(i)}) \right \|_2^2  + \| \frac{1}{N_\text{gen}}\sum_{i=1}^{N_\text{gen}}\mathbf{y}^{(i)} - 1 \|_2^2
	\end{equation*}
	\STATE Update $\phi$ using Adam \cite{adam}.
	\ENDWHILE
	\STATE \textbf{Output:} $G(. ; \phi_*)$.
	\end{algorithmic}  
\end{algorithm}

\begin{algorithm}[h!]
	\caption{Heuristic for Sampling From a Given Pickands Copula \citep[Algorithm 1]{hofert2018hierarchical}}
	 \label{alg:sampling}
	\begin{algorithmic}
	\STATE \textbf{Input:} $A(\mathbf{w})$, $N_\text{max} > 1 \in \mathbb{N}$\\ 
	\STATE Optimize a generator $G(\cdot; \phi)$ using Algorithm~\ref{alg:train_gen}. \\
	\FOR{$i \in \{1, \ldots, N_\text{max}\}$}
	\STATE Generate $\mathbf{y}^{(i)}$ where $\mathbf{y}^{(i)} = G(\mathbf{z}^{(i)}; \phi_*), \mathbf{z}^{(i)} \sim p_z$. 
	\STATE Sample $\{\xi^{(i)}\}_1^{N_\text{max}}$ from the Poisson process by sampling $\epsilon_k \sim \text{Exp}(1)$ and $\xi^{(i)} = 1 / \sum_{k=1}^{i} \epsilon_k$.
	\ENDFOR
	\STATE Compute the component-wise maxima as: $M = \max_{1 \leq i \leq N_\text{max}} \{ \xi^{(i)} \odot \mathbf{y}^{(i)} \}$.
	\STATE \textbf{Output:} $M$.
	\end{algorithmic}  
\end{algorithm}

\section{Sampling Examples}
\section{Background on EVT}
\label{sec:bg}
The main idea behind EVT is to establish a form of the central limit theorem for the  maxima of appropriately scaled random variables.
\subsection{Main Definitions and Theorems}
\begin{theorem}
If $C$ is a $d-$variate extreme value copula then there exists a tail dependence function $\ell: [0, \infty)^d \to [0, \infty)$ such that:
\begin{equation}
    C \left(u_1, \cdots, u_d \right) = e^{ -\ell\left(-\log u_1, \cdots, -\log u_d \right)}, 
\end{equation}
where $\left(u_1, \cdots,u_d \right) \in (0, 1]^d$.
Using the homogeneity property of $\ell$, the extreme value copula $C$ can be rewritten as: 
\begin{equation}
    \begin{split}
    C \left(u_1, \cdots, u_d \right)  = e^{ \left( \sum_{k=1}^d \log u_k\right) A\left(\frac{\log u_1}{\sum_{k=1}^d \log u_k}, \cdots, \frac{\log u_d}{\sum_{k=1}^d \log u_k} \right)  },
    \end{split}
\end{equation}
where $A$ is known as the Pickands dependence function, which can be thought of as the restriction of $\ell$ to the unit simplex $\Delta_{d-1} = \{ \mathbf{w}=\left( w_1, \cdots, w_d\right) \in [0, \infty)^d: \sum_{k=1}^d w_k=1 \}$. The Pickands function $A$ is known to be fully d-max-decreasing and satisfies: \vspace{-0.2cm}
\begin{equation}\label{pickandsbounds}
\max_{1 \leq k  \leq d }  w_k \leq A(w_1, \cdots, w_d) \leq 1 
\end{equation}
for all $\mathbf{w} = \left(w_1, \cdots, w_d\right) \in \Delta_{d-1}$. 
\end{theorem}
\begin{definition}[Tail dependence function]
A function $\ell: [0, \infty)^d \to [0, \infty)$ is a tail dependence function if for all $\left(x_1, \cdots, x_d \right) \in [0, \infty)^d$, the following conditions are satisfied:
\begin{itemize}
    \item (i) $\ell$ is fully d-max-decreasing and homogeneous of order $1$, i.e. $\ell(cx_1, \cdots, c x_d) = c   \times  \ell(x_1, \cdots, x_d)$, for all $c > 0$. 
    \item (ii) $\max_{ 1 \leq k \leq d} x_k \leq \ell(x_1, \cdots, x_d) \leq \sum_{k=1}^d x_k$.
\end{itemize}
\label{def:stdf}
\end{definition}
\subsection{Spectral Decomposition of Stationary Max-Stable Processes}
Stationary max-stable processes can be intuitively interpreted as i.i.d. samples from infinite dimensional extreme value distributions (i.e. distributions over functions).
A stationary max-stable process can be decomposed by the spectral representation defined in \cite{de1984spectral} which we recall in Proposition~\ref{prop:spectral}. 
\begin{proposition}[Spectral Representation of Max-Stable Processes \citep{de1984spectral}]
\label{prop:spectral}
Suppose that $M(t)$ has unit Fr\'echet margins and is stationary.
Then, $M(t)$ can be written as:
\begin{equation}
    M(t) = \max_{i \geq 1 } \xi_i Y_i^{+}(t), \quad t \in \mathcal{T}.
\label{eqn:spectral_maxstable}
\end{equation}
$\{Y_i(t) \}_{i \geq 1}$ are i.i.d. copies of a continuous stochastic process $Y$ defined on $\mathcal{T}$ such that $\mathbb{E}[Y^{+}(t)]=1$ with $Y^{+}(t) = \max \{0, Y(t) \}$ and $\xi_i$ is the $i$th realization of an independent Poisson point process on $[0, \infty)$ with intensity $\xi^{-2} d \xi$.
\end{proposition}
\vspace{-10pt}





\end{document}